\documentclass{article}

\usepackage{arxiv}

\usepackage[utf8]{inputenc} % allow utf-8 input
\usepackage[T1]{fontenc}    % use 8-bit T1 fonts
\usepackage{hyperref}       % hyperlinks
\usepackage{url}            % simple URL typesetting
\usepackage{booktabs}       % professional-quality tables
\usepackage{amsfonts}       % blackboard math symbols
\usepackage{nicefrac}       % compact symbols for 1/2, etc.
\usepackage{microtype}      % microtypography
\usepackage{lipsum}%
\usepackage{graphicx}%
\usepackage{graphicx}%
\usepackage{multirow}%
\usepackage{amsmath,amssymb,amsfonts}%
\usepackage{amsthm}%
\usepackage[mathscr]{eucal}%
\usepackage[title]{appendix}%
\usepackage{xcolor}%
\usepackage{textcomp}%
\usepackage{manyfoot}%
\usepackage{booktabs}%
\usepackage{algorithm}%
\usepackage{algorithmicx}%
\usepackage{algpseudocode}%
\usepackage{listings}%
\usepackage[edges]{forest}%
\usepackage{pgfplots}%
\pgfplotsset{compat=1.17}%
\usepackage{tikz}%
\usetikzlibrary{arrows.meta}%
\usepackage{array}%
\usepackage{lscape}%
\usepackage{booktabs}%
\usepackage{tabularx}%
\usepackage{rotating}%
\usepackage[utf8]{inputenc}%
\usepackage[T1]{fontenc}%
\usepackage{mathrsfs}%
\usepackage{lmodern}%
\usepackage{orcidlink}%
\graphicspath{ {./images/} }

\title{\textbf{Advancing Talking Head Generation}: A Comprehensive Survey of Multi-Modal Methodologies, Datasets, Evaluation Metrics, and Loss Functions}

\author{
Vineet Kumar Rakesh~\orcidlink{0009-0000-7102-6564} \\
Engineering Sciences, Homi Bhabha National Institute \\
Training School Complex, Anushaktinagar, Mumbai, Maharashtra 400094, India \\
Computer and Informatics Group, VECC \\
1/AF, Bidhannagar, Kolkata, West Bengal 700064, India \\
\texttt{vineet@vecc.gov.in} \\
\And
Soumya Mazumdar~\orcidlink{0009-0006-3521-9557} \\
Department of Computer Science and Business Systems \\
Gargi Memorial Institute of Technology \\
Baruipur, Kolkata, West Bengal 700144, India \\
\texttt{reachme@soumyamazumdar.com} \\
\And
Research Pratim Maity~\orcidlink{0009-0007-5326-4180} \\
Department of Computer Science and Business Systems \\
Gargi Memorial Institute of Technology \\
Baruipur, Kolkata, West Bengal 700144, India \\
\texttt{researchpratimmaity2004@gmail.com} \\
\And
Sarbajit Pal~\orcidlink{0009-0009-5246-7052} \\
Computer and Informatics Group, VECC \\
1/AF, Bidhannagar, Kolkata, West Bengal 700064, India \\
Engineering Sciences, Homi Bhabha National Institute \\
Training School Complex, Anushaktinagar, Mumbai, Maharashtra 400094, India \\
\texttt{sarbajit@vecc.gov.in} \\
\And
Amitabha Das~\orcidlink{0009-0003-1460-8308} \\
School of Nuclear Studies and Application \\
Jadavpur University \\
Salt Lake City, Kolkata, West Bengal 700106, India \\
\texttt{amitabhad.snsa@jadavpuruniversity.in} \\
\And
Tapas Samanta~\orcidlink{0000-0003-0521-0747} \\
Computer and Informatics Group, VECC \\
1/AF, Bidhannagar, Kolkata, West Bengal 700064, India \\
Engineering Sciences, Homi Bhabha National Institute \\
Training School Complex, Anushaktinagar, Mumbai, Maharashtra 400094, India \\
\texttt{tsamanta@vecc.gov.in} \\
}

\begin{document}
\maketitle
\newpage
\begin{abstract}
Talking Head Generation has emerged as a transformative technology in computer vision, enabling the synthesis of realistic human faces synchronized with image, audio, text, or video inputs. This paper provides a comprehensive review of methodologies and frameworks for talking head generation, categorizing approaches into 2D--based, 3D--based, Neural Radiance Fields (NeRF)--based, diffusion--based, parameter-driven techniques and many other techniques. It evaluates algorithms, datasets, and evaluation metrics while highlighting advancements in perceptual realism and technical efficiency critical for applications such as digital avatars, video dubbing, ultra-low bitrate video conferencing, and online education. The study identifies challenges such as reliance on pre--trained models, extreme pose handling, multilingual synthesis, and temporal consistency. Future directions include modular architectures, multilingual datasets, hybrid models blending pre--trained and task-specific layers, and innovative loss functions. By synthesizing existing research and exploring emerging trends, this paper aims to provide actionable insights for researchers and practitioners in the field of talking head generation. For the complete survey, code, and curated resource list, visit our GitHub repository: https://github.com/VineetKumarRakesh/thg.
\end{abstract}

% keywords can be removed

\section{Introduction}

%\footnote{\hyperref[github]{https://github.com/vineet692/thg}

Deep Learning (DL) and Artificial Neural Network (ANN) have altered computer vision, enabling breakthroughs in the area of Talking Head Generation (THG) and synthesis of realistic human faces for speech articulation \cite{gowda-2023}. THG is a research subject that seeks to generate realistic video pictures of human faces that speak in synchronization with arbitrary audio, image, text and others. as input. Today, this technology finds uses in digital avatars, video dubbing in movies, online schooling and video conferencing \cite{shin-2024}.

Early approaches concentrated on 2D--based algorithms applying deep generative models, but they failed to capture 3D structural information, which resulted in less realistic facial dynamics and perspective modifications \cite{shin-2024}. Recent breakthroughs in 3D scene representation approaches, notably Neural Radiance Fields (NeRF) \cite{mildenhall-2020}, have boosted realism, allowed free-view control for movies, and improved picture quality. Innovative frameworks like Wav2NeRF \cite{shin-2024} have emerged to tackle issues such as correctly morphing lip movements in rhythm with audio and managing high-frequency details \cite{shin-2024}. The subject has developed from early rule--based approaches to sophisticated deep learning methods employing Generative Adversarial Networks (GANs) and attention processes \cite{gowda-2023}.

The synthesis of talking heads encompasses portrait production, driving mechanisms, and editing techniques, enabling the adjustment of attributes such as emotion, head posture, and eye blinking \cite{meng-2024}. This study thoroughly examines the existing methodology and approaches for creating THG and assessing algorithms, datasets, and evaluation matrices. It arranges processes into separate groupings, stressing their contributions and drawbacks. Focusing on technical efficiency and perceptual realism is crucial for real-time interaction and high visual quality applications—the research also provides a comparative review of publicly accessible technologies for THG. 

Portrait production has improved dramatically by utilizing generative techniques, including unconditional and conditional methods. Unconditional techniques provide random visuals without prior labels or data inputs, while conditional methods offer controlled outputs based on descriptive inputs \cite{meng-2024}. However, making high-quality, realistic portraits remains tricky due to intricate geometry and appearance. Talking head synthesis, a driving mechanism for animating facial characteristics, has shown remarkable performance utilizing audio--driven and video--driven techniques. Advanced deep learning algorithms have boosted real-time rendering capabilities, yet ongoing constraints like temporal inconsistency and training data necessitate continued effort. Editing in talking head synthesis enables control over avatar modification, although decoupling is challenging due to the connected nature of attributes.

Different sorts of deep learning architecture are utilized to implement any model in THG. Deep learning architectures, such as Convolutional Neural Network (CNN) \cite{6248110} and Recurrent neural network (RNN) ~\cite{10.1001/archneurpsyc.1933.02240140009001}, have proved crucial in AI breakthroughs \cite{mienye-2024A}. CNNs, first created for image processing \cite{mienye-2024A}, have proved their capacity to tackle tough visual problems with better accuracy than earlier approaches \cite{krizhevsky-2017}. RNNs, on the other hand, are created to handle sequence data and have applications in voice recognition and language modeling \cite{sherstinsky-2020}. Including LSTM units has increased the efficacy of these models by eliminating the vanishing gradient issue \cite{mienye-2024A}.

The universal approximation property of neural networks, initially established by Hornik et al. \cite{hornik-1989}, states that a neural network with a single hidden layer may approximate any continuous function on a compact subset of Rn to any desired degree of accuracy. This theorem explains why even primitive systems may convey intricate relationships in data. Neural networks are often divided into shallow and deep learning models, with shallow networks having one or two hidden layers, limited in their ability to learn complex features. In contrast, deep neural networks with multiple hidden layers can perform hierarchical feature extraction and more sophisticated data representations \cite{mienye-2024A}.

Talking-head video creation includes synthesizing lip motion sequences matching a driving source, such as voice or text. This technique also evaluates facial features like emotions and head movements. Early techniques employed cross-modal retrieval \cite{thies2020face2facerealtimefacecapture} and HMM--based algorithms \cite{shen-2023}, but they had high requirements. Recent breakthroughs in deep learning technology have permitted talking-head video generation approaches, utilizing 2D and 3D--based processes. A considerable amount of data must be applied for training, testing, and validation to train a model, build a talking head employing the above mentioned approaches, or even employ pre-existing models properly. Accuracy rises with the quantity of data, but only up to a limit; beyond that, fresh data may bring declining returns in performance. To ensure that the model does not start memorizing the input, rather, the model learns, Backpropagation was invented by Linnainmaa (1976) \cite{linnainmaa-1976}, is a method used to change link weights in neural networks to reduce learning errors. It was popularized in the 1980s and was later developed to correct mistakes during learning. Based on automated differentiation, this technique assures that the model learns without memory retention. It distributes error amounts across connections and determines the gradient of the cost function. Weight updates may be done using techniques such as stochastic gradient descent \cite{ollivier-2015}, extreme learning machines \cite{huang-2006}, no-prop networks \cite{widrow-2012}, weightless networks, and non-connectionist neural networks.

Publicly available datasets have allowed for important improvements in talking head synthesis. The model we have studied indicates that VoxCeleb \cite{nagrani-2017}, VoxCeleb2 \cite{chung-20018}, and TalkingHead--1KH~\cite{tcwang-no-date} are among the most commonly used datasets for video-driven tasks, while CREMA-D~\cite{cheyneycomputerscience-no-date}, LRW~\cite{chung-2016}, and MEAD~\cite{kaisiyuan2020mead} have become prominent datasets for audio-driven tasks, according to our several recent surveys on talking head generation. Large-size, high-resolution datasets like FFHQ~\cite{nvlabs-no-date}, HDTF~\cite{zhang-2021}, and CelebV-HQ~\cite{zhu2022celebvhqlargescalevideofacial} evolved as a result of the growing necessity for higher image quality brought about by the rise of application scenarios. In-depth details about each dataset, such as picture size, modality, and subject perspective, have been acquired; these factors have not been fully investigated in past evaluations of talking head synthesis.

While training a model, a critical component in training models for optimal accuracy is the design and implementation of a robust loss function, which assesses the difference between anticipated outputs and actual ground truth. A loss function is used in training models to improve parameters by examining the difference between anticipated and expected outcomes \cite{terven-2023}. Although convexity assures that any local minimum is also a global minimum, which is theoretically ideal, many effective deep learning loss functions are non-convex. However, they may be efficiently improved utilizing modern approaches \cite{terven-2023}. An efficient loss function is created to be resilient to outliers and differentiable to allow smooth gradient--based optimization. Convex loss functions are popular since they may be optimized using gradient--based approaches. Differentiability is vital for allowing gradient--based optimization. Robustness indicates that loss functions can support outliers without being influenced by a small number of extreme values. Smoothness provides a steady gradient without sudden transitions or spikes. Sparsity encourages sparse output, suited for high-dimensional data and tiny features. In non-convex conditions, a monotonic drop in the loss function does not assure convergence to the optimal solution, even if it would demonstrate consistent optimization progress. 

As per Chen et al. \cite{chen-2020A}, the performance of talking head synthesis models is significantly examined based on four fundamental criteria which are identity preservation, visual quality, lip synchronization, and natural motion. However, models usually focus on increases in a single component and qualitative and quantitative assessments are commonly applied. Quantitative measures offer a more objective analysis, but qualitative judgments rely on direct observation and could be subjective. This section includes some well-known quantitative criteria for a complete examination of talking head synthesis models.

Additionally, the multilingual component of talking head synthesis is another problem. The paucity of annotated, high-quality datasets across numerous languages, with diverse phonetic patterns, lip movements, and cultural intricacies, inhibits the general deployment of these systems \cite{liu-2024}. To tackle this challenge, researchers should foster multilingual datasets and employ sophisticated methodologies like self-supervised and transfer learning. Complementary options include approaches like Long Short-Term Memory (LSTM) ~\cite{10.1162/neco.1997.9.8.1735} networks to acquire mouth landmarks from audio, the Wav2Lip model \cite{prajwal-2020} to automate lip-syncing, and a strategy for reenactment that stresses visual signals for managing different voices and modifying face gestures depending on audio-derived speech styles \cite{thies-2019}.

The methodical technique clarifies the present situation and brings intriguing prospects for further inquiry. This article seeks thorough insights, including for researchers in the THG research topic. Despite the great benefits of THG, it tackles several issues, such as its dependence on pre--trained models, the management of extreme postures, and the necessity for high-quality datasets and sophisticated algorithms \cite{liu-2024}. These limits may hamper innovation and adaptation for varied use cases. Future research could consider training individual components or modules on large-scale, diverse datasets to develop more modular and adaptable frameworks. Hybrid architectures that blend pre--trained models with task-specific layers or modules may give both pre--trained knowledge and targeted flexibility advantages. To solve problems such as large-angle poses, researchers aim to enhance datasets and apply multi-view training methodologies to capture a larger range of face orientations \cite{liu-2024}. Maintaining temporal consistency is critical for providing smooth visual outputs, and tackling this involves high-quality datasets and innovative approaches \cite{liu-2024}.

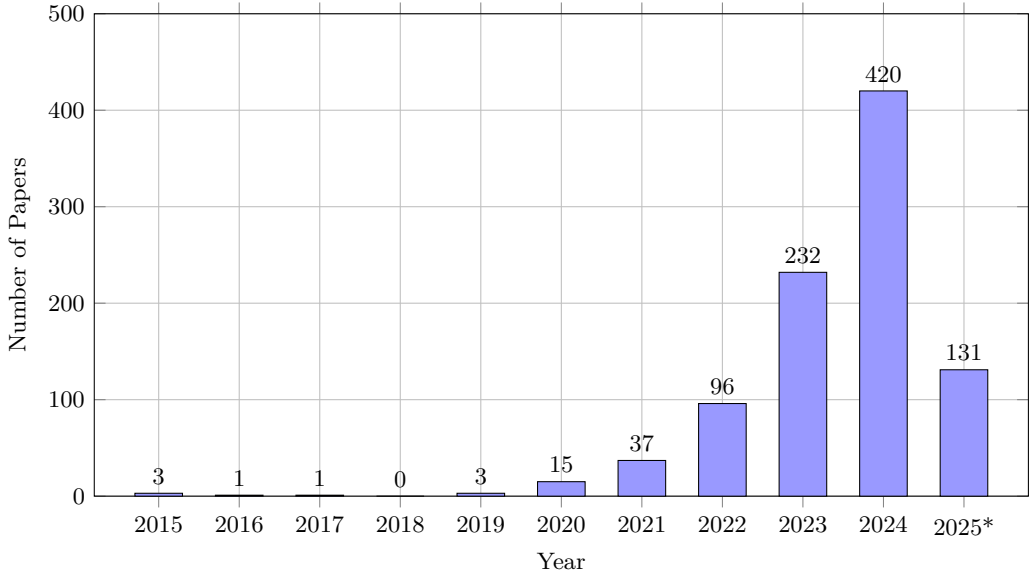
\begin{figure}[htbp]
    \centering
    \begin{tikzpicture}
    \label{fig:enter-label}
    \begin{axis}[
        ybar,
        bar width=18pt,
        width=14cm,
        height=8cm,
        ymin=0,
        ymax=500,
        ytick={0,100,200,300,400,500},
        ylabel={Number of Papers},
        xlabel={Year},
        symbolic x coords={2015,2016,2017,2018,2019,2020,2021,2022,2023,2024,2025*},
        xtick=data,
        nodes near coords,
        nodes near coords align={vertical},
        every node near coord/.append style={font=\small},
        enlarge x limits=0.08,
        grid=major,
        tick label style={font=\small},
        label style={font=\small}
    ]
    \addplot[fill=blue!40!white] coordinates {
        (2015,3)
        (2016,1)
        (2017,1)
        (2018,0)
        (2019,3)
        (2020,15)
        (2021,37)
        (2022,96)
        (2023,232)
        (2024,420)
        (2025*,131)
    };
    \end{axis}
    \end{tikzpicture}
    \caption{Number of papers published per year with the keyword ``Talking Head Generation'' from 2015 to 2025\protect\footnotemark}
    \footnotetext{Publication till April, 2025.}
\end{figure}

The field of THG is experiencing consistent growth each day. Figure \ref{fig:enter-label} illustrates the annual number of reported works in this area based on data collected from Google Scholar, including publications from academic publishers, professional societies, online repositories, universities, and other relevant websites under the keyword "THG" and its associated terms. For 2025, the publication count covers the period from January through April.

A comprehensive survey has been developed, categorizing various highly cited scholarly works based on input modality, model architecture, and training paradigms. This framework enhances our understanding of the evolution of THG.
\begin{itemize}
  \item An in-depth analysis of critical components, including portrait generation, driving mechanisms, and editing techniques, has been conducted. This analysis reveals the technical trade-offs associated with identity preservation, realism, and controllability.
  \item A thorough comparison of widely utilized datasets has been performed. This comparison emphasizes key factors: resolution, modality, subject diversity, and applicability to various learning tasks.
  \item An extensive review of commonly employed loss functions and evaluation metrics has been undertaken. This review addresses their significance in optimizing model performance and improving perceptual quality.
  \item Several key research gaps have been identified, including multilingual synthesis, pose diversity, temporal consistency, and the risk of over reliance on pretrained models. Recommendations for future research include the exploration of modular architectures and the adoption of diverse training strategies.
\end{itemize}

Although several recent overviews Gowda et al. (2023)~\cite{gowda-2023} and Nguyen-Le et al. (2024)~\cite{nguyen-le-2024} provide broad surveys of 2D/3D pipelines up to 2024, but offer limited discussion of emerging neural radiance field and diffusion-based approaches. Dedicated NeRF reviews~\cite{mildenhall-2020} and general 3D generation summaries~\cite{li-2024} tend to treat audio-driven lip-sync and modular design only peripherally. At the same time, security-focused deepfake detection works \cite{nguyen-le-2024} \cite{liu-2024} do not fully trace the evolution of underlying generative methodologies. Similarly, existing evaluations of perceptual metrics \cite{gandhi-2024} and foundational THG principles \cite{chen-2020A} predate recent advances in real-time rendering and temporal consistency. By contrast, the present survey spans innovations from 2017 through April 2025, offering detailed architectural and loss-function comparisons across ten distinct THG paradigms and explicitly linking each method to a dozen practical applications—ranging from video conferencing and remote education to visual effects. It is also the first to unify 3D Gaussian splatting and latent consistency models within a single analytical framework, and it proposes a set of ethical safeguards—such as robust watermarking protocols and the curation of multilingual datasets—to support responsible deployment of THG technologies.

\begin{figure}[htbp]
\centering
\resizebox{0.9\textwidth}{!}{%
\begin{forest}
for tree={
  draw,
  rounded corners,
  align=center,
  font=\sffamily\small,
  text width=4cm,
  edge={->, >={Stealth[length=3mm, width=2mm]}, line width=0.4pt, shorten >=1pt},
  minimum height=0.6cm,
  grow'=0,
  child anchor=west,
  parent anchor=east,
  l sep=1.8cm,
  s sep=2mm,
  fill=white,
}
[{\textbf{Advancing THG}}, fill=blue!20
 [1. Introduction, fill=green!30]
 [2. THG, fill=orange!30
  [2.1 Methodology]
  [2.2 Dataset]
  [2.3 Loss Function]
  [2.4 Evaluation Metrics]
 ]
 [3. Test Evaluation, fill=pink!30
  [3.1 Experimental Setup]
  [3.2 Results and Discussion]
 ]
 [4. Limitation \& Challenge, fill=gray!30]
 [5. Conclusion, fill=gray!20]
 [6. Future Direction, fill=gray!20]
 [Acknowledgement, fill=gray!10]
 [References, fill=gray!10]
]
\end{forest}
}
\caption{Structure of the Review Paper on Talking Head Generation}
\label{fig:talking-head-outline}
\end{figure}
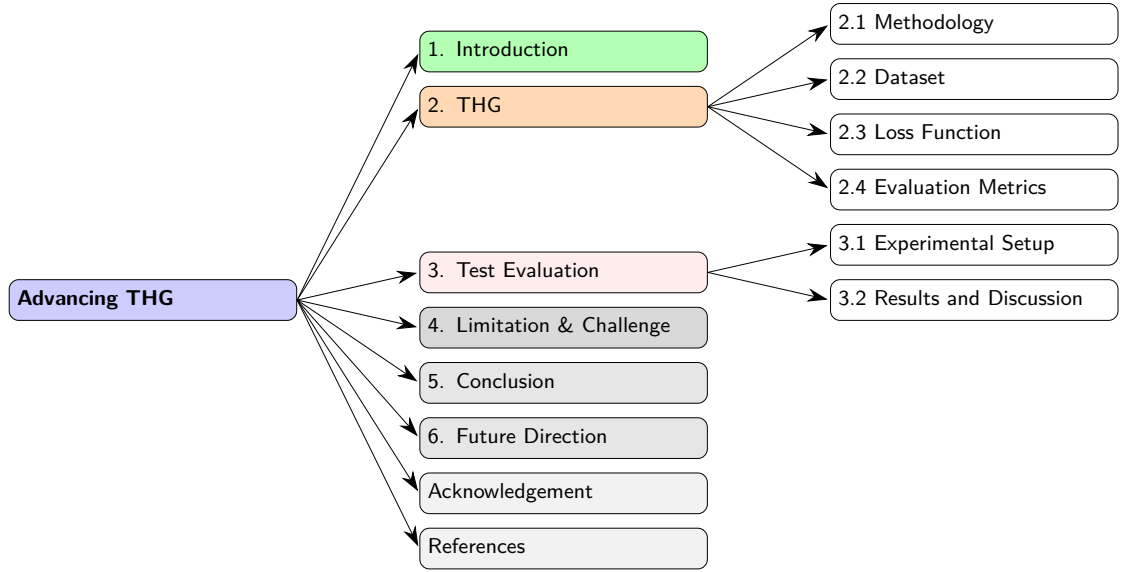

This article outlines main ideas of talking head development, research scope, ethical difficulties, and methods. It describes the architecture of a proposed system and discusses training and inference processes, the dataset, loss function, and evaluation metrics. A complete overview of the whole research may be viewed in the below Figure \ref{fig:talking-head-outline}.

\section{Talking Head Generation (THG)}
Walter Pitts and Warren McCulloch introduced the foundational concepts of Deep Learning (DL) and Artificial Neural Networks (ANNs) in 1943 \cite{mcculloch-1943}, who showed the capacity of theoretical artificial neuron networks to complete basic logical tasks. However, recent developments in processing power, notably the introduction of high-performance Graphics Processing Units (GPUs), have substantially facilitated the deployment of complicated deep learning models, such as those applied in THG \cite{hagos-2024}. Our investigation thoroughly examined over few hundreds of articles linked to the THG. 

\subsection{Comparison with Existing Relevant Review Articles}

This comprehensive review identified approximately 100 methods for creating talking heads that were frequently referenced in various credible academic sources and selected for further research. The papers examined in these publications were released mainly between 2017 and April 2025. We also included a selection of previous works to trace the origins and foundational frameworks of the subject. The evaluation of the systems for THG focused on their performance in ten essential areas: picture--based, audio--based, text--based, video--based, 2D--based, 3D--based, distortion--based, Neural Radiance Fields (NeRF) based, parameter efficiency, and 3D animation. Many techniques are used in the THG process to create lifelike human faces coordinated with parametric, textual, video, or audio inputs. This section divides these approaches into ten paradigms by examining their designs, datasets, advantages, disadvantages, and applications.

\subsubsection{Image--based THG}
Image--based THG is a technique that animates a human head by transferring motion information learned from a driving video. This method generates realistic head movements and facial expressions from a single source image. The base architecture consists of four core components: motion encoding via self-learned key points, a local affine transformation module, an occlusion aware generator, and an extended equivariance loss. The system learns to extract key points from the source image, encoding the motion information by tracking their trajectories. The architecture utilizes local affine transformations to capture complex motion dynamics. The occlusion-aware Generator fills in or infers missing information, ensuring the generated frames remain coherent even during significant motion or occlusion events.
\begin{figure}
  \centering
  \includegraphics[width=1\linewidth]{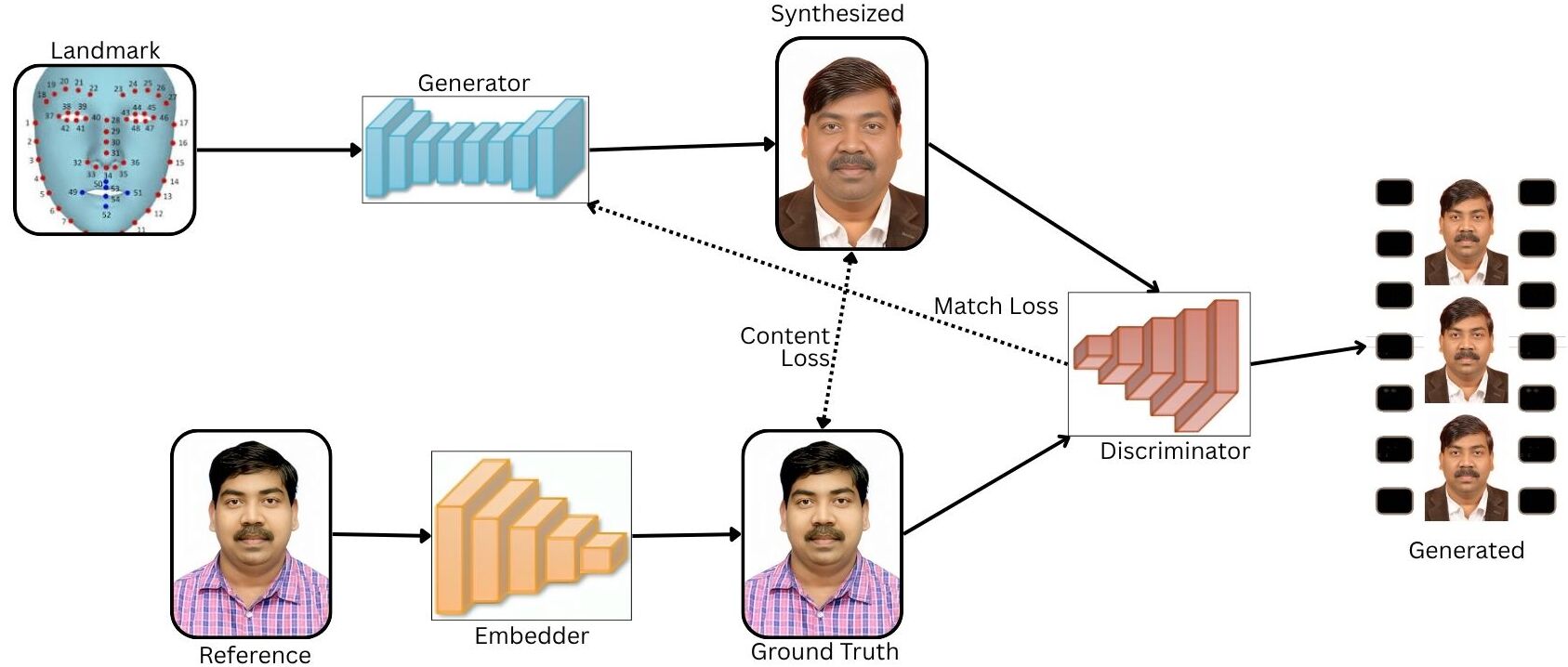}
  \caption{Generalized Approach for Image based}
  \label{Figure.2.1.1.}
\end{figure}
A theoretical framework for synthesizing talking heads using images is presented, along with a neural network architecture similar to a GAN, in \ref{Figure.2.1.1.}, supplemented by an embedding and landmark--based motion module. The model has four components: Input: Landmark and Source Image; a generator that utilizes landmarks and identity information from the source image; an embedder that extracts identity features from the source image to inform the Generator; synthesized output that produces a frame emulating the pose/expression of the input landmark while retaining the identity from the source image; ground truth representing the authentic target frame corresponding to the historic; loss functions that evaluate pixel/feature similarity and embedding similarity between generated and ground truth images to maintain identity and realism; a discriminator that attempts to differentiate between real and synthesized images to yield more realistic frames; Output: generated video, a sequence of synthesized frames that forms a talking head video, where head movements and expressions correspond to the landmark inputs. The shown approach generates realistic talking head films from a single source picture and driving markers, often derived from an alternative video. Loss functions such as Content Loss and Match Loss guarantee realism and identity coherence, while the Discriminator compels the Generator to provide photorealistic outcomes.

\begin{sidewaystable}[htbp]
\scriptsize
\caption{Comparison of Image Based Approaches}
\label{tab:image_based_1}
\centering
\begin{tabular}{@{}lll>{\raggedright\arraybackslash}p{4.2cm}>{\raggedright\arraybackslash}p{4.2cm}l@{}}
\toprule
\textbf{Method} & \textbf{Arch.} & \textbf{Dataset} & \textbf{Highlights} & \textbf{Limitations} & \textbf{N-shot} \\
\midrule
SMA~\cite{zhao2025synergizingmotionappearancemultiscale} & VQGAN & VoxCeleb1~\cite{nagrani-2019}, CelebV-HQ~\cite{zhu2022celebvhqlargescalevideofacial} & Joint motion + appearance codebooks for smooth motion flow. & Keypoint motion estimation needed to reduce appearance leakage. & One \\
TS-Net~\cite{ni2022crossidentityvideomotionretargeting} & GAN & FaceForensics & Dual-branch with warp-free and GRID~\cite{10.1121/1.5042758} transformation. & Inconsistent motion, lacks high-freq detail. & One \\
DaGAN~\cite{hong2022depthawaregenerativeadversarialnetwork} & GAN & VoxCeleb1~\cite{nagrani-2019}, CelebV & Self-supervised geometry for synthetic faces. & Quality degrades with poor input; costly. & One \\
SAFA~\cite{wang2021safastructureawareface} & FAN & Voxceleb1 & 3DMM-driven structure-aware animation. & Sensitive to occlusion, overfitting. & One \\
Face2Face\cite{thies2020face2facerealtimefacecapture} & CNN & Face2Face & Real-time expression transfer from monocular view. & Lambertian assumptions + latency issues. & One \\
CrossID-GAN~\cite{9157543} & GAN & 300VW & Multi-ID reenactment with landmark mapping. & Needs real-time and accurate landmarks. & One \\
MarioNETte~\cite{ha2019marionettefewshotfacereenactment} & U-Net & VoxCeleb1~\cite{nagrani-2019}, CelebV & Identity via image attention + landmark transform. & Pose and detection errors limit scale. & One \\
FOMM~\cite{siarohin-2020} & Monkey-Net & Multiple & Decouples appearance/motion; handles occlusion. & Requires object-specific tuning + data. & One \\
AAO~\cite{siarohin2019animatingarbitraryobjectsdeep} & Monkey-Net & Multiple & Dense motion + keypoint prediction. & Fails with poor keypoints. & One \\
ReenactGAN~\cite{wu2018reenactganlearningreenactfaces} & GAN & WFLW, CelebV & Reenactment via boundary space transfer. & Slow, data hungry. & One \\
X2Face~\cite{wiles2018x2facenetworkcontrollingface} & U-Net & VGG-Face, VoxCeleb & Self-supervised pose/exp control. & Poor generalization to large poses. & One \\
GATH~\cite{pham-2018} & CNN & CACD, GTAV & AU--based expression synthesis. & Ethical concerns + data load. & One \\
\bottomrule
\end{tabular}
\end{sidewaystable}

The numerous perspectives of image--based THG, such as its highlights, limitations with the main dataset for training, and basic architecture of the various models, are evaluated in Table \ref{tab:image_based_1}, focusing on ways to employ one-shot for creation. The primary issues involve disentangling and altering the THG's look and motion components. The SMA~\cite{zhao2025synergizingmotionappearancemultiscale} technique employs VQGANs and jointly learns codebooks for both aspects, allowing for more relatable control over motion flows. TS-Net~\cite{ni2022crossidentityvideomotionretargeting}, a dual-branch technique, promotes identity retention and resistance against occlusions. DaGAN~\cite{hong2022depthawaregenerativeadversarialnetwork}, a GAN--based technique, focuses on self-supervised geometry learning. However, its performance may be impeded by extreme postures and substantial occlusions. SAFA~\cite{wang2021safastructureawareface}, a structure-aware technique, integrates 3D Morphable Models (3DMMs) to produce more realistic animations with higher perceptual quality. Face2Face\cite{thies2020face2facerealtimefacecapture}, a revolutionary technology for real-time face replication, exhibits possibilities for live interactive applications.
CrossID-GAN~\cite{9157543}, a network intended for multi-identity face reconstruction, displays resilience and efficacy in qualitative and quantitative tests. MarioNETte~\cite{ha2019marionettefewshotfacereenactment}, a U-Net--based architecture, focuses on strong identity retention. However, its performance depends on reliable landmark recognition and issues managing significant pose fluctuations, boosting inference speed, and scaling to higher-definition movies. The FOMM~\cite{siarohin-2020} is a flexible framework for picture animation that can handle many object types beyond faces. "Monkey-Net" architecture learns motion representations from driving footage and applies them to source photos. AAO~\cite{siarohin2019animatingarbitraryobjectsdeep}refines the picture animation pipeline by employing a key point detector, a Dense Motion prediction network, and a Motion Transfer Network. ReenactGAN~\cite{wu2018reenactganlearningreenactfaces} seeks photorealistic and real-time face reenactment by transferring facial motions from a source to a target video. However, its success relies on the availability of adequate training data and may struggle with hidden identities or extreme positions. X2Face~\cite{wiles2018x2facenetworkcontrollingface} provides a neural network model that uses another face or modality as input to change the expression and location of a target face. GATH~\cite{pham-2018} uses continuous Action Unit (AU) coefficients to generate facial emotions from still images automatically. Quality and diversity are the cornerstones of its success. Furthermore, it is important to address the ethical issues of face expression synthesis carefully. Future research in image--based THG will primarily address these issues, explore more reliable and effective architectures, develop innovative methods for handling challenging scenarios, enhance the temporal coherence of produced videos, and investigate strategies that can generalize successfully to unseen identities while maintaining high fidelity and controllability.

In addition to the most cited paper, various models and their applications in the fields, including facial recognition, video conferencing, and facial recognition. Tencent has developed the HunyuanPortrait ~\cite{xu2025hunyuanportraitimplicitconditioncontrol} model in 2025, while ByteDance has developed the X-Portrait ~\cite{xie2024xportraitexpressiveportraitanimation} model in 2023. IIIT Hyderabad developed the AVFR-GAN ~\cite{agarwal2022audiovisualfacereenactment} in 2022, an audio-visual face reenactment model animating a source image by transferring head motion from a driving video. Tsinghua University has developed one-shot high-resolution editable talking face generation via pre-trained StyleGAN \cite{karras2019stylebasedgeneratorarchitecturegenerative} in 2022. Samsung AI has developed one-shot megapixel neural head avatars focusing on cross-driving synthesis in 2022. Finally, Samsung AI has developed a multi-dimensional face recognition model called MegaPortraits \cite{drobyshev2023megaportraitsoneshotmegapixelneural} in 2022. These models create realistic, realistic, and interactive facial recognition models. These models have been developed to create realistic, realistic, and interactive facial recognition models. Various institutions and researchers have developed them, and their use in various fields has shown promising results. A few other relatively highly cited models include LivePortrait~\cite{guo2025liveportraitefficientportraitanimation}, SMA~\cite{zhao2025synergizingmotionappearancemultiscale}, a multiscale framework that synergizes motion and appearance for enhanced realism; MCNET~\cite{hong2023implicitidentityrepresentationconditioned}, which leverages implicit identity representations conditioned on input frames for coherent video generation; TPSM~\cite{zhao2022thinplatesplinemotionmodel}, introducing thin‑plate spline motion modeling for smooth facial deformations; StyleHEAT~\cite{yin2022styleheatoneshothighresolutioneditable}, an editable, high‑resolution one‑shot facial animation technique; DAM~\cite{tao2022structureawaremotiontransferdeformable}, which integrates structure‑aware deformable motion transfer; StyleMask~\cite{bounareli2022stylemaskdisentanglingstylespace}, disentangling style spaces to offer fine‑grained control over facial attributes; AniFaceGAN~\cite{wu2022anifacegananimatable3dawareface}, providing 3D‑aware animatable face generation; IW~\cite{mallya2022implicitwarpinganimationimage}, using implicit warping for seamless image animation; LIA~\cite{wang2022latentimageanimatorlearning}, a latent image animator that learns dynamic representations directly in the latent space; and many others. The state of the art in producing realistic, interactive, and high-fidelity facial representations has been greatly enhanced by these models, which collectively cover a wide range of capabilities, from effective, real-time animation to ultra-high-resolution editable outputs.

\subsubsection{Audio--Based THG}

Audio-driven techniques use cross-modal alignment and acoustic feature extraction to synthesize facial emotions and lip movements from speech data. Google Scholar citation metrics show that Talk3D~\cite{ko2024talk3dhighfidelitytalkingportrait}, EMO\cite{tian2024emoemoteportraitalive}, and GC-AVT~\cite{9878472} are important models. While EMO~\cite{tian2024emoemoteportraitalive} employs a ReferenceNet with dynamic audio-visual attention to capture subtle emotional expressions, Talk3D~\cite{ko2024talk3dhighfidelitytalkingportrait} uses a 3D-aware generative prior and audio-guided attention U-Net. Editing emotions and styles is possible with GC-AVT~\cite{9878472}, although complicated backdrops provide difficulties. Challenges include limited datasets for non-English phonetic patterns and speech variability. Future approaches may involve real-time rendering pipelines for low-latency applications and self-supervised learning for cross-lingual adaptability.
\begin{figure}
  \centering
  \includegraphics[width=1\linewidth]{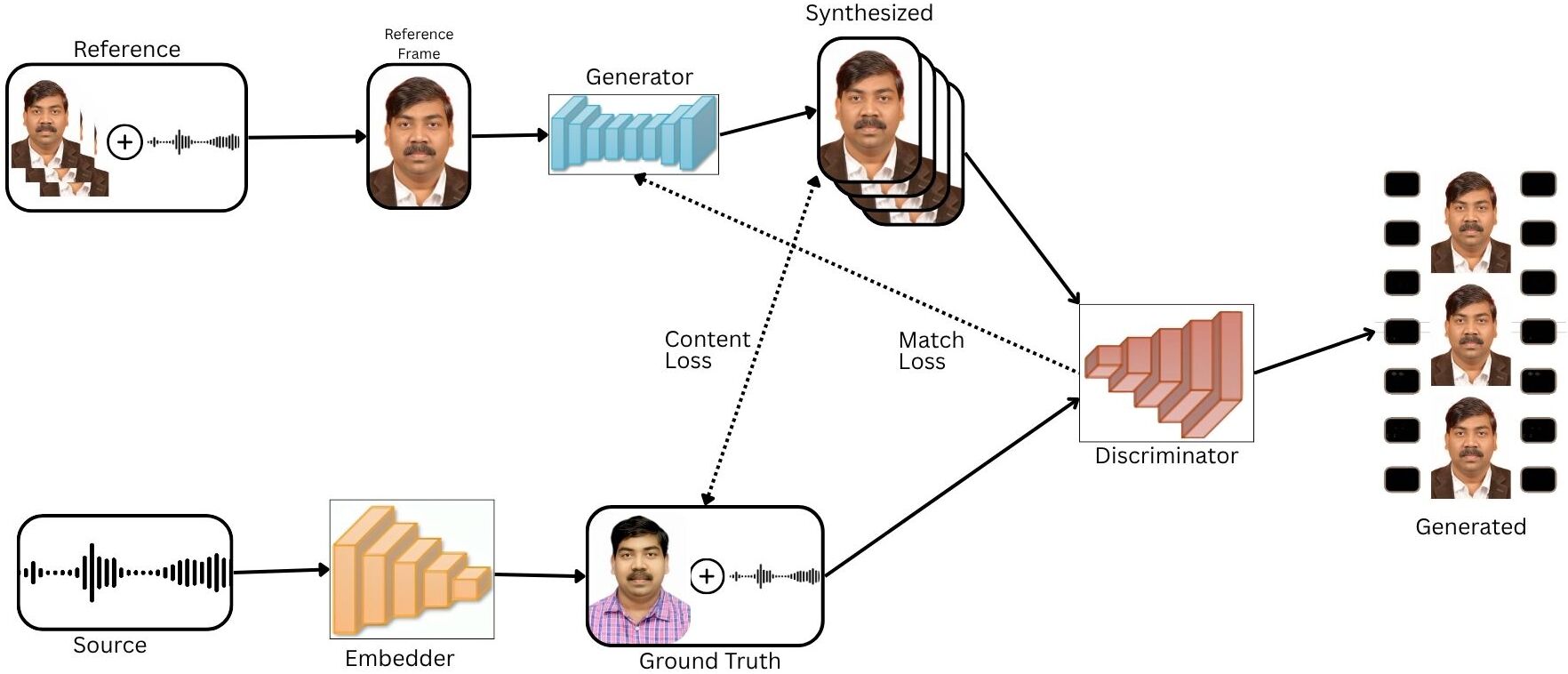}
  \caption{Generalized Approach for Audio based}
  \label{fig:2.1.2}
\end{figure}
A theoretical framework for audio--based talking head synthesis is shown with a neural network architecture akin to a GAN in Figure \ref{fig:2.1.2}, supplemented by an embedding and landmark--based motion module, which involves a pre--trained model (implicitly represented by the Generator and Discriminator) and a reference visual identity (multiple image frames). The input is the source audio, which determines the required speech and lip movements of the created talking head. The reference video illustrates the visual identity of the person speaking. The model components include the Embedder, which extracts phonetic and prosodic information from the source audio, and the Generator, which synthesizes a new sequence of video frames where the person from the Reference Frame appears to be speaking the content of the Source Audio, with synchronized lip movements and potentially matching head poses or expressions. The Discriminator operates as an antagonistic critic, receiving two inputs: the synthesized video frames created by the Generator and the Ground Truth. Its purpose is to discern between the realistically seeming synthetic video and genuine video frames, producing a score reflecting its confidence in the "realness" of the input. The figure \ref{fig:2.1.2} highlights two types of losses that guide the training of the Generator: Content Loss, which measures the difference between the content (e.g., lip movements, facial expressions) of the Synthesized video and the Ground Truth, and Match Loss, derived from the output of the Discriminator, encouraging the Generator to produce video frames that are indistinguishable from real video frames. By attempting to "fool" the Discriminator, the Generator learns to generate increasingly photorealistic and natural-looking talking head sequences. The trained model's end product is a video that shows the individual from the Reference Video speaking the content of the Source Audio, complete with realistic facial expressions and well-coordinated lip motions.  This simplified explanation outlines a common method for generating audio-driven talking heads using deep learning techniques, likely involving a Generative Adversarial Network (GAN) architecture. It highlights the importance of a reference image for establishing visual identity, audio embeddings for controlling facial movements, and adversarial learning to produce realistic outputs.

\begin{sidewaystable}[htbp]
\scriptsize
\caption{Comparison of Audio Based Approaches}
\label{tab:audio_based_1}
\centering
\begin{tabular}{@{}lll>{\raggedright\arraybackslash}p{4.2cm}>{\raggedright\arraybackslash}p{4.2cm}l@{}}
\toprule
\textbf{Method} & \textbf{Arch.} & \textbf{Dataset} & \textbf{Highlights} & \textbf{Limitations} & \textbf{N-shot} \\
\midrule
OmniHuman-1~\cite{lin2025omnihuman1rethinkingscalinguponestage} & 3DVAE & CelebV-HQ~\cite{zhu2022celebvhqlargescalevideofacial}, RAVDESS~\cite{livingstone-2018} & Diffusion Transformer with motion conditioning for realism. & Scalability, data dependency, and ethical concerns. & One \\
EMO~\cite{tian2024emoemoteportraitalive} & RefNet & CREMA & Audio–facial correlation enhances realism. & Expression complexity, quality, dependency. & One \\
Talk3D~\cite{ko2024talk3dhighfidelitytalkingportrait} & GAN & AD-NeRF~\cite{guo2021adnerfaudiodrivenneural}, Obama Weekly & Audio-driven 3D prior and attention U-Net. & Prep complexity, artifacts, poor generalization. & Multi \\
MODA~\cite{liu2023modamappingonceaudiodrivenportrait} & LSGAN & HDTF~\cite{zhang-2021}, LSP~\cite{lu2021livespeechportraitsrealtime} & Dual-attention and facial composer. & Expression variance, compute cost. & One \\
GC-AVT~\cite{9878472} & GAN & VoxCeleb2~\cite{chung-20018}, MEAD~\cite{kaisiyuan2020mead} & Granular control over lip, pose, and expression. & Low-res output and poor backgrounds. & Multi \\
Flow-guided~\cite{gu2025flowguideddiffusionvideoinpainting} & 3DMM & HDTF~\cite{zhang-2021} & AV flow-guided with motion realism. & Coherence, cropping, and style gaps. & One \\
Audio2Head~\cite{wang2021audio2headaudiodrivenoneshottalkinghead} & RNN & LRW, GRID~\cite{10.1121/1.5042758}, VoxCeleb & Predicts pose and motion from audio/image. & Identity mismatch, ethical misuse. & One \\
DAVS~\cite{zhou2019talkingfacegenerationadversarially} & GAN & LRW & Enhances lip realism and intelligibility. & Speed and variability limitations. & One \\
LMGG~\cite{chen2018lipmovementsgenerationglance} & GAN & GRID~\cite{10.1121/1.5042758}, LDC, LRW & Cross-modal speech–lip fusion. & Realism + scale lacking, costly. & Multi \\
Facial Reenact.~\cite{hornik-1989} & GAN & CelebA & cGAN + RNN for audio–face sync. & Speaker variability, weak realism. & Multi \\
VisemeNet~\cite{zhou2018visemenetaudiodrivenanimatorcentricspeech} & LSTM & GRID~\cite{10.1121/1.5042758}, SAVEE, BIWI 3D & LSTM viseme curves improve sync robustness. & emotion diversity is limited. & Multi \\
\bottomrule
\end{tabular}
\end{sidewaystable}

The various approaches to audio--based THG are reviewed in Table \ref{tab:audio_based_1} This review highlights the strengths and limitations of the major datasets used for training and the fundamental architecture of different models. One notable approach is OmniHuman--1, which utilizes a Diffusion Transformer--based architecture for end-to-end human animation, producing incredibly realistic human films. It adds motion-related circumstances directly into the training process, resulting in excellent realism. EMO~\cite{tian2024emoemoteportraitalive}, another N-shot technique, focuses on the dynamic interaction between auditory cues and facial movements, seeking to capture more expressive facial signals. Talk3D~\cite{ko2024talk3dhighfidelitytalkingportrait}, another N-shot technique, stresses the realistic reconstruction of face geometry using a tailored 3D-aware generative prior and an audio-guided attention U-Net architecture. However, the drawbacks of Talk3D~\cite{ko2024talk3dhighfidelitytalkingportrait} include its lack of generalizability beyond lifelike human faces and its probable limits in non-human characters. The article explores numerous ways to make high-fidelity, multi-person talking portraits, including MODA~\cite{liu2023modamappingonceaudiodrivenportrait}, GC-AVT~\cite{9878472}, Flow-guided One-shot, Audio2Head~\cite{wang2021audio2headaudiodrivenoneshottalkinghead}, and DAVS~\cite{zhou2019talkingfacegenerationadversarially}. MODA~\cite{liu2023modamappingonceaudiodrivenportrait} is a complete system that concentrates on audio and visual characteristics.
 GC-AVT~\cite{9878472} is an audio-visual talking head model that allows granular control over lip movements, head positions, and facial expressions. GC-AVT~\cite{9878472} largely focuses on accommodating varied speakers and a spectrum of emotions in speech. Flow-guided One-shot~\cite{zhang-2021} is a flow-guided talking face creation framework, particularly built for high-definition facial movies. Audio2Head~\cite{wang2021audio2headaudiodrivenoneshottalkinghead} is an audio-driven talking-head approach that seeks to make photorealistic films from a single reference picture of the target individual. DAVS~\cite{zhou2019talkingfacegenerationadversarially} is a one-shot learning technique that leverages a single reference picture for numerous identity and audio inputs. However, the study notes the potential for exploitation of such technology for harmful reasons and aims to disclose code and models to minimize this. DAVS~\cite{zhou2019talkingfacegenerationadversarially} is a system that utilizes voice recordings to build realistic facial pictures, concentrating on enhancing lip motion patterns. It has potential uses in automated lip reading and video retrieval. The "LMGG" is a one-shot solution that integrates audio and visual embeddings to establish synchronization between produced lip movements and speech. However, it suffers constraints in accuracy, generalization, photorealism, and processing expenses. The "N-Shot" approach employs recurrent neural networks and conditional generative adversarial networks to build photorealistic faces with accurate lip synchronization. The VisemeNet \cite{zhou2018visemenetaudiodrivenanimatorcentricspeech} approach is meant to produce animator-centric speech motion curves but has issues addressing speaker variability and introducing emotional context. Overall, these audio--based THG algorithms indicate breakthroughs in human animation but also confront issues in scalability, data reliance, and controlling computing needs for real-time applications.

Audio-based models have seen a significant expansion with several innovative models. These include ACTalker~\cite{hong2025audiovisualcontrolledvideodiffusion} (2025) integrates multimodal control signals—including pose, expression, and audio—to guide high‐fidelity video diffusion. AniPortrait~\cite{wei2024aniportraitaudiodrivensynthesisphotorealistic} (2024) transforms static portraits into expressive animations driven solely by speech. EDTalk~\cite{tan2024edtalkefficientdisentanglementemotional} (2024) employs efficient disentanglement to capture and transfer emotional nuances in facial movements. EchoMimic~\cite{chen2024echomimiclifelikeaudiodrivenportrait} (2024) mimics vocal subtleties to produce lifelike, audio‐driven portrait animations. FD2Talk~\cite{yao2024fd2talkgeneralizedtalkinghead} (2024) generalizes talking‐head generation across diverse faces and speaking styles with a unified architecture. FaceChain‐ImagineID~\cite{xu2024facechainimagineidfreelycraftinghighfidelity} (2024) offers free‑form identity manipulation to craft high‑fidelity talking heads from arbitrary source images. FlowVQTalker~\cite{tan2024flowvqtalkerhighqualityemotionaltalking} (2024) leverages flow‑based quantization to achieve emotionally rich and identity‐consistent talking faces. MuseTalk~\cite{zhang2025musetalkrealtimehighfidelityvideo} (2025) operates in a latent video space to deliver real‐time, high‐fidelity lip‐sync results on unseen speakers. ReSyncer~\cite{guan2024resyncerrewiringstylebasedgenerator} (2024) provides a plug‑and‑play framework for resynchronizing any facial model with arbitrary audio inputs, enhancing lip‐sync accuracy. Real3DPortrait~\cite{ye2024real3dportraitoneshotrealistic3d} (2024) reconstructs and animates realistic 3D avatars from a single image using audio cues. Emotional Conversation~\cite{liang2024emotionalconversationempoweringtalking} (2024) empowers talking heads with contextual emotional modulation to make dialogues more engaging. Make Your Actor Talk~\cite{yu2024makeactortalkgeneralizable} (2024) focuses on generalizable audio‑driven actor synthesis across varied domains. RealTalk~\cite{ji2024realtalkrealtimerealisticaudiodriven} (2024) achieves real‑time, high‑fidelity audio‑driven portrait generation optimized for interactive applications. AAAI~\cite{tan2024saystyle} (2024) incorporates style‐transfer techniques to produce stylized talking‐head animations. Style2Talker~\cite{tan2024style2talkerhighresolutiontalkinghead} (2024) extends this by generating high‑resolution talking‐head videos with fine‑grained style control. A model like ManiTalk~\cite{fang-2024} (2024) advances audio-driven talking head generation by enabling explicit manipulation of facial details such as eyebrows, eyelids, and pupils through a three-stage pipeline. The method combines synchronized landmark generation with parameterized facial control and image warping, achieving state-of-the-art lip synchronization and identity preservation. Its focus on stylized expression control aligns with emerging trends in personalized avatar synthesis. Finally, THQA~\cite{zhou2024thqaperceptualqualityassessment} (2024) introduces a perceptual quality assessment metric specifically designed for evaluating talking‑head animations. Together, these innovations significantly broaden the capabilities of audio‑driven facial animation, enabling more expressive, controllable, and realistic talking heads.

\subsubsection{Video--Based THG}

Video-driven techniques focus on preserving identity and ensuring temporal coherence by replaying source movies while transferring motion from driving sequences. According to Google Scholar citation metrics, the HiDe-NeRF~\cite{li2023oneshothighfidelitytalkingheadsynthesis}, DFA-NeRF~\cite{yao2022dfanerfpersonalizedtalkinghead}, and Neural Talking-Head~\cite{wang2021oneshotfreeviewneuraltalkinghead} models are important. DFA-NeRF~\cite{yao2022dfanerfpersonalizedtalkinghead} offers high fidelity in results but is time-consuming, while HiDe-NeRF~\cite{li2023oneshothighfidelitytalkingheadsynthesis} maintains identity even under significant deformations. Neural Talking-Head employs H.264 compression with unsupervised 3D keypoints for bandwidth-efficient video conferencing. Occlusion handling and performance dips on invisible identities are among the difficulties. Future directions include advanced training methods for improving realism and the development of temporal transformers.
\begin{figure}
  \centering
  \includegraphics[width=1\linewidth]{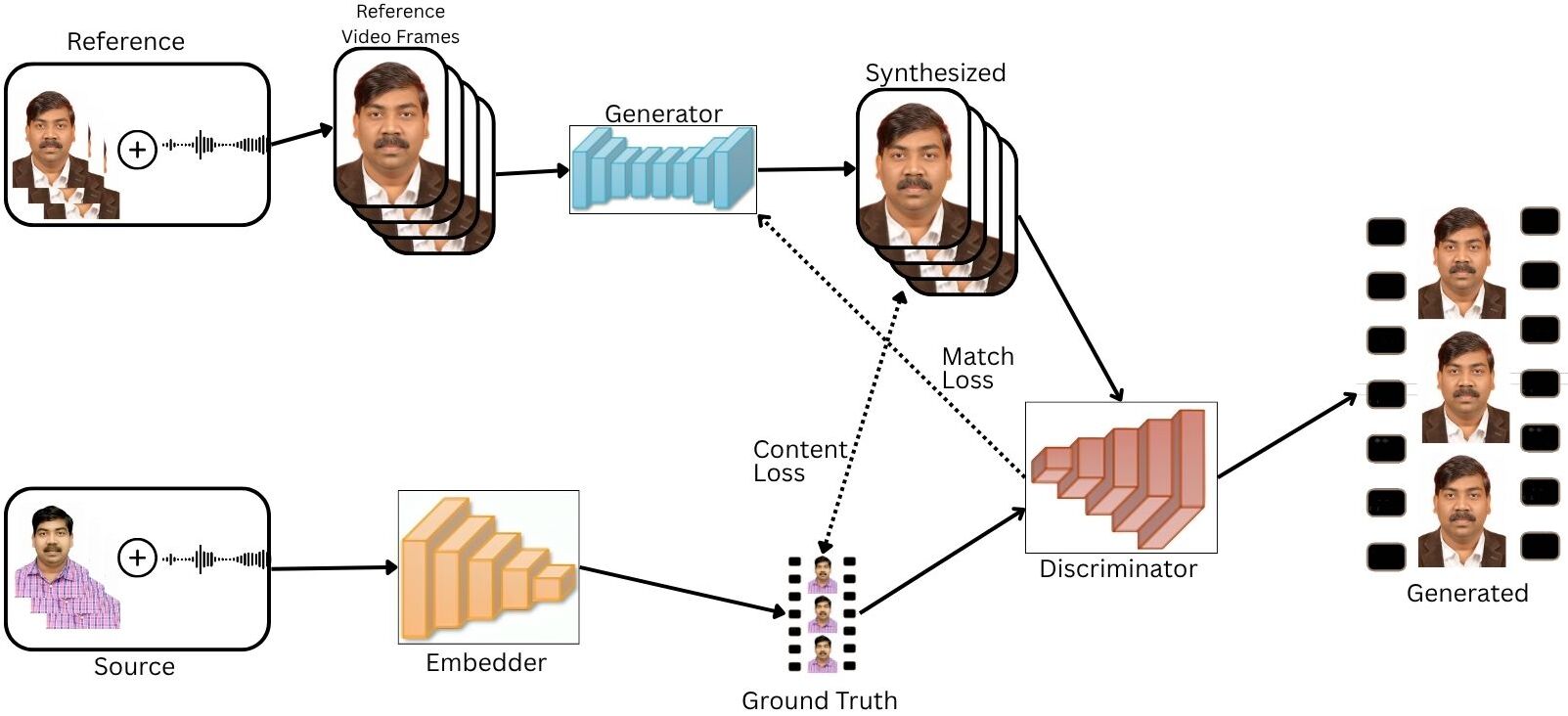}
  \caption{Generalized Approach for Video--based}
  \label{fig:2.3.1}
\end{figure}
A video--based talking head synthesis system producing a talking head from audio and reference posture data is shown in Figure \ref{fig:2.3.1}. Comprising two parts, the model is the Embedder, which gathers phonetic and prosodic data from the original audio, and the Generator, which produces a new sequence of video frames to simulate the speaker's speech and lip movements. It lets one perfectly reproduce the speaker's unique style. The Generator produces a new sequence of video frames that gives the impression that the person in the reference video speaks the words from the source audio. The Generator creates a new sequence of video frames that makes it look like the person in the reference video speaks the words from the source audio.  This method entails synchronizing lip motions and matching the pose from the reference frame. The Discriminator, an adversarial critic, takes two inputs: the synthesized video frames created by the Generator and the Ground Truth, which presumably relates to genuine video frames of the same person uttering certain audio and demonstrating a specific stance. The figure \ref{fig:2.3.1} highlights two types of losses that guide the training of the Generator: Content Loss, which measures the difference between the content (e.g., lip movements, facial expressions, and importantly, the pose) of the Synthesized video and the Ground Truth, and Match Loss, which is derived from the output of the Discriminator. This loss drives the Generator to generate video frames that are indistinguishable from genuine video frames in terms of visual quality and coherence. By challenging the Discriminator, the Generator learns to create increasingly photorealistic and natural-looking talking head sequences that align with the required position. The ultimate result of the trained model is the generated video, which is a series of video frames portraying the person from the Reference Video saying the content of the Source Audio, with synced lip movements and adopting the position from the given Reference position frame. The picture depicts a system where a pre--trained model combines a reference visual identity, target audio, and desired position to build a new talking head video. The training involves mapping audio information and desired postures to realistic facial movements, overall body posture, and appearance by minimizing Content Loss and Match Loss through adversarial learning.

\begin{sidewaystable}[htbp]
\scriptsize
\caption{Comparison of Video Based Approaches}
\label{tab:video_based_1}
\centering
\begin{tabular}{@{}lll>{\raggedright\arraybackslash}p{4.2cm}>{\raggedright\arraybackslash}p{4.2cm}l@{}}
\toprule
\textbf{Method} & \textbf{Arch.} & \textbf{Dataset} & \textbf{Highlights} & \textbf{Limitations} & \textbf{N-shot} \\
\midrule
DISCOHEAD~\cite{huang-2025} & ResNet-18 & Obama, GRID~\cite{10.1121/1.5042758}, KoEBA & Dense motion estimator with encoder for expressive mouth region synthesis. & Needs high-quality input; struggles with complex backgrounds. & Multi \\
HiDe-NeRF~\cite{li2023oneshothighfidelitytalkingheadsynthesis} & 3DMM & VoxCeleb1~\cite{nagrani-2019}, TH-1KH & Preserves identity under deformation for 3D face synthesis. & Occlusion challenges; DeepFake misuse risk. & Multi \\
DFA-NeRF~\cite{yao2022dfanerfpersonalizedtalkinghead} & NeRF & LRS2~\cite{Chung17}, HDTF~\cite{zhang-2021} & Long rendering time; diarization dependency. & Multi \\
MMVID~\cite{lin2023mmvidadvancingvideounderstanding} & GAN & MUG, iPER, VoxCeleb & Fuses visual modalities for diverse generation. & Motion consistency still limited. & Multi \\
Face-Dubbing++~\cite{waibel2022facedubbinglipsynchronousvoicepreserving} & GAN & LRS2~\cite{Chung17} & Multilingual dubbing with sync accuracy. & Prosody and ASR mismatch hurt realism. & Multi \\
Neural Talking-Head~\cite{wang2021oneshotfreeviewneuraltalkinghead} & GAN & VoxCeleb2~\cite{chung-20018}, TH-1KH & Bandwidth-efficient synthesis with unsupervised 3D keypoints. & Prone to artifacts under pose variation. & Multi \\
FT~\cite{zakharov2019fewshotadversariallearningrealistic} & VGG19 & VoxCeleb1~\cite{nagrani-2019}, VoxCeleb2~\cite{chung-20018} & Few-shot generation with meta-learning init. & Trade-off in realism and generalization. & Multi \\
RHM~\cite{chen-2020A} & GAN & CREMA-D~\cite{cheyneycomputerscience-no-date}, LRS3-TED~\cite{afouras-2018} & 3D-aware hybrid embedding for photo-realism. & Complex design, scalability issues. & Multi \\
vid2vid~\cite{wang2019fewshotvideotovideosynthesis} & GAN & YouTube, Street-scene & Few-shot with adaptive weights. & Fails with unseen CG styles. & Multi \\
Speech2Vid~\cite{chung2017saidthat} & CNN & VoxCeleb, LRW & Real-time video generation from speech + image. & Weakness with accents and wide expressions. & Multi \\
\bottomrule
\end{tabular}
\end{sidewaystable}

Several video--based THG techniques are evaluated in Table \ref{tab:video_based_1}, emphasizing multi-shot or few-shot learning paradigms. These methods utilize existing video data to create new talking head sequences, typically focusing on reenactment, dubbing, or generating fresh content. The examination analyzes architectural designs, datasets employed, important developments, and inherent limits of these video-driven systems, intending to make realistic and cohesive talking head films. DISCOHEAD~\cite{huang-2025} provides an innovative approach to creating realistic talking heads, emphasizing enhanced efficiency through a dense motion estimator and encoder. It focuses on managing the mouth area according to spoken sounds, ensuring proper lip synchronization. The architecture of DISCOHEAD~\cite{huang-2025} leverages a ResNet-18 backbone to extract visual features from input video frames, which are subsequently processed by the dense motion estimator and encoder to learn face motions, notably in the mouth region, driven by the audio. HiDe-NeRF~\cite{li2023oneshothighfidelitytalkingheadsynthesis} proposes a revolutionary technique for producing talking heads by exploiting Neural Radiance Fields (NeRF), allowing for realistic face distortion while scrupulously retaining the subject's identity. The method utilizes a 3D Morphable Model (3DMM) as a fundamental component, enabling new view synthesis and audio-driven facial deformations while prioritizing identity preservation. However, the research highlights several significant challenges associated with HiDe-NeRF~\cite{li2023oneshothighfidelitytalkingheadsynthesis}, including difficulties in handling face occlusions, pose bias prevalent in training datasets, and the potential misuse of such realistic creations for harmful purposes, such as "DeepFakes." DFA-NeRF~\cite{yao2022dfanerfpersonalizedtalkinghead} is a unique framework that leverages Neural Radiance Fields (NeRF) for high-fidelity, individualized talking head creation. It highlights the usefulness of deep neural networks, notably NeRFs, in creating realistic and identity-specific talking head synthesis. MMVID~\cite{lin2023mmvidadvancingvideounderstanding} is a multimodal video-generating framework meant to increase the quality, consistency, and variety of videos produced. Face-Dubbing++~\cite{waibel2022facedubbinglipsynchronousvoicepreserving} provides a powerful neural system for voice-preserving, lip-synchronous video translation, incorporating many unique neural network models. Neural Talking-Head~\cite{wang2021oneshotfreeviewneuraltalkinghead} is a video synthesis model designed for video conferencing applications. It aims to provide visual quality comparable to established video compression standards while using significantly less bandwidth. FT~\cite{zakharov2019fewshotadversariallearningrealistic} is a multi-shot system with few-shot learning capabilities that creates highly realistic images of human heads from just a few views of a person. RHM offers a 3D-aware generative network linked with a hybrid embedding and composition module, allowing the development of controlled, photorealistic, and temporally coherent talking-head films with natural head motions. The analyzed video--based THG methods employ diverse architectures such as ResNet-18, NeRF integrated with 3DMMs, specialized GANs with attention mechanisms and hybrid embeddings, and encoder-decoder CNNs, trained on datasets like Obama, VoxCeleb, LRS2, and in a few cases, self-collected video data. Finally, the Enhanced Temporal Representation and Spatial Alignment \cite{dong-2025} method addresses motion inconsistency in video-driven synthesis by introducing temporal representation augmentation (TRA) and spatial alignment correction (SAC). These innovations improve motion feature learning and head-pose consistency, producing smoother, artifact-free talking videos with superior visual fidelity.

\subsubsection{Text--Based THG}

\begin{figure}
  \centering
  \includegraphics[width=1\linewidth]{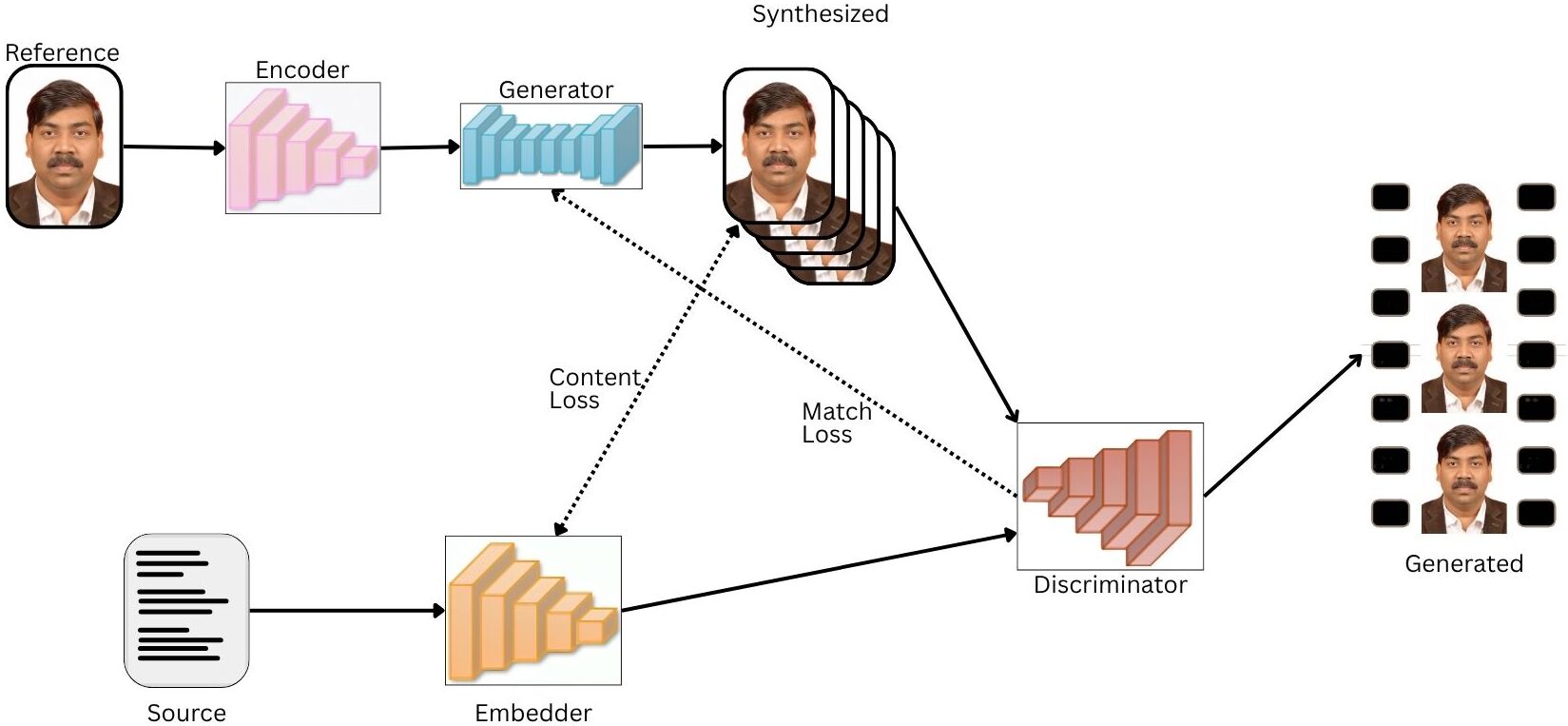}
  \caption{Generalized Approach for Text-based}
  \label{fig:2.1.4}
\end{figure}

Text-to-video models utilize emotion management and visual prediction modules to generate talking heads that respond to textual prompts. According to Google Scholar citation metrics, InstructAvatar~\cite{wang2024instructavatartextguidedemotionmotion}, TalkCLIP~\cite{ma2024talkcliptalkingheadgeneration}, and FT2TF~\cite{diao2024ft2tffirstpersonstatementtexttotalking} are important models. A video--based talking head synthesis system producing a talking head from audio and reference posture data is shown in Figure \ref{fig:2.1.4}. Comprising two parts, the model is the Embedder, which gathers phonetic and prosodic data from the original audio, and the Generator, which produces a new sequence of video frames to simulate the speaker's speech and lip movements. It lets one perfectly reproduce the speaker's unique style. The Generator produces a new sequence of video frames that gives the impression that the person in the reference video speaks the words from the source audio.
A theoretical framework for text--based talking head synthesis is shown in Figure \ref{fig:2.1.4}. This framework employs a pre--trained model to generate a new talking head video using a target text and a reference visual identity. The input is the source text, which dictates the intended speech and lip motions, and the reference picture gives the visual identity of the person who will be "talking." A video--based talking head synthesis system producing a talking head from audio and reference posture data is shown in Figure \ref{fig:2.1.4}. Comprising two parts, the model is the Embedder, which gathers phonetic and prosodic data from the original audio, and the Generator, which produces a new sequence of video frames to simulate the speaker's speech and lip movements. It lets one perfectly reproduce the speaker's unique style. The Generator produces a new sequence of video frames that gives the impression that the person in the reference video speaks the words from the source audio. The Discriminator serves as an antagonistic critic, accepting two inputs: the synthetic video frames created by the Generator and the Ground Truth, which presumably corresponds to genuine video frames of the same person saying the same or comparable material. The loss calculation and training (implicitly) entail two losses: Content Loss and Match Loss. Content Loss evaluates the difference between the content (e.g., lip movements, facial emotions) of the synthesized video and the Ground Truth, guided by the source text. The GeneratorGenerator aims to minimize Content loss to ensure that the generated lip movements and emotions align with the provided text. Match Loss is obtained from the output of the Discriminator, asking the Generator to produce video frames that are visually identical to real video frames in terms of coherence and quality. As illustrated in Figure \ref{fig:2.1.4}, the video--based talking head synthesis system generates a talking head by fusing reference posture information with audio. To ensure a flawless replication of the speaker's distinct style, the model comprises an Embedder that collects phonetic and prosodic information from the original audio and a Generator that generates a new video frame sequence to mimic the speaker's speech and lip movements.

\begin{sidewaystable}[htbp]
\scriptsize
\caption{Comparison of Text Based Approaches}
\label{tab:text_based_1}
\centering
\begin{tabular}{@{}lll>{\raggedright\arraybackslash}p{4.2cm}>{\raggedright\arraybackslash}p{4.2cm}l@{}}
\toprule
\textbf{Method} & \textbf{Arch.} & \textbf{Dataset} & \textbf{Highlights} & \textbf{Limitations} & \textbf{N-shot} \\
\midrule
InstructAvatar~\cite{wang2024instructavatartextguidedemotionmotion} & VAE & MEAD~\cite{kaisiyuan2020mead} & Text-guided expressive 2D avatars with improved interactivity. & emotion granularity and text precision need work. & One \\
EMO~\cite{tian2024emoemoteportraitalive} & VAE & CREMA & Realism via dynamic audio-facial alignment. & Data quality, expression complexity. & One \\
TalkCLIP~\cite{ma2024talkcliptalkingheadgeneration} & T2SS & MEAD~\cite{kaisiyuan2020mead}, HDTF~\cite{zhang-2021}, VoxCeleb2~\cite{chung-20018} & Text-driven synthesis via CLIP modulation. & Abstract expressions, mouth artifacts. & One \\
FT2TF~\cite{diao2024ft2tffirstpersonstatementtexttotalking} & GAN & LRS2~\cite{Chung17}, LRS3 & Text + vision fusion yields SOTA talking faces. & High training cost, expressiveness gaps. & Multi \\
Difftalk~\cite{shen-2023} & GAN & HDTF~\cite{zhang-2021} & Denoising diffusion for coherent motion. & Relies on reference image, heavy compute. & Multi \\
Face-Dubbing++~\cite{waibel2022facedubbinglipsynchronousvoicepreserving} & LSTM & LRS2~\cite{Chung17}, TED, etc. & Voice-preserving dubbing with sync accuracy. & ASR + prosody mismatches, fragility. & Multi \\
GC-AVT~\cite{9878472} & GAN & VoxCeleb2~\cite{chung-20018}, MEAD~\cite{kaisiyuan2020mead} & Fine-grained pose + lip + expression control. & Limited realism from background masking. & Multi \\
Text2Video~\cite{zhang2022text2videotextdriventalkingheadvideo} & GAN & VidTIMIT & Phoneme-pose mapping for text-driven synthesis. & Lacks realism, generalization, speaker control. & Multi \\
Write-a-Speaker~\cite{li2021writeaspeakertextbasedemotionalrhythmic} & 3DMM & Mocap & Text to head animation with rhythm and emotion control. & Motion capture needed; performance cost. & Multi \\
Flow Guided\cite{zhang-2021} & 3DMM & HDTF & Flow-guided video from audio-text inputs. & Poor coherence, style inconsistency. & One \\
Audio2Head~\cite{wang2021audio2headaudiodrivenoneshottalkinghead} & RNN & LRW, GRID~\cite{10.1121/1.5042758}, VoxCeleb & Audio-driven with pose + motion field prediction. & Misuse potential, identity mismatch. & One \\
Speech2Vid~\cite{chung2017saidthat} & CNN & VoxCeleb, LRW & Real-time generation from audio + image. & Accent handling, realism issues. & One \\
LMGG~\cite{chen2018lipmovementsgenerationglance} & CNN & GRID~\cite{10.1121/1.5042758}, LDC, LRW & Cross-modal synced lip movement generation. & Weak realism + efficiency. & Multi \\
Facial Reenactment~\cite{jalalifar2018speechdrivenfacialreenactmentusing} & GAN & CelebA & RNN + cGAN for expressive face sync. & Still lacks realism, consistency. & Multi \\
ObamaNet~\cite{kumar2017obamanetphotorealisticlipsynctext} & LSTM & Char2Wav, Pix2Pix & Full pipeline: text to lip-sync video. & Language support + ethics safeguards needed. & Multi \\
\bottomrule
\end{tabular}
\end{sidewaystable}

Various text--based talking head-generating strategies are assessed in Table \ref{tab:text_based_1}, focusing on multi-shot or few-shot learning paradigms. These techniques use pre-existing video data to produce new talking head sequences, often concentrating on tasks such as reenactment, dubbing, or developing original material. The research examines architectural designs, employed datasets, significant developments, and inherent constraints of video-driven systems to produce realistic and cohesive talking head films. DISCOHEAD~\cite{huang-2025} introduces an innovative method for creating realistic talking heads, emphasizing enhanced efficiency through a dense motion estimator and encoder. This approach focuses on managing the mouth area by spoken sounds, ensuring proper lip synchronization. DISCOHEAD~\cite{huang-2025}'s architecture employs a ResNet-18 backbone to extract visual features from input video frames, which are then processed by a dense motion estimator and encoder to learn facial motions, specifically in the mouth region, influenced by the audio. HiDe-NeRF~\cite{li2023oneshothighfidelitytalkingheadsynthesis} introduces an innovative method for generating talking heads using Neural Radiance Fields (NeRF), enabling realistic face deformation while rigorously maintaining the individual's identity. The approach utilizes a 3D Morphable Model (3DMM) as a core element, allowing innovative view synthesis and audio-driven face deformations while prioritizing identity preservation. The article highlights several significant issues with HiDe-NeRF~\cite{li2023oneshothighfidelitytalkingheadsynthesis}, including difficulties in managing face occlusions, bias in training datasets related to poses, and the potential for misuse in creating "DeepFakes." DFA-NeRF~\cite{yao2022dfanerfpersonalizedtalkinghead} is an innovative framework that utilizes Neural Radiance Fields (NeRF) to create high-fidelity, personalized talking heads. This framework demonstrates the effectiveness of deep neural networks, particularly NeRFs, in synthesizing realistic and identity-specific talking heads. MMVID~\cite{lin2023mmvidadvancingvideounderstanding} is a multimodal video generation framework designed to enhance the quality, consistency, and variety of the videos produced. Face-Dubbing++~\cite{waibel2022facedubbinglipsynchronousvoicepreserving} provides a powerful neural system for voice-preserving, lip-synchronous video translation, incorporating many unique neural network models. Developed for video conferencing applications, Neural Talking-Head~\cite{wang2021oneshotfreeviewneuraltalkinghead} is a neural talking-head video synthesis model that aims to use significantly less bandwidth while maintaining good visual quality comparable to established video compression standards. FT~\cite{zakharov2019fewshotadversariallearningrealistic} is a sophisticated multi-shot system that creates remarkably lifelike portraits of human heads using the power of few-shot learning. It creates vivid, immersive, realistic representations of people by capturing their essence and personality with only a few image perspectives. RHM offers a 3D-aware generative network linked with a hybrid embedding and composition module, allowing the development of controlled, photorealistic, and temporally coherent talking-head films with natural head motions. The reviewed methods for video--based THG utilize various architectures, including ResNet-18, NeRF combined with 3DMMs, specialized GANs equipped with attention mechanisms and hybrid embeddings, and encoder-decoder CNNs. These methods are trained on datasets such as Obama, VoxCeleb, LRS2~\cite{Chung17}, and self-collected video data.

Further advancements in text-conditioned generation have opened new pathways for controllable and expressive talking head synthesis. GenCA~\cite{sun2024gencatextconditionedgenerativemodel} introduces a powerful text-conditioned generative model that interprets textual inputs to guide fine-grained facial animation, offering a new level of semantic control in audiovisual generation tasks. T3M~\cite{peng2024t3mtextguided3d} extends this concept to 3D space, proposing a text-guided 3D avatar generation framework that produces high-quality and identity-consistent 3D heads from natural language prompts, bridging the gap between text understanding and 3D modeling. STAR~\cite{chai2024starskeletonawaretextbased4d} advances toward full 4D synthesis by introducing a skeleton-aware, text-driven system that generates temporally coherent, expressive facial performances aligned with the described motion, making it suitable for dynamic human avatar applications. Complementing these approaches, Text-to-Video~\cite{wang2023texttovideotwostageframeworkzeroshot} proposes a two-stage zero-shot framework that maps text directly to video sequences, enabling general and flexible video generation, which can include talking heads as a subset of its broader generative scope.

\subsubsection{2D--Based THG}

2D methods utilize landmark-driven warping and attention techniques, prioritizing computational efficiency. Style transmit, MetaPortrait~\cite{zhang2023metaportraitidentitypreservingtalkinghead}, and MakeItTalk~\cite{10.1145/3414685.3417774} are important models based on citation metrics from Google Scholar. These models transmit speaking styles across identities, although they have drawbacks such as artifacts in non-frontal perspectives. 2D warping struggles to accurately illustrate 3D head rotations and often loses high-frequency information, leading to depth ambiguity and issues with photorealism. Future advancements include diffusion--based detail enhancement and hybrid 2D-3D pipelines to improve efficiency and realism.

\begin{sidewaystable}[htbp]
\scriptsize
\caption{Comparison of 2D Based Approaches}
\label{tab:twod_based_1}
\centering
\begin{tabular}{@{}lll>{\raggedright\arraybackslash}p{4.2cm}>{\raggedright\arraybackslash}p{4.2cm}l@{}}
\toprule
\textbf{Method} & \textbf{Arch.} & \textbf{Dataset} & \textbf{Highlights} & \textbf{Limitations} & \textbf{N-shot} \\
\midrule
Style Transfer~\cite{10678562} & ResNet50 & VoxCeleb2~\cite{chung-20018}, RAVDESS~\cite{livingstone-2018} & 2D animation using audio and style reference from one image. & Needs style frames; heavy computation. & Multi \\
Diffused Heads~\cite{stypułkowski2023diffusedheadsdiffusionmodels} & GAN & LRW, CREMA & Autoregressive diffusion for smooth expressive motion. & Too slow for real-time. & Multi \\
MetaPortrait~\cite{zhang2023metaportraitidentitypreservingtalkinghead}& 3DMM & VoxCeleb2~\cite{chung-20018}, HDTF~\cite{zhang-2021} & Identity preservation using dense landmarks and generative priors. & Blurring under occlusion; alpha-blend needed. & Multi \\
LSP~\cite{lu2021livespeechportraitsrealtime} & LSTM & — & Real-time facial animation from live speech. & Affected by lighting, emotion, tracking. & Multi \\
PC-AVS~\cite{zhou2021posecontrollabletalkingfacegeneration} & GAN & VoxCeleb2~\cite{chung-20018}, LRW & Pose control + lip-sync from raw face image input. & Training bias; not fully real-time. & One \\
MakeItTalk~\cite{10.1145/3414685.3417774}& AUTOVC & VoxCeleb2~\cite{chung-20018} & Expression-aware synthesis from one image + audio. & Sparse landmarks may distort; lacks 3D cues. & One \\
Video Rewrite~\cite{10.1145/258734.258880} & Beier-Neely & HMM, TIMIT & Classic method using phoneme-aligned mouth patches. & Weak for expressive motion; limited matching. & One \\
\bottomrule
\end{tabular}
\end{sidewaystable}

A variety of 2D model--based algorithms for generating talking heads have been explored in Table \ref{tab:twod_based_1}. These algorithms aim to synthesize realistic facial movements using a single source image or a limited set of pictures guided by audio or other control inputs. Techniques like style transfer, diffusion models, and landmark manipulation are routinely applied to generate credible outcomes. Style Transfer~\cite{10678562} is an innovative method for creating audio-driven talking head animations. Drawing on learned style references, it generates 2D animations using a single input image and an audio stream. The technique utilizes a ResNet50 backbone for visual feature extraction and is evaluated on the VoxCeleb2~\cite{chung-20018} and RAVDESS~\cite{livingstone-2018} datasets. Diffused Heads~\cite{stypułkowski2023diffusedheadsdiffusionmodels} is an autoregressive diffusion model for making realistic talking head movies to attain state-of-the-art outcomes in expressiveness and smoothness. It leverages a GAN architecture inside an autoregressive diffusion framework and is assessed on the LRW and CREMA datasets. MetaPortrait~\cite{zhang2023metaportraitidentitypreservingtalkinghead} is a unique framework for identity-preserving one-shot talking head synthesis, combining dense facial landmarks, meta-learning approaches, 3D convolution for temporal modeling, and a generative prior. It is examined using VoxCeleb2~\cite{chung-20018} and HDTF~\cite{zhang-2021} datasets, indicating its capacity to manage various identities and create high-definition talking head movies from a single picture. LSP~\cite{lu2021livespeechportraitsrealtime} is a multi-shot approach that employs a single input picture to build a talking head from a single reference image. LSP~\cite{lu2021livespeechportraitsrealtime} is a live system that generates individualized talking-head animation using deep neural networks focused on capturing specific face dynamics and head movements for high-fidelity outcomes. It is trained on photorealistic photos and has been tested via user research. Key characteristics of LSP~\cite{lu2021livespeechportraitsrealtime} include its capacity to build unique talking-head animations in a live system, obtain high-fidelity face details, and control overhead posture accurately. However, it has difficulties recording plosive consonants, high-speed speech, emotive audio, shadows and lighting reflections, and realistic motions beyond simple head movements. PC-AVS~\cite{zhou2021posecontrollabletalkingfacegeneration} is a multi-shot framework designed for creating posture-controllable talking faces using non-aligned raw facial photos and an implicit low-dimensional pose code. It utilizes a GAN architecture and has been evaluated on the VoxCeleb2~\cite{chung-20018} and LRW datasets. The model can handle multiple identities and deliver accurate lip synchronization based on audio while allowing for control over facial positioning. MAKEItTalk is a one-shot approach for making talking-head videos from a single-face photograph utilizing audio input. It employs AUTOVC, an autoencoder--based voice conversion paradigm adaptable for visual feature creation. The approach has drawbacks relating to sparse landmark representation for video output and may struggle with managing complicated facial emotions beyond simple lip movements. Video Rewrite~\cite{10.1145/258734.258880} is a facial animation technique that automatically generates new lip movements from a source video for movie dubbing. It utilizes a Beier-Neely warping technique for image modification and employs Hidden Markov Models (HMMs) for phoneme alignment and selection. However, this technique has limitations regarding the availability and accuracy of phoneme-aligned mouth images in the source video. Handling complex facial emotions beyond simple lip movements may also be difficult.

\subsubsection{3D-Based THG}

Morphable models are utilized in 3D techniques to distinguish between em, posture, and geometry. ADL~\cite{FANG2024103925} (Audio-Driven Lip Sync), PV3D~\cite{xu2023pv3d3dgenerativemodel}, and JambaTalk~\cite{jafari2024jambatalkspeechdriven3dtalking} are important models according to Google Scholar citation metrics.ADL~\cite{FANG2024103925} backs multilingual lip sync; JambaTalk~\cite{jafari2024jambatalkspeechdriven3dtalking} runs the speech-to-motion transformer. PV3D~\cite{xu2023pv3d3dgenerativemodel} dely uses a 3D-aware GAN that cleverly includes motion dynamics despite its difficulties in capturing long-term dynamics, allowing a more interesting and immersive representation of movement in three dimensions. Rising computer expenses and the tediousness of data collecting create major issues. Still, neural rendering and single-camera footage's unsupervised 3D reconstruction offer fascinating possibilities for more development.
\begin{figure}
  \centering
  \includegraphics[width=1\linewidth]{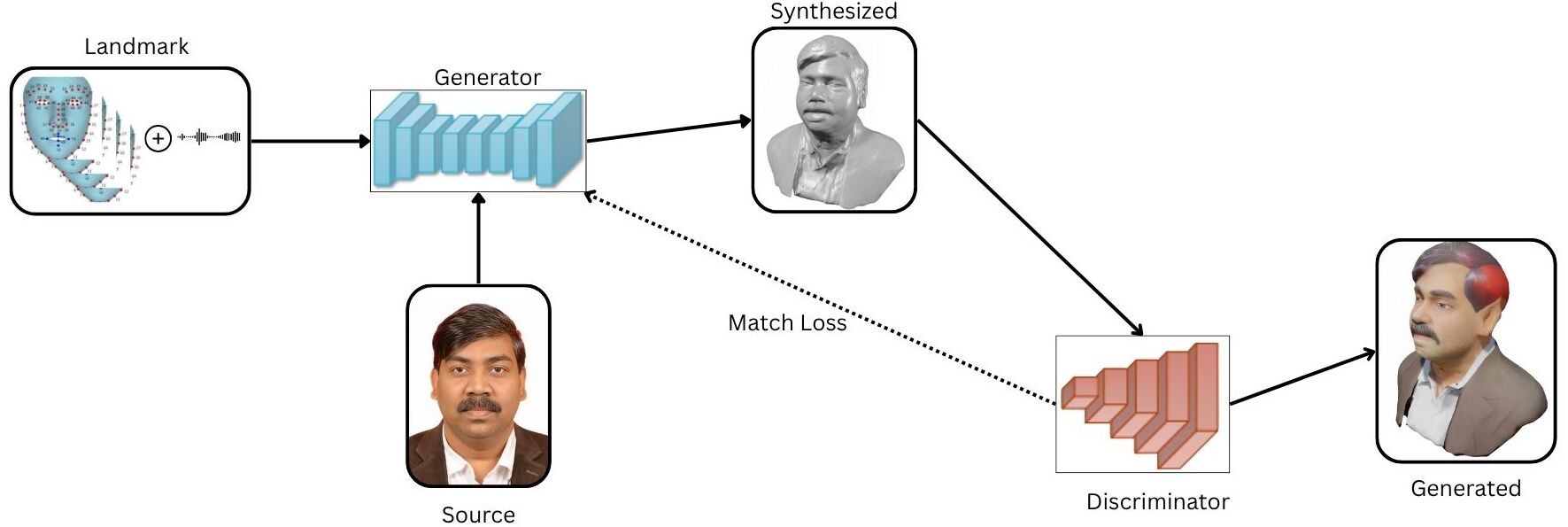}
  \caption{Generalized Approach for 3D-based}
  \label{fig:2.6.1}
\end{figure}
A theoretical architecture for synthesizing text--based talking heads is illustrated in Figure \ref{fig:2.6.1}. This architecture includes two inputs: the source image and the landmark sequence. The source image provides the visual identity and starting position of the individual whose 3D talking head is being created. Meanwhile, the landmark sequence conveys the appropriate facial expressions and movements. The model's core, the Generator, takes two inputs: the original picture and the landmark sequence. The procedure calls for animating a 3D model of a talking head that resembles the person in the original image according to a given sequence of landmarks. From this 3D model, 2D video frames may be produced. The Discriminator acts as an adversarial critic that receives two inputs: the synthetic 3D model and a realistic 3D talking head. Its role is to differentiate between the rendered output and the genuine 3D models, producing a score that reflects its confidence in the "realness" of the input. This adversarial setup is a critical feature of GAN, which is widely utilized for producing realistic 3D material. One sort of loss leads to the training of the Generator: Match Loss, generated from the output of the Discriminator. By attempting to "fool" the Discriminator, the Generator learns to build increasingly photorealistic and natural-looking 3D models that accurately animate according to the input landmarks. Other implicit losses may be linked to the precision of the produced 3D geometry and texture relative to the source picture and the fidelity of the animation to the input landmarks. The 3D--based THG model uses a pre--trained framework that includes a Generator and a Discriminator to create realistic and articulate 3D head models.

\begin{sidewaystable}[htbp]
\scriptsize
\caption{Comparison of 3D-Based Approaches}
\label{tab:threed-based_1}
\centering
\begin{tabular}{@{}lll>{\raggedright\arraybackslash}p{4.2cm}>{\raggedright\arraybackslash}p{4.2cm}l@{}}
\toprule
\textbf{Method} & \textbf{Arch.} & \textbf{Dataset} & \textbf{Highlights} & \textbf{Limitations} & \textbf{N-shot} \\
\midrule
PACT~\cite{HUANG2025100698} & 3D U-Net & Self-Data & Overcomes skull distortion via deep reconstruction. & Sensitive to skull shape variance. & One \\
AniArtAvatar~\cite{LI2025129706} & 3DMM & Wonder3D, NeuS & Artistic avatar generation with body + face motion. & Style deformation and inconsistency. & One \\
JambaTalk~\cite{jafari2024jambatalkspeechdriven3dtalking} & Mamba & VOCA~\cite{williams-no-date}, FaceFormer & Two-stage modeling for speech + laughter. & emotion/gesture modeling missing. & Multi \\
ADL~\cite{FANG2024103925} & OM-Net & VOCA~\cite{williams-no-date}SET-lip, ADLSET & Language-agnostic lip sync with Wav2Vec2.0. & Ignores upper-face emotion dynamics. & Multi \\
FHP~\cite{info:doi/10.2196/55476} & FFNN & StateFarm, Pose & Posture detection in real-time use-cases. & Poor cross-domain generalization. & One \\
TTSF~\cite{jang2024facesspeakjointlysynthesising} & 3DMM & LRS3, VoxCeleb2~\cite{chung-20018} & TTS fused with facial generation for avatars. & Voice consistency and realism gaps. & One \\
LaughTalk~\cite{sung-bin-2023} & 3DMM & VoxCeleb2~\cite{chung-20018}, CelebV-HQ~\cite{zhu2022celebvhqlargescalevideofacial} & Expressive generation for laughter + speech. & Weakness under emotional extremes. & Multi \\
AE-NeRF~\cite{li2023aenerfaudioenhancedneural} & 3DMM & NeRF, HDTF~\cite{zhang-2021} & Audio-enhanced NeRF for sync + fidelity. & Needs better speed benchmarking. & One \\
PV3D~\cite{xu2023pv3d3dgenerativemodel} & GAN & VoxCeleb, CelebV-HQ~\cite{zhu2022celebvhqlargescalevideofacial} & 3D-consistent motion with editable views. & Needs more diverse training data. & Multi \\
DVP~\cite{kim2018deepvideoportraits} & cGAN & Leader Dataset & Transfers expression, pose, blink for reenactment. & Breaks under extreme motion. & Multi \\
\bottomrule
\end{tabular}
\end{sidewaystable}

Table \ref{tab:threed-based_1} presents various 3D model--based methods for generating talking heads. These methods utilize 3D models and specific architectures to address particular challenges in this field. By directly modeling the 3D structure of the face, these techniques aim to enhance control over facial expressions, head posture, and overall realism. One notable learning--based image reconstruction method that effectively overcomes the distortions caused by the skull, a significant challenge for transcranial applications, is PACT~\cite{HUANG2025100698} (Photoacoustic Computed Tomography). AniArtAvatar~\cite{LI2025129706} provides a method for creating 3D-aware art avatars, allowing users to control features such as head positions, shoulder movements, and facial expressions through manual animation. The method uses a pre--trained 2D diffusion model for texture synthesis, an SDF--based neural surface representation for accurate geometry, and a 3DMM for the underlying facial structure and motion. For 3D talking heads, JambaTalk~\cite{jafari2024jambatalkspeechdriven3dtalking} offers a two-phase training method to effectively convey speech and laughing signals. In addition to a new collection of laughing videos, the work integrates VOCA~\cite{williams-no-date}, FaceFormer, CodeTalker, FaceDiffuser~\cite{stan2023facediffuserspeechdriven3dfacial}, and ScanTalk databases. The perceived quality of face recognition systems may be limited by limitations such as the limited availability of training set patterns and audio-visual data. Future research should explore control mechanisms that utilize gaze, emotions, gestures, and facial movements, as these elements may change independently in the real world. This study proposes a 3D talking lip system designed to interact with unknown individuals and support multiple languages. The technology features a pipeline that employs Wav2Vec2.0 with a Transformer for audio feature extraction, Viseme Fixing for visual alignment, and techniques for mapping lip landmarks. Additionally, it includes automated data processing and a head-mounted device for capturing 3D lip motions. Most likely emphasizing multilingual lip motions, the project trains the system using the VOCA SET-lip~\cite{williams-no-date} and ADLSET datasets. The project includes Frequent Human Pose (FHP~\cite{info:doi/10.2196/55476}), which emphasizes 3D human posture identification for daily activities. Its precise 3D human posture identification is via a graph convolutional network. The paper advocates improving the identification of 3D human postures using artificial intelligence. It also points out some negatives, such as completely capturing the varied postural changes happening in real life and natural biases in the training datasets. TTSF~\cite{jang2024facesspeakjointlysynthesising} (Text-to-audio and Face) combines Talking Face Generation technology with text-to-speech (TTS) to produce real talking faces and synchronized audio outputs straight from text input. It addresses problems like ensuring consistency between the generated speech and facial motions and generating realistic head positions. LaughTalk~\cite{sung-bin-2023} is a 3D talking head model with FLAME~\cite{FLAME:SiggraphAsia2017} parameters to simulate speaking and laughing. The 2D laughing clips are used to train the model, which is presumably learning to relate speech and laughter audio characteristics to the 3D FLAME parameters that control head movements and facial emotions. LaughTalk~\cite{sung-bin-2023} outperforms earlier systems, which generally have trouble with non-speech facial expressions. However, it also faces challenges in effectively conveying speech and laughter, suggesting that further research may be necessary to improve the variety and sensitivity of laughter and its smooth integration with speech. The improved audio-driven talking head synthesis technique AE-NeRF~\cite{li2023aenerfaudioenhancedneural} aims to improve generalization ability, audio-lip synchronization, and picture quality, particularly in few-shot scenarios. It aims to produce NeRF--based talking heads with proper lip synchronization by combining a 3DMM with NeRF (Neural Radiance Fields). PV3D~\cite{xu2023pv3d3dgenerativemodel} is a generative framework for 3D-aware portrait films that uses discriminators for more realism and explicitly models motion dynamics, including camera condition techniques, to maximize performance in static animation and view-consistent motion editing applications. A generative neural network is used in the ground-breaking technology known as Deep Video Portraits (DVP~\cite{kim2018deepvideoportraits}) to create realistic reanimations of portrait videos. It realistically transfers 3D head position, rotation, facial emotions, and eye blinking from a source to a target actor. However, DVP~\cite{kim2018deepvideoportraits} faces challenges such as managing excessive head motions or occlusions, relying on artificial face models for motion transfer, effectively transferring non-facial aspects, and possibly having limited generalization across various actor types. However, DVP~\cite{kim2018deepvideoportraits} faces challenges such as managing excessive head motions or occlusions, relying on artificial face models for motion transfer, effectively transferring non-facial aspects, and possibly having limited generalization across various actor types. The part-based local-global conditional GAN approach uses photometric FLAME reconstructions and 3D feature points to enable expression transfer across extreme pose variations. Its modular design with Parts Generation Networks (PGNs) and Fusion Networks (PFN)~\cite{rashid-2023} demonstrates robustness to occlusions while preserving identity attributes. In summary, the techniques for generating talking heads based on 3D models show significant advancements in several areas. These improvements include overcoming physical distortions in imaging, creating controllable 3D art avatars, generating realistic speech and laughter, synthesizing 3D talking lips for various languages, analyzing and potentially utilizing 3D human pose information, developing comprehensive text-to-talking face systems, enhancing audio-driven synthesis using NeRF, producing view-consistent 3D portrait videos, and transferring photorealistic facial expressions.

\subsubsection{Parameter--Based Approaches}

Parametric techniques manipulate facial movements using action units (AUs) or blend shape coefficients, enabling precise adjustments. According to Google Scholar citation metrics, the models DiscoFaceGAN~\cite{deng2020disentangledcontrollablefaceimage} (Contrastive Learning for Disentangled Factors), FEP~\cite{sato-2017} (Facial Expression Parameters), and PIR~\cite{huang2023parametricimplicitfacerepresentation} (Parametric Implicit Representation) \cite{huang2023parametricimplicitfacerepresentation} are significant. These methods combine implicit detail and explicit control, but they have limitations like overfitting to training identities, degradation under extreme lighting, and high demand for training data. Future efforts could include universal parametric spaces for cross-dataset compatibility and neural network--based nonlinear blend shape models.

\begin{sidewaystable}[htbp]
\scriptsize
\caption{Comparison of Parameter--Based THG}
\label{tab:parameter_based}
\centering
\begin{tabular}{@{}lll>{\raggedright\arraybackslash}p{4.2cm}>{\raggedright\arraybackslash}p{4.2cm}l@{}}
\toprule
\textbf{Method} & \textbf{Arch.} & \textbf{Dataset} & \textbf{Highlights} & \textbf{Limitations} & \textbf{N-shot} \\
\midrule
PIR~\cite{huang2023parametricimplicitfacerepresentation} & 3DMM & HDTF~\cite{zhang-2021} & Combines parametric + implicit models with augmentation for quality reenactment. & High complexity, data dependence, ethical concerns. & One \\
AdaIN~\cite{huang2017arbitrarystyletransferrealtime} & STN & RAVDESS~\cite{livingstone-2018} & Temporal regularized style transfer for dubbing. & Low expression precision, privacy risks. & Multi \\
DiscoFaceGAN~\cite{deng2020disentangledcontrollablefaceimage} & 3DMM & FFHQ~\cite{nvlabs-no-date} & Imitative + contrastive learning with 3D priors. & Sensitive to lighting/pose; lacks gaze control. & Multi \\
FEP~\cite{sato-2017} & HMM & Self & HMM + DNN with facial embeddings for pixel sync. & Needs large training; overfitting risk. & One \\
\bottomrule
\end{tabular}
\end{sidewaystable}

A few 3D model--based methods for THG are summarized in Table \ref{tab:parameter_based}. These methods typically use parametric models of the human face, such as 3D Morphable Models (3DMMs), to represent and modify facial geometry and texture. By adjusting the parameters of these models depending on audio or other input, they attempt to make realistic and controlled talking head animations. The architectural designs, datasets used, significant advancements, and intrinsic limitations of these parameter-driven approaches are all examined in this review. A few 3D model--based methods for THG are summarized in Table \ref{tab:parameter_based}. These methods typically use parametric models of the human face, such as 3D Morphable Models (3DMMs), to represent and modify facial geometry and texture. By adjusting the parameters of these models depending on audio or other input, they attempt to make realistic and controlled talking head animations. The architectural designs, datasets used, significant advancements, and intrinsic limitations of these parameter-driven approaches are all examined in this review. Combining explicit and implicit neural representation techniques,  PIR~\cite{huang2023parametricimplicitfacerepresentation} (Parametric Implicit Representation) offers a ground-breaking framework for audio-driven face reenactment that produces excellent facial animation. It uses data augmentation methods, conditional picture synthesis techniques, and contextual data to increase performance and resilience.  PIR~\cite{huang2023parametricimplicitfacerepresentation} is tested using the HDTF~\cite{zhang-2021} and PC-AVS~\cite{zhou2021posecontrollabletalkingfacegeneration} datasets, commonly used for audio-driven THG and face reenactment applications. User-generated content (UGC) audio-driven dubbing employs AdaIN~\cite{huang2017arbitrarystyletransferrealtime} (Adaptive Instance Normalization). The technique presumably translates the speaking style (visual lip movements) from a source video to a target face using AdaIN~\cite{huang2017arbitrarystyletransferrealtime} inside a Style Translation Network (STN), depending on the audio of the dubbed clip. Temporal regularization and a semi-parametric video renderer produce realistic and fluid results. DiscoFaceGAN~\cite{deng2020disentangledcontrollablefaceimage} is an advanced method for synthesizing images that employs contrastive learning, imitative-contrastive learning, and 3D priors to generate controlled facial images while separating variations in facial features. It utilizes a GAN architecture conditioned on a specific parameter, 3DMM. When tested on the FFHQ~\cite{nvlabs-no-date} dataset, a high-quality collection of human faces, DiscoFaceGAN~\cite{deng2020disentangledcontrollablefaceimage} aims to produce diverse and lifelike facial images with carefully managed features. Facial Expression Parameters (PEP) suggests a two-step synthesis method to create realistic talking face animation from pixels. It uses Deep Neural Networks (DNNs), which incorporate facial expressions to achieve a stronger link between audio and visual context, and Hidden Markov Models (HMMs) for temporal modeling. FEP~\cite{sato-2017} uses HMMs to replicate the temporal sequence of phonemes or visemes derived from the audio input. The corresponding lip movements and facial expressions are then generated at the pixel level by DNNs, conditioned on facial expression parameters that presumably provide greater control over the final animation. The reviewed parameter--based THG methods utilize a variety of architectures, including 3D Morphable Models (3DMMs) with implicit neural representations, Style Translation Networks with Adaptive Instance Normalization (AdaIN~\cite{huang2017arbitrarystyletransferrealtime}), Generative Adversarial Networks (GANs) conditioned on 3D priors and contrastive learning, and Hidden Markov Models (HMMs) integrated with Deep Neural Networks (DNNs) and facial expression parameters. These methods have been evaluated using HDTF~\cite{zhang-2021}, PC-AVS~\cite{zhou2021posecontrollabletalkingfacegeneration}, RAVDESS~\cite{livingstone-2018}, FFHQ~\cite{nvlabs-no-date}, and self-collected video data. The study demonstrates significant progress in face image generation, automated dubbing, facial expression parameter integration, and high-quality reenactment. It also highlights drawbacks such as the reliance on parametric models, the inability to control lighting and poses, potential overfitting, and further improvements in facial expression realism and animation quality. Using AdaIN~\cite{huang2017arbitrarystyletransferrealtime} (Adaptive Instance Normalization) inside a Style Translation Network (STN), the technique translates speaking styles from a source video to a target face.  Temporal regularization and a semi-parametric video renderer produce realistic and fluid results. DiscoFaceGAN~\cite{deng2020disentangledcontrollablefaceimage} is an advanced method for synthesizing images that employs contrastive learning, imitative-contrastive learning, and 3D priors to generate controlled facial images while separating variations in facial features. It utilizes a GAN architecture conditioned on a specific parameter, 3DMM. When tested on the FFHQ~\cite{nvlabs-no-date} dataset, a high-quality collection of human faces, DiscoFaceGAN~\cite{deng2020disentangledcontrollablefaceimage} aims to produce diverse and lifelike facial images with carefully managed features. Facial Expression Parameters (PEP) suggests a two-step synthesis method to create realistic talking face animation from pixels. It uses Deep Neural Networks (DNNs), which incorporate facial expressions to achieve a stronger link between audio and visual context, and Hidden Markov Models (HMMs) for temporal modeling. FEP~\cite{sato-2017} uses HMMs to replicate the temporal sequence of phonemes or visemes derived from the audio input. The corresponding lip movements and facial expressions are then generated at the pixel level by DNNs, conditioned on facial expression parameters that presumably provide greater control over the final animation. The reviewed parameter--based THG methods utilize a variety of architectures, including 3D Morphable Models (3DMMs) with implicit neural representations, Style Translation Networks with Adaptive Instance Normalization (AdaIN~\cite{huang2017arbitrarystyletransferrealtime}), Generative Adversarial Networks (GANs) conditioned on 3D priors and contrastive learning, and Hidden Markov Models (HMMs) integrated with Deep Neural Networks (DNNs) and facial expression parameters. These methods have been evaluated using HDTF~\cite{zhang-2021}, PC-AVS~\cite{zhou2021posecontrollabletalkingfacegeneration}, RAVDESS~\cite{livingstone-2018}, FFHQ~\cite{nvlabs-no-date}, and self-collected video data. They demonstrate notable progress in several areas, including the transfer of speaking styles for automated dubbing, the creation of controllable and disentangled face images, the incorporation of facial expression parameters for improved audio-visual correspondence in pixel-level animation, and the combination of parametric and implicit representations for high-quality reenactment. The expressiveness of the underlying parametric models is often relied upon, extreme poses and lighting conditions are difficult to control, overfitting to training data is a possibility, and additional work is required to control and improve the realism of generated facial expressions and animation quality overall. While not strictly a THG method, the multi-feature fusion algorithm \cite{cheng-2023} demonstrates the value of parameterized facial analysis through its lightweight CNN architecture. Fusing eye, mouth, and head movement features with lane-departure data highlights the potential for hybrid parameter-driven systems in real-time applications requiring facial behavior understanding.

\subsubsection{NeRF--Based THG}

Neural Radiance Fields (NeRFs) enable photographic novel-view synthesis and dynamic facial deformations through volumetric scene representation. Google Scholar cites AD-c~\cite{guo2021adnerfaudiodrivenneural}, IMavatar\cite{zheng2022imavatarimplicit}, and CVTHead~\cite{ma2023cvtheadoneshotcontrollablehead} as the main models for generating avatars from single photos and detecting fine-scale wrinkles. However, real-time rendering, training stability, and slow inference times are major challenges. Future research will focus on creating portable Neural Radiance Fields substitutes and exploring innovative hash encoding techniques for enhanced efficiency and performance.

\begin{sidewaystable}[htbp]
\scriptsize
\caption{Comparison of NeRF--Based Approaches}
\label{tab:nerf_based_1}
\centering
\begin{tabular}{@{}lll>{\raggedright\arraybackslash}p{4.2cm}>{\raggedright\arraybackslash}p{4.2cm}l@{}}
\toprule
\textbf{Method} & \textbf{Arch.} & \textbf{Dataset} & \textbf{Highlights} & \textbf{Limitations} & \textbf{N-shot} \\
\midrule
CVTHead~\cite{ma2023cvtheadoneshotcontrollablehead} & 3DMM & VoxCeleb & One-shot neural avatar w/ expression + pose control. & Limited by DECA mesh reconstruction. & One \\
Head3D~\cite{ni20233dawaretalkingheadvideomotion} & ConvLSTM & VoxCeleb, FaceForensics & Canonical frames w/ interpretable 3D motion transfer. & Parsing errors, side-view issues. & Multi \\
SSP-NeRF\cite{liu2022semanticawareimplicitneuralaudiodriven} & 3DMM & NVP, Obama Synth & Semantic NeRF w/ dynamic sampling. & Rendering slow; poor language generalization. & Multi \\
HeadNeRF~\cite{ha2019marionettefewshotfacereenactment} & 3DMM & CelebAMask-HQ, FFHQ~\cite{nvlabs-no-date} & Fine-grain NeRF face model. & Fails on headgear, wild generalization. & Multi \\
IMavatar\cite{zheng2022imavatarimplicit} & 3DMM & VOCA~\cite{williams-no-date}, COMA & Monocular avatars w/ expression modeling. & Heavy compute, coarse high-frequency detail. & Multi \\
ROME~\cite{khakhulin2022realisticoneshotmeshbasedhead} & FLAME & VoxCeleb2~\cite{chung-20018}, H3DS & One-shot head reenactment with realism. & Lacks fine detail; smoothed geometry. & One \\
FNeVR~\cite{zeng2022fnevrneuralvolumerendering} & CNN & VoxCeleb1~\cite{nagrani-2019} & Volume rendering to fix motion deformation. & Compute-heavy; poor pose editability. & One \\
DFA-NeRF~\cite{yao2022dfanerfpersonalizedtalkinghead} & 3DMM & LRS2~\cite{Chung17}, HDTF~\cite{zhang-2021} & Audio-driven NeRF w/ lip-sync alignment. & Single speaker support; slow rendering. & Multi \\
NerFACE~\cite{gafni2020dynamicneuralradiancefields} & SRNs & Wild videos & 4D NeRF reenactment for telepresence. & Lacks eye/torso motion control. & Multi \\
AD-NeRF~\cite{guo2021adnerfaudiodrivenneural}& GAN & Wild videos & View- and audio-adaptive NeRF synthesis. & Language mismatch degrades realism. & Multi \\
\bottomrule
\end{tabular}
\end{sidewaystable}

Neural Radiance Fields (NeRFs) enable photographic novel-view synthesis and dynamic facial deformations through volumetric scene representation. Google Scholar cites AD-NeRF~\cite{guo2021adnerfaudiodrivenneural}, IMavatar\cite{zheng2022imavatarimplicit}, and CVTHead~\cite{ma2023cvtheadoneshotcontrollablehead} as the main models for generating avatars from single photos and detecting fine-scale wrinkles. However, real-time rendering, training stability, and slow inference times are major challenges. Future research will focus on creating portable Neural Radiance Fields substitutes and exploring innovative hash encoding techniques for enhanced efficiency and performance. Various methods for THG based on 3D models have been reviewed in Table \ref{tab:nerf_based_1}. These methods utilize Neural Radiance Fields (NeRF) or combine them with other techniques, such as 3D Morphable Models (3DMMs), to achieve photorealistic and controlled talking head synthesis. NeRF--based algorithms learn a volumetric model of the scene, allowing for innovative view synthesis and detailed rendering. CVTHead~\cite{ma2023cvtheadoneshotcontrollablehead} offers a novel method for creating programmable neural head avatars from a single reference image. It aims to maintain visual quality while quickly representing human heads in various positions and emotions. To achieve effective and controllable rendering of the head avatar with different expressions and poses derived from the 3DMM parameters, CVTHead~\cite{ma2023cvtheadoneshotcontrollablehead} uses a 3DMM as a foundational representation of head geometry and texture. It probably combines this with neural rendering techniques, perhaps inspired by NeRF. Head3D~\cite{ni20233dawaretalkingheadvideomotion} proposes a 3D-aware talking-head video motion transmission network. The motion from a driving video or audio is transferred to the target identity using this 3D model. One of Head3D~\cite{ni20233dawaretalkingheadvideomotion}'s most notable features is its ground-breaking 3D-aware Method for Transferring Motion in Talking Head Videos. This sophisticated method goes beyond previous methods and performs exceptionally well when identities change in the real world. It enhances the realism of animated faces and produces stunningly clear and visually interpretable 3D canonical head models, bringing virtual characters to life like never before. SSP-NeRFF~\cite{liu2022semanticawareimplicitneuralaudiodriven} introduces an innovative approach that utilizes Neural Radiance Fields to produce high-quality video portraits. We need to enhance our methods to improve how we create talking heads with NeRF technology. Integrating Dynamic Ray Sampling modules and incorporating semantic awareness is crucial, potentially through 3D Morphable Models (3DMMs) or other facial analysis techniques. A parametric head model based on NeRF, HeadNeRF, was created to improve the quality of high-fidelity headshots. To replicate the intricate appearance and possible non-rigid deformations of the face, a Neural Radiance Field combined with a 3DMM offers a parametric description of head geometry and posture. IM Avatar provides a generative neural network (GNN) based on NeRF to generate a variety of headgear. In contrast to standard 3D morphable face models, IMavatar\cite{zheng2022imavatarimplicit} is a method for learning implicit head avatars from monocular films that improve geometry and expression space by learning an implicit head representation. Monocular video sequences create an implicit neural representation of the head avatar that can be manipulated to generate a variety of positions and emotions. The VOCA~\cite{williams-no-date} and COMA datasets, which focus on voice-driven face animation and provide a rich space of facial emotions, respectively, are used to evaluate the method. Intending to achieve competitive head geometry recovery and high-quality rendering from a single image, ROME~\cite{khakhulin2022realisticoneshotmeshbasedhead} is a technique for creating realistic one-shot mesh--based human head avatars. It uses FLAME as a simple mesh and regresses the FLAME parameters from a single input image, most likely using deep learning techniques. VoxCeleb2~\cite{chung-20018} and H3DS (Human 3D Scans) are used to evaluate ROME~\cite{khakhulin2022realisticoneshotmeshbasedhead}, which emphasizes recovering precise 3D head geometry and providing realistic representations for a range of users. By overcoming motion deformation and complex modeling challenges, FNeVR~\cite{zeng2022fnevrneuralvolumerendering}, a CNN--based network, aims to improve computer vision tasks, particularly talking head creation. It uses neural volume rendering to improve performance on talking-head benchmarks. However, it faces limitations like high computational complexity, which can impact rendering speed and resource requirements, pose modification problems, and limit generalization to highly diverse and unknown input data. DFA-NeRF~\cite{yao2022dfanerfpersonalizedtalkinghead} is a novel neural radiance field framework designed for personalized THG to surpass existing methods in producing high-fidelity lip-synchronized talking heads. It is NeRF--based and learns a neural radiance field conditioned on audio and identification. LRS2~\cite{Chung17} and HDTF~\cite{zhang-2021} datasets, which are high-resolution talking head video datasets with synchronized audio, are used to evaluate the method. NeRFACE uses dynamic neural radiance fields to mimic human facial appearance and produce realistic talking head images that outperform video--based reenactment methods. This strategy greatly benefits applications in solving virtual reality (VR) or augmented reality (AR) telepresence. High-quality, two-minute self-gathered videos train the method, emphasizing learning from real-world data to attain high fidelity. ADNeural scene representation networks (perhaps a variation of NeRF) conditioned on audio input are used in the GAN--based NeRF architecture. It allows the generated talking heads' audio signals, viewing instructions, and background images to be altered without restriction. However, the article points out flaws like strange mouths and body parts in the generated movies, which are most likely the result of discrepancies between the driving language and the training data. A variety of architectures are used in the reviewed NeRF--based and hybrid THG methods, including CNNs for neural volume rendering, 3DMMs with neural rendering, Conv LSTM for 3D motion transfer, NeRFs with semantic awareness and dynamic ray sampling, parametric head models enhanced with NeRF, implicit neural representations, FLAME for one-shot avatar creation, and GANs conditioned on NeRF for audio-driven synthesis. Neural volume rendering for better animation, high-fidelity video portraits, 3D-aware motion transfer, controllable neural head avatars from single images, learning implicit head avatars with improved geometry and expression, and enabling flexible audio, viewpoint, and background adjustments are just a few of the areas where the techniques demonstrate advancements.

Furthermore, other NeRF-based models have significantly advanced the realism and controllability of 3D facial synthesis and talking head generation. 3DFaceShop~\cite{tang20223dfaceshopexplicitlycontrollable3daware} presents a system for explicitly controllable, 3D-aware facial generation using neural rendering techniques that enable intuitive editing through sparse sketch-based inputs. Next3D~\cite{sun2023next3dgenerativeneuraltexture} leverages generative neural textures within a 3D framework to create high-quality, animatable facial avatars with photorealistic fidelity, contributing toward fully interactive virtual humans. NeRFInvertor~\cite{yin2022nerfinvertorhighfidelitynerfgan} bridges GANs and NeRF by inverting 2D facial images into high-fidelity volumetric representations, enabling fine-grained editing and robust identity preservation. DFRF~\cite{shen2022learningdynamicfacialradiance} introduces a dynamic facial radiance field that captures appearance and expression dynamics, enhancing temporal coherence and expressiveness in facial animations. 

\subsubsection{Diffusion--Based THG}

Diffusion models, known for their temporal coherence and expression synthesis capabilities, are utilized to create high-quality films. According to Google Scholar citation metrics, MoDiTalker~\cite{kim2024moditalkermotiondisentangleddiffusionmodel}, DreamTalk~\cite{ma2024dreamtalkemotionaltalkinghead}, and DAE-Talker~\cite{Du_2023} are significant models, each with its limitations. DAE-Talker~\cite{Du_2023} has state-of-the-art lip-sync accuracy, DreamTalk~\cite{ma2024dreamtalkemotionaltalkinghead} generates strong emotions, and MoDiTalker~\cite{kim2024moditalkermotiondisentangleddiffusionmodel} differentiates between lip and non-lip motions. Challenges include fine-grained editing, high VRAM requirements, and computational cost. Future research could focus on reinforcement learning for reward-guided generation and latent consistency models.

\begin{sidewaystable}[htbp]
\scriptsize
\caption{Comparison of Diffusion--Based Approaches}
\label{tab:diffusion_based_2}
\centering
\begin{tabular}{@{}lll>{\raggedright\arraybackslash}p{4.2cm}>{\raggedright\arraybackslash}p{4.2cm}l@{}}
\toprule
\textbf{Method} & \textbf{Arch.} & \textbf{Dataset} & \textbf{Highlights} & \textbf{Limitations} & \textbf{N-shot} \\
\midrule
MoDiTalker~\cite{kim2024moditalkermotiondisentangleddiffusionmodel} & GAN & LRS3-TED~\cite{afouras-2018}, HDTF~\cite{zhang-2021} & Motion-disentangled diffusion for subtle lip movements. & High sampling time, stochasticity, ethics concerns. & Multi \\
EMOTALKER\cite{zhang2024emotalkeremotionallyeditabletalking} & CLIP & MEAD~\cite{kaisiyuan2020mead}, CREMA-D~\cite{cheyneycomputerscience-no-date} & emotion intensity control in speech-driven avatars. & Generalization issues, realism-emotion trade-offs. & One \\
DiffTalk\cite{shen-2023} & RNN & HDTF~\cite{zhang-2021} & Audio-driven diffusion with temporal coherence. & Requires face reference image, slow, artifacts. & Multi \\
Diffused Heads~\cite{stypułkowski2023diffusedheadsdiffusionmodels} & GAN & CREMA, LRW & Autoregressive diffusion for smooth expressions. & Slow generation; not real-time suitable. & One \\
Diffusion Video~\cite{bigioi2023speechdrivenvideoediting} & U-Net & GRID~\cite{10.1121/1.5042758}, CREMA-D~\cite{cheyneycomputerscience-no-date} & Denoising diffusion for multi-speaker video edits. & Long training, lip sync issues. & One \\
DAE-Talker~\cite{Du_2023} & DAE & LibriTTS & Latent DAE boosts lip sync and pose accuracy. & Poor pose modeling; resource intensive. & One \\
DreamTalk~\cite{ma2024dreamtalkemotionaltalkinghead} & 3DMM & MEAD~\cite{kaisiyuan2020mead}, HDTF~\cite{zhang-2021} & Style-aware emotional diffusion model. & Lacks realism for subtle emotional cues. & Multi \\
DiT-Head~\cite{mir2023ditheadhighresolutiontalkinghead} & DiT & HDTF~\cite{zhang-2021} & High-fidelity transformer--based diffusion. & English-only; deepfake + compute risks. & Multi \\
FaceDiffuser~\cite{stan2023facediffuserspeechdriven3dfacial} & HuBERT & BIWI, Multiface, BEAT & 3D animation via nondeterministic diffusion. & Dataset limits, slow inference, subjective evals. & Multi \\
PAVDP~\cite{yu2022talkingheadgenerationprobabilistic} & 3DMM & VoxCeleb1~\cite{nagrani-2019} & Probabilistic priors for facial motion. & Stripe artifacts; weak high-freq detail. & One \\
\bottomrule
\end{tabular}
\end{sidewaystable}

A range of diffusion model--based strategies in THG has been examined in A range of diffusion model--based strategies in THG has been examined in Table \ref{tab:diffusion_based_2} that leverage diffusion models, a class of generative models notable for their capacity to create high-quality and varied outputs by learning to reverse a slow noising process. These methods aim to improve artificial talking head movies' temporal coherence, expressiveness, and realism. MoDiTalker~\cite{kim2024moditalkermotiondisentangleddiffusionmodel} is a diffusion model specifically designed to enhance the generation of talking heads. It separates motion into audio-driven lip movements and more general facial expressions, using distinct modules for audio-to-motion and motion-to-video conversion. The model aims to create realistic and synchronized talking heads from audio by employing a GAN architecture within a diffusion framework. The portrait animation technique EMOTalker aims to improve emotion recognition when producing talking heads. While enabling the creation of customizable facial emotions through an emotion Intensity Block, it attempts to preserve the original photo identity. The model uses an architecture based on Contrastive Language-Image Pre-training (CLIP) to understand and produce facial expressions associated with different emotions, potentially enabling text--based emotion management.DiffTalk~\cite{shen-2023} proposes a talking head synthesis method that generates high-quality visuals in sync with the original audio through audio-driven, temporally coherent denoising. Using a Recurrent Neural Network (RNN) architecture, it learns to predict the corresponding facial movements over time by capturing the temporal relationships in the audio. The technique creates a single talking head using a single image as input in the multi-shot Diffused Heads~\cite{stypułkowski2023diffusedheadsdiffusionmodels} technique. Its dependence on reference face photos, the computational expense of repeated denoising techniques, and the possibility of artifacts in the output videos are some of its drawbacks. It seeks to enhance the entire synthesis process by utilizing diffusion models' capacity to synthesize high-quality video frames sequentially. Diffused Heads~\cite{stypułkowski2023diffusedheadsdiffusionmodels}, an autoregressive diffusion model, aims to create realistic talking head movies. Using a GAN architecture within an autoregressive diffusion framework, the model generates successive video frames, each dependent on the audio input and the frames that came before it. The CREMA and LRW datasets, which focus on emotional expressions and extensive lip reading datasets, are used to evaluate it. Diffusion Video Editing~\cite{bigioi2023speechdrivenvideoediting} is an innovative speech-driven method. This technique showcases the ability of diffusion models to generate new content and modify existing films based on audio input. It has potential applications in tasks like lip synchronization and adjusting facial expressions to match new audio. The method integrates a U-Net architecture inside a denoising diffusion model to enable targeted changes to the video content based on the audio, presumably learning to denoise video frames conditioned on the input speech. DAE-Talker~\cite{Du_2023} uses latent representations learned by a Denoising Autoencoder (DAE) to improve speech-driven talking face synthesis. By utilizing powerful feature representations that the DAE has learned from audio, it aims to outperform earlier methods regarding lip synchronization accuracy, video quality, pose naturalness, and pose controllability. Pose accuracy and control, efficient pose modeling, denoising performance (within the DAE), generalization to unseen identities or speaking styles, the need for improved evaluation metrics for THG, the DAE's data dependency, and the system's overall computational complexity are some of the issues that the paper recognizes DAE-Talker~\cite{Du_2023} faces. DreamTalk~\cite{ma2024dreamtalkemotionaltalkinghead} is a diffusion model framework designed to create realistic talking heads exhibiting various emotions. Controlling the emotional style of the generated talking head consists of a style predictor, a style-aware lip expert, and a denoising network. The method addresses problems such as emotion recognition in low-emotion-intensity audio, speaker identity retention, and consistent expressions. DiT-Head~\cite{mir2023ditheadhighresolutiontalkinghead} is a novel talking head synthesis pipeline that uses Diffusion Transformers (DiT) with audio input for scalability and competitive performance. It has demonstrated promise for many applications and is evaluated using the HDTF~\cite{zhang-2021} dataset, a common benchmark for talking head synthesis. However, its reliance on data from English speakers and the substantial computer resources required for training and inference limit it. A non-deterministic deep learning model, FaceDiffuser~\cite{stan2023facediffuserspeechdriven3dfacial}, was developed for speech-driven 3D face animation synthesis. It uses an internal dataset in addition to 3D vertex and blend shape datasets to learn realistic 3D facial motions directly from speech. The BIWI, Multiface, BEAT, and UUDaMM datasets are used to train the model, which allows it to learn realistic speech-driven 3D facial deformations. With a focus on probabilistically sampling holistic lip-irrelevant facial gestures to match the input audio while maintaining photorealism and naturalness, PAVDP~\cite{yu2022talkingheadgenerationprobabilistic} offers a novel framework for one-shot audio-driven talking head creation. It uses a generator network (G) that was trained similarly to PC-AVS~\cite{zhou2021posecontrollabletalkingfacegeneration} but focuses on using probabilistic diffusion prior models to predict non-lip facial movements. A variety of architectures are used in the reviewed diffusion--based THG techniques, including GANs in diffusion frameworks, CLIP--based models, RNNs with denoising, Diffusion Transformers, U-Nets for video editing, Denoising Autoencoders, and specialized diffusion models with style-aware elements. These methods focus on advances in speech-driven video editing, expressiveness, smoothness, emotionally editable talking heads, motion disentanglement, pose control, lip synchronization, and high-quality synthesis. Some main disadvantages include long sampling times, high computing costs, the potential for artifacts, challenges with generalization and identity retention, reliance on large and diverse datasets, and ethical issues related to creating realistic synthetic media.Table \ref{tab:diffusion_based_2} that leverages diffusion models, a class of generative models notable for their capacity to create high-quality and varied outputs by learning to reverse a slow noising process. These methods aim to improve artificial talking head movies' temporal coherence, expressiveness, and realism. MoDiTalker~\cite{kim2024moditalkermotiondisentangleddiffusionmodel} is a diffusion model specifically designed to enhance the generation of talking heads. It separates motion into audio-driven lip movements and more general facial expressions, using distinct modules for audio-to-motion and motion-to-video conversion. The model aims to create realistic and synchronized talking heads from audio by employing a GAN architecture within a diffusion framework. The portrait animation technique emoTalker aims to improve Emotion recognition when producing talking heads. While enabling the creation of customizable facial Emotions through an Emotion Intensity Block, it attempts to preserve the original photo identity. The model uses an architecture based on Contrastive Language-Image Pre-training (CLIP) to understand and produce facial expressions associated with different Emotions, potentially enabling text--based Emotion management.DiffTalk~\cite{shen-2023} proposes a talking head synthesis method that generates high-quality visuals in sync with the original audio through audio-driven, temporally coherent denoising. It captures the temporal relationships in the audio and learns to predict the corresponding facial movements over time using a Recurrent Neural Network (RNN) architecture. The method utilizes a single image as input in the multi-shot Diffused Heads~\cite{stypułkowski2023diffusedheadsdiffusionmodels} technique, allowing for the creation of a single talking head. However, it has some disadvantages, including its reliance on reference face photos, the computational cost associated with repeated denoising techniques, and the potential introduction of artifacts in the output videos. By leveraging diffusion models' capability to synthesize high-quality video frames sequentially, it aims to improve the overall synthesis process. Diffused Heads~\cite{stypułkowski2023diffusedheadsdiffusionmodels}, an autoregressive diffusion model, aims to create realistic talking head movies. Using a GAN architecture within an autoregressive diffusion framework, the model generates successive video frames, each dependent on the audio input and the frames that came before it. The CREMA and LRW datasets, which focus on Emotional expressions and extensive lip reading datasets, are used to evaluate it. Diffusion Video Editing~\cite{bigioi2023speechdrivenvideoediting} is an innovative speech-driven method. This technique showcases the ability of diffusion models to generate new content and modify existing films based on audio input. It has potential applications in tasks like lip synchronization and adjusting facial expressions to match new audio. The method integrates a U-Net architecture inside a denoising diffusion model to enable targeted changes to the video content based on the audio, presumably learning to denoise video frames conditioned on the input speech. DAE-Talker~\cite{Du_2023} uses latent representations learned by a Denoising Autoencoder (DAE) to improve speech-driven talking face synthesis. By utilizing powerful feature representations that the DAE has learned from audio, it aims to outperform earlier methods regarding lip synchronization accuracy, video quality, pose naturalness, and pose controllability. Pose accuracy and control, efficient pose modeling, denoising performance (within the DAE), generalization to unseen identities or speaking styles, the need for improved evaluation metrics for THG, the DAE's data dependency, and the system's overall computational complexity are some of the issues that the paper recognizes DAE-Talker~\cite{Du_2023} faces. DreamTalk~\cite{ma2024dreamtalkemotionaltalkinghead} is a diffusion model framework designed to create realistic talking heads exhibiting various Emotions. Controlling the Emotional style of the generated talking head consists of a style predictor, a style-aware lip expert, and a denoising network. The method addresses problems such as Emotion recognition in low-Emotion-intensity audio, speaker identity retention, and consistent expressions. DiT-Head~\cite{mir2023ditheadhighresolutiontalkinghead} is a novel talking head synthesis pipeline that uses Diffusion Transformers (DiT) with audio input for scalability and competitive performance. It has demonstrated promise for many applications and is evaluated using the HDTF~\cite{zhang-2021} dataset, a common benchmark for talking head synthesis. However, its reliance on data from English speakers and the substantial computer resources required for training and inference limit it. A non-deterministic deep learning model, FaceDiffuser~\cite{stan2023facediffuserspeechdriven3dfacial}, was developed for speech-driven 3D face animation synthesis. It uses an internal dataset in addition to 3D vertex and blend shape datasets to learn realistic 3D facial motions directly from speech. The BIWI, Multiface, BEAT, and UUDaMM datasets are used to train the model, which allows it to learn realistic speech-driven 3D facial deformations. With a focus on probabilistically sampling holistic lip-irrelevant facial gestures to match the input audio while maintaining photorealism and naturalness, PAVDP~\cite{yu2022talkingheadgenerationprobabilistic} offers a novel framework for one-shot audio-driven talking head creation. It uses a generator network (G) that was trained similarly to PC-AVS~\cite{zhou2021posecontrollabletalkingfacegeneration} but focuses on using probabilistic diffusion prior models to predict non-lip facial movements. The Adaptive Diffusion Landmark Dynamic Rendering (DLDR) \cite{ying-2025} framework also leverages transformer-based audio semantic mapping and landmark alignment to achieve precise audio-visual synchronization. Combining diffusion models with dynamic rendering demonstrates significant improvements in mouth-shape realism and perceptual quality on benchmark datasets. A variety of architectures are used in the reviewed diffusion--based THG techniques, including GANs in diffusion frameworks, CLIP--based models, RNNs with denoising, Diffusion Transformers, U-Nets for video editing, Denoising Autoencoders, and specialized diffusion models with style-aware elements. These methods focus on advances in speech-driven video editing, expressiveness, smoothness, Emotionally editable talking heads, motion disentanglement, pose control, lip synchronization, and high-quality synthesis. Some main disadvantages include long sampling times, high computing costs, the potential for artifacts, challenges with generalization and identity retention, reliance on large and diverse datasets, and ethical issues related to creating realistic synthetic media.

\subsubsection{3D Animation--Based THG}

3D animation pipelines integrate motion capture, physics simulations, and rigged models to deliver film and VFX quality results. Learn2Talk~\cite{zhuang2024learn2talk3dtalkingface}, MultiTalk~\cite{sungbin2024multitalkenhancing3dtalking}, and Speech4Mesh~\cite{10378326} are important models based on Google Scholar citation metrics. While MultiTalk~\cite{sungbin2024multitalkenhancing3dtalking} uses VQ-VAE with language embeddings and self-collected multilingual films, Learn2Talk~\cite{zhuang2024learn2talk3dtalkingface} uses TCN for lip-sync and vertex accuracy. Speech4Mesh~\cite{10378326} uses the FLAME--based~\cite{FLAME:SiggraphAsia2017} audio2mesh network and the MEAD~\cite{kaisiyuan2020mead} dataset. Obstacles include realistic and creative control. Future possibilities for realistic cloth and hair modeling include generative rigging and differentiable physics.

\begin{sidewaystable}[htbp]
\scriptsize
\caption{Comparison of 3D Animation-Based Approaches}
\label{tab:animation_based}
\centering
\begin{tabular}{@{}lll>{\raggedright\arraybackslash}p{4.2cm}>{\raggedright\arraybackslash}p{4.2cm}l@{}}
\toprule
\textbf{Method} & \textbf{Arch.} & \textbf{Dataset} & \textbf{Highlights} & \textbf{Limitations} & \textbf{N-shot} \\
\midrule
MultiTalk~\cite{sungbin2024multitalkenhancing3dtalking} & VQ-VAE & MultiTalk~\cite{sungbin2024multitalkenhancing3dtalking} & Multilingual avatars with style embeddings for lip sync. & Language limits and non-standard speech issues. & Multi \\
CSTalk~\cite{liang2024cstalkcorrelationsupervisedspeechdriven} & TCN & Live Link Face & 3D lip + Emotion sync aligned to speech. & Narrow emotion range; complex cues poorly handled. & Multi \\
Learn2Talk~\cite{zhuang2024learn2talk3dtalkingface} & TCN & BIWI, VOCASET~\cite{williams-no-date} & 2D generation fused with 3D for clarity. & Weak realism in blinking/emotion; multitasking issues. & Multi \\
DiffSpeaker~\cite{ma2024diffspeakerspeechdriven3dfacial} & GRU & BIWI, VOCASET~\cite{williams-no-date} & Fast transformer-based 3D facial motion. & Struggles with realism + 4D audio mapping. & Multi \\
Media2Face~\cite{zhao2024media2facecospeechfacialanimation} & VAE & BIWI, VOCASET~\cite{williams-no-date} & Diffusion + parametric asset + dataset fusion. & Compute-heavy; dataset-dependent. & Multi \\
PMMTalk~\cite{han2023pmmtalkspeechdriven3dfacial} & CNN & 3D-CAVFA, VOCASET~\cite{williams-no-date} & Pseudo-multimodal lip-blendshape sync. & No head/exp control; inference slow. & Multi \\
3DiFACE~\cite{thambiraja20233difacediffusionbasedspeechdriven3d} & Transformer & VOCASET~\cite{williams-no-date} & Personalized diffusion-based 3D animation. & Small dataset, editing limits. & Multi \\
Speech4Mesh~\cite{10378326} & FLAME & MEAD~\cite{kaisiyuan2020mead}, VoxCeleb2~\cite{chung-20018} & Audio2Mesh system handles 4D mesh scarcity. & Prone to overfitting, poor generalization. & One \\
PV3D~\cite{xu2023pv3d3dgenerativemodel} & GAN & VoxCeleb, CelebV-HQ~\cite{zhu2022celebvhqlargescalevideofacial}, TH-1KH & GAN + motion modeling for 3D-aware output. & Quality limited by 2D data. & One \\
3D-TalkEmo~\cite{wang20213dtalkemolearningsynthesize3d} & StarGAN & Custom & Emotion-adaptive 3D talking head generation. & Low scalability; data/tuning bottlenecks. & Multi \\
\bottomrule
\end{tabular}
\end{sidewaystable}

The domain of 3D animation-based talking head generation has witnessed a surge in approaches aiming to produce expressive, synchronized, and controllable facial animations from audio or multimodal inputs. Tables \ref{tab:animation_based} and \ref{tab:animation_based} summarize a wide array of such methods, categorizing them based on architecture, datasets, core contributions, limitations, and the number of input instances (N-shot). These methods explore the fusion of generative modeling with 3D mesh control, emotional expressiveness, and personalized motion synthesis.
In Table \ref{tab:animation_based}, the first entry, MultiTalk, employs a VQ-VAE framework to enable multilingual avatar synthesis with style embeddings tailored for accurate lip synchronization. Despite its robustness in language-specific articulation, it struggles with non-standard speech and linguistic nuances. CSTalk, built upon Temporal Convolutional Networks (TCNs), integrates 3D lip and emotional synchronization. However, it has a limited emotional range and cannot effectively capture subtle or complex affective cues. Learn2Talk innovates by combining 2D image generation and 3D animation for enhanced clarity. While effective for simple speech, it lacks realism in secondary expressions like blinking and exhibits task-switching challenges. DiffSpeaker introduces a GRU-based design to drive facial motion using fast, transformer-like sequences, but its realism and capacity for handling 4D audio integration remain limited. Media2Face stands out for merging diffusion modeling with parametric animation and dataset blending, achieving fine-grained lip movement. However, this comes at the cost of heavy computational requirements and a high dependency on dataset alignment.
Table \ref{tab:animation_based} continues this trend, focusing on models pushing the speech-driven 3D mesh synthesis envelope. PMMTalk leverages CNNs for pseudo-multimodal blendshape generation, achieving effective lip synchronization but lacking comprehensive control over head poses and facial expressions. It also suffers from slow inference times. 3DiFACE adopts a transformer-based diffusion approach tailored to personalized mesh outputs. Despite the strength of personalization, its reliance on limited datasets constrains generalizability and editing flexibility. Speech4Mesh addresses the scarcity of annotated 4D meshes by proposing an Audio2Mesh pipeline using the FLAME model. However, it tends to overfit and generalize poorly across domains. PV3D, using GANs, combines 3D-aware generation with motion modeling. While it offers stylistically rich outputs using datasets like VoxCeleb and CelebV-HQ, its dependency on 2D video data affects 3D consistency. Finally, 3D-TalkEmo, based on StarGAN, focuses on emotion-adaptive synthesis for expressive talking heads. While it provides rich emotional dynamics, its scalability is hindered by tuning complexity and limited training data.
Collectively, these methods showcase a growing emphasis on integrating multimodal cues (speech, emotion, style) with 3D-aware modeling. The dominant architectures span traditional RNNs and CNNs to modern transformers, diffusion models, and variational frameworks. Despite improvements in fidelity and personalization, challenges persist—particularly regarding scalability, inference efficiency, emotional generalization, and non-standard linguistic or acoustic input handling.
Regarding datasets, VOCASET and BIWI dominate due to their detailed 3D annotations, while newer datasets like MultiTalk and 3D-CAVFA offer specific benefits such as multilinguality and multimodal fidelity. Most methods operate in a multi-shot regime, requiring several input samples or training iterations per subject. However, methods like Speech4Mesh and PV3D progress toward one-shot generation, reflecting efforts to reduce data reliance.
In summary, the surveyed methods reflect a dynamic research trajectory aimed at bridging expressiveness, realism, and control in 3D talking heads, with each contributing uniquely across the axes of animation realism, emotional fidelity, and computational feasibility.

Various diffusion 3D animation models based on THG have been investigated in Table~\ref{tab:animation_based} techniques for creating 3D talking head animations, generally controlled by audio input. MultiTalk~\cite{sungbin2024multitalkenhancing3dtalking}, CSTalk~\cite{liang2024cstalkcorrelationsupervisedspeechdriven}, Learn2Talk~\cite{zhuang2024learn2talk3dtalkingface}, DiffSpeaker~\cite{ma2024diffspeakerspeechdriven3dfacial}, Media2Face~\cite{zhao2024media2facecospeechfacialanimation}, and BIWI are some of the methods. The multilingual 2D video dataset MultiTalk~\cite{sungbin2024multitalkenhancing3dtalking} aims to generate expressive and lifelike 3D head positions and facial gestures. It uses a VQ-VAE architecture to create a discrete latent representation of facial movements that can be conditioned by language and audio embeddings. A speech-driven 3D facial animation technique, CSTalk~\cite{liang2024cstalkcorrelationsupervisedspeechdriven}, aims to outperform techniques in several areas, such as handling data constraints, achieving precise lip alignment with the audio, and producing a wider range of Emotions in the 3D face. To interpret sequential input, such as audio, and produce corresponding temporal sequences of 3D facial motions, it uses a TCN architecture. Some disadvantages are the computational complexity of training and managing 3D facial animation models, the potentially limited range of possible Emotional expressions, and the subjectivity involved in interpreting and producing subtle Emotional cues. By combining knowledge from audio-video synchronization networks with experience from 2D talking face production, Learn2Talk~\cite{zhuang2024learn2talk3dtalkingface} aims to improve both 2D and 3D talking face research. It aims to improve the generated talking heads' overall speech perception, vertex correctness (in 3D models), and lip-sync accuracy. It uses a TCN architecture comparable to CSTalk~\cite{liang2024cstalkcorrelationsupervisedspeechdriven}, demonstrating how well it models the temporal relationship between facial and audio motions for improved synchronization and animation. A Transformer--based network, DiffSpeaker~\cite{ma2024diffspeakerspeechdriven3dfacial}, was created to improve speech-driven 3D face animation performance. The main innovation is creating parallel face movements for better performance and faster inference speeds than sequential or autoregressive models. Its reliance on audio-4D data, its inability to generalize to invisible identities or speech patterns, and the potential trade-offs between achieving flawless lip synchronization and producing genuine non-verbal facial Emotions are some of its limitations. Using a Generalized Neural Parametric face asset, the M2F-D dataset, and a Media2Face~\cite{zhao2024media2facecospeechfacialanimation} diffusion model, Media2Face~\cite{zhao2024media2facecospeechfacialanimation} is a three-step method for creating 3D face animations from speech. Realistic and emotive animations conditioned on voice input are created using the diffusion model framework's VAE architecture. Like Learn2Talk~\cite{zhuang2024learn2talk3dtalkingface} and DiffSpeaker~\cite{ma2024diffspeakerspeechdriven3dfacial}, the method is evaluated on the BIWI and VOCA~\cite{williams-no-date}SET datasets. A new framework called PMMTalk~\cite{han2023pmmtalkspeechdriven3dfacial} aims to improve the precision of speech-driven 3D face animation. It utilizes pseudo-multi-modal characteristics to take advantage of information from various modalities, such as text and audio. To capture the complex relationship between speech and facial motions, PMMTalk~\cite{han2023pmmtalkspeechdriven3dfacial} uses a CNN architecture to extract audio data and map it to face shape coefficients. 3DiFACE~\cite{thambiraja20233difacediffusionbasedspeechdriven3d} is a novel method for editing and animating 3D faces using speech. A lightweight audio-conditioned diffusion model creates stochasticity and provides motion editing capabilities in the resulting animations. The diffusion component enables the creation of different and potentially adjustable animations, while the Transformer architecture depicts the interaction between audio and 3D facial motions. The study lists limitations like comparatively small training datasets, the challenge of achieving accurate motion control and editing capabilities, and the computational complexity of diffusion--based models. A 3D face animation system called Speech4Mesh~\cite{10378326} manages controllability and data scarcity in speech-driven 3D facial animation. For more control over the generated animations, it encodes speaking factors, generates 4D talking head data, and trains an audio2mesh network to translate audio to 3D face models. FLAME parameters are based on audio characteristics, and the technology trains an audio2mesh network that uses FLAME as the underlying format for 3D face models. A generative framework called PV3D~\cite{xu2023pv3d3dgenerativemodel} was created to create 3D-aware portrait films. By offering methods for describing motion dynamics and incorporating discriminators that operate in both spatial and temporal dimensions, it expands static GANs into the video domain. The VoxCeleb, CelebV-HQ~\cite{zhu2022celebvhqlargescalevideofacial}, and TalkingHead-1KH~\cite{tcwang-no-date} datasets evaluate PV3D~\cite{xu2023pv3d3dgenerativemodel}, which focuses on producing realistic and excellent 3D-aware video portraits. A deep neural network called 3D-3D-TalkEmo~\cite{wang20213dtalkemolearningsynthesize3d} is designed to produce dynamic 3D talking-head animations with modifiable Emotion states. In terms of the quality and controllability of the generated Emotional expressions in 3D talking heads, it aims to surpass earlier methods. The framework uses a StarGAN design, which is renowned for its ability to translate images across multiple domains. The explored 3D animation model--based THG techniques, which primarily use multi-shot learning, make use of a variety of architectures, including CNNs, FLAME, StarGANs, TCNs, Transformers, VQ-VAEs, and conventional VAEs inside diffusion frameworks. Advances in multilingual lip synchronization, improved lip alignment and Emotional expression, improved lip-sync and vertex accuracy, parallel motion generation for faster inference, high-fidelity and expressive animation through diffusion models, precise lip-syncing with blend shape coefficients, personalized animation with stochasticity and editing, addressing data scarcity through 4D data generation, producing 3D-aware portrait videos with motion dynamics, and producing vibrant 3D talking heads with adjustable Emotions are all highlighted in these methods.

\subsection{DATASET}

THG depends on high-quality datasets to train lip synchronization, Emotion synthesis, and pose control models. Key datasets used in the field include audio-driven datasets such as VoxCeleb, CREMA-D~\cite{cheyneycomputerscience-no-date}, LRW~\cite{chung-2016} (Lip Reading in the Wild), MEAD~\cite{kaisiyuan2020mead}, Video-Driven datasets like TalkingHead-1KH~\cite{tcwang-no-date}, HDTF~\cite{zhang-2021}, CelebV-HQ~\cite{zhu2022celebvhqlargescalevideofacial}, Text-Driven datasets like CelebrV-Text, Emotion-Specific datasets like RAVDESS~\cite{livingstone-2018}, VoxCeleb1~\cite{nagrani-2019}, 3D and Multi-View datasets like Multiface, Vocaset, and GRD. Audio-driven datasets focus on synchronizing facial movements with speech inputs. VoxCeleb1~\cite{nagrani-2019} contains over 1,251 and 6,112 speakers from YouTube, with 153,000+ audio-video clips. These datasets are used for cross-identity reenactment, speaker verification, and Emotion-aware synthesis. However, they have limitations such as lab control, limited ethnic diversity, low resolution, lack of non-frontal poses, small sample size, and lab environment. Video-driven datasets are used for motion transfer and identity preservation across videos. TalkingHead-1KH~\cite{tcwang-no-date} contains 1,000+ hours of talking head videos from YouTube, while HDTF~\cite{zhang-2021} includes 362 speakers and 10,000 clips with extreme poses. CelebV-HQ~\cite{zhu2022celebvhqlargescalevideofacial} contains 15,653 high-resolution videos of celebrities, while BIWI--3D provides 3D scans of 14 speakers with head pose annotations. Emotion-specific datasets focus on nuanced Emotional expressions, with RAVDESS~\cite{livingstone-2018} containing 24 actors performing eight Emotions in speech and song. VoxCeleb1~\cite{nagrani-2019} contains 652 naturalistic Emotional dialogues from 12 speakers, while 3D and Multi-View datasets enable 3D-aware synthesis and free-view rendering. Vocaset contains 3D facial animations from 12 speakers, while GRD contains 33,000 clips of 34 speakers reciting grammatically fixed sentences. Cross-modal datasets combine multiple modalities (text, audio, video) using LRS3-TED~\cite{afouras-2018} and GRD. These datasets have limitations, mainly English and limited Emotion labels, but can improve lip-reading models' accuracy. THG relies heavily on diverse, high-quality datasets for tasks like lip synchronization, Emotion synthesis, and pose control. These datasets have their limitations, such as a bias towards frontal views and English speakers, limited ethnic diversity, and limitations in the use of different modalities. Researchers can develop more effective and efficient models for generating realistic and engaging talking head content by analyzing these datasets and identifying their limitations.

\begin{sidewaystable}[htbp]
\scriptsize
\caption{Overview of Various Datasets}
\label{tab:dataset}
\centering
\begin{tabular}{@{}l>{\raggedright\arraybackslash}p{4cm}cccccccccc@{}}
\toprule
\textbf{DATASET} & \textbf{HIGHLIGHTS} & \textbf{YEAR} & \textbf{HOUR} & \textbf{SPEAKERS} & \textbf{SENTENCES} & \textbf{ENV.} & \textbf{VIEW} & \textbf{CATEGORY} & \textbf{FPS} & \textbf{RESOLUTION} \\
\midrule
CelebV-Text~\cite{yu2023celebvtextlargescalefacialtextvideo} & Speech-driven facial animation & 2023 & 279 & 70000 & 1.4M & Wild & Frontal & Video & -- & 512×512 \\
MMFace4D~\cite{wu-2023} & 4D motion capture for face animation & 2023 & 36 & 431 & 11K & Lab & Frontal & Video, Audio & 30 & -- \\
CelebV-HQ~\cite{zhu2022celebvhqlargescalevideofacial} & High-quality celebrity videos & 2022 & 65 & 15653 & -- & Wild & -- & Video & -- & 512×512 \\
Multiface (3D)~\cite{facebookresearch-no-date} & 3D facial scan dataset & 2022 & -- & 13 & -- & Lab & Multi-view & Image, Audio & 1 & 2048×1334, 1024×1024 \\
Responsive Listening Head~\cite{10.1145/3414685.3417774} & 3D avatars responding to audio & 2022 & 1.5 & 67 & 483 & Wild & Frontal & Video, Audio, Image & 30 & 384×384 \\
BIWI (3D)~\cite{doubiiu-no-date} & Annotated 3D face scans & 2021 & 1.44 & 14 & 1109 & Lab & Multi-view & Video & 25 & -- \\
HDTF~\cite{zhang-2021} & Dataset for head pose estimation & 2021 & 15.8 & 362 & 10K & Wild & -- & Video & -- & 1280×720, 1920×1080 \\
TalkingHead-1KH~\cite{tcwang-no-date} & Emotion-rich talking videos & 2021 & 1000 & -- & -- & Wild & -- & Video & -- & -- \\
MeshTalk~\cite{richard-2021} & 3D talking heads driven by speech & 2021 & 13 & 250 & 12.5K & Lab & -- & Video & 30 & 800×800 \\
MEAD~\cite{kaisiyuan2020mead} & Emotion-labeled high-res video/audio & 2020 & 39 & 60 & 20 & Lab & Multi-view & Video, Audio & 30 & 1920×1080 \\
FaceForensics++~\cite{ondyari-no-date} & Deepfake detection corpus & 2019 & 5.7 & 1000 & 1000+ & Wild & Frontal & Video, Image & -- & 512×512 \\
FFHQ~\cite{nvlabs-no-date} & GAN-ready diverse face photos & 2019 & -- & -- & -- & Wild & -- & Image & 1 & -- \\
VOCA~\cite{williams-no-date} & Vocal tract visualization via video & 2019 & 0.5 & 12 & 40 & Lab & Frontal & Video, Audio & 60 & 2000×2000 \\
Lombard~\cite{10.1121/1.5042758} & Speech under noise conditions & 2018 & 3.6 & 54 & 5400 & Lab & Multi-view & Video, Audio & -- & 720×480, 864×480 \\
LRS2~\cite{Chung17} & Lip-reading dataset from BBC TV & 2018 & 224.5 & 500+ & 1.4M & Wild & Multi-view & Video, Audio, Text & -- & 224×224 \\
LRS3-TED~\cite{afouras-2018} & TED-based AV speech dataset & 2018 & 438 & 5000 & 152K & Wild & Multi-view & Video, Audio, Text & -- & 224×224 \\
MELD~\cite{declare-lab-no-date} & Emotion-rich dialogue videos & 2018 & 13.7 & 407 & -- & Wild & -- & Video, Audio, Text & -- & -- \\
RAVDESS~\cite{livingstone-2018} & Emotion expression by actors & 2018 & 7 & 24 & 2 & Lab & Frontal & Video, Audio & -- & 1920×1080, 1280×720 \\
MODALITY~\cite{zhou2024enhancingsnnbasedspatiotemporallearning} & Multimodal affective dataset & 2017 & 31 & 35 & 5800 & Lab & -- & Audio, Video & -- & 1920×1080 \\
ObamaSet~\cite{andreotti-2024} & Lip-sync and deepfake research dataset & 2017 & 14 & 1 & -- & Wild & -- & Video & -- & -- \\
VoxCeleb~\cite{nagrani-no-date} & Celeb voice video collection & 2017 & 352 & 1200 & 153.5K & Wild & -- & Video & 25 & 224×224 \\
VoxCeleb1~\cite{nagrani-2017} & Emotion in AV communication & 2016 & 18 & 12 & 652 & Lab & Frontal & Audio, Video, Image & -- & 1440×1080 \\
VoxCeleb2~\cite{nagrani-2019} & Unconstrained celeb speech data & 2017 & 352 & 1251 & 153K & Wild & -- & Video & 30 & 224×224 \\
LRW~\cite{chung-2016} & Word-level lip-reading videos & 2016 & 173 & 1000 & 539K & Wild & Frontal & Video, Text & -- & 256×256 \\
SAVEE~\cite{jackson-2015} & Emotional speech videos & 2015 & -- & 480 & -- & Lab & Frontal & Video, Audio & 60 & -- \\
TCD-TIMIT~\cite{7050271} & AV corpus for speech recognition & 2015 & 11.1 & 62 & 6900 & Lab & Multi-view & Video, Audio & -- & 1920×1080 \\
CREMA-D~\cite{cheyneycomputerscience-no-date} & Emotionally expressive AV clips & 2014 & 11.1 & 91 & 12 & Lab & -- & Video, Audio & -- & 960×720 \\
GRID~\cite{10.1121/1.5042758} & Speech + lip reading video corpus & 2006 & 27.5 & 34 & 33K & Lab & Frontal & Video, Audio & -- & 360×288, 720×576 \\
DPCD~\cite{metaverse-ai-lab-thu-no-date} & 3D Emotion + dynamic pose scans & 2004 & 29.75 & 5 & 48.6K & Wild & Multi-view & Video, Audio, Text & 25 & -- \\
CAVSR1.0~\cite{li-100080} & Early lip sync + facial animation set & 1998 & -- & 123 & 600 & Lab & Frontal & Video, Audio & -- & -- \\
\bottomrule
\end{tabular}
\end{sidewaystable}

A variety of training and evaluation datasets influences research on THG. The varied character of this activity is shown in the datasets' disparities in size, recording environment, view kinds, included modalities, motion capture data presence, and resolution. Emphasizing their important qualities relevant to building avatar and video synthesis algorithms, Table \ref{tab:dataset} thoroughly summarizes this domain's most often used datasets. Creating efficient avatar and video synthesis algorithms depends on the size and duration of these datasets. A notable resource for applications requiring large amounts of diverse audio-visual data is VoxCeleb2 (2018), which offers an astounding 2400 hours of video with 6100 speakers and 1.1 million words. With 438 hours of TED Talks from 5000 speakers and 152,000 phrases, LRS3-TED~\cite{afouras-2018} (2018) similarly provides various speaking styles and linguistic material for lip-reading and speech-to-text research. Conversely, CelebV-Text~\cite{yu2023celebvtextlargescalefacialtextvideo} emphasizes lip sync features, which are perfect for training algorithms concentrating on accurate mouth movements; datasets like VOCA~\cite{williams-no-date} (2019) provide high-quality 3D facial motion capture data. Recording conditions greatly alter the characteristics of the dataset. Many datasets are classified as "WILD," meaning they were recorded in unrestricted, real-world settings. Examples of these datasets include VoxCeleb (general) (2017), VoxCeleb1~\cite{nagrani-2019} (2017), VoxCeleb2 (2018), LRS2~\cite{Chung17} (BBC) (2018), LRS3-TED~\cite{afouras-2018} (2018), FFHQ~\cite{nvlabs-no-date} (2019), HDTF~\cite{zhang-2021} (2021), TalkingHead-1KH~\cite{tcwang-no-date} (2021), CelebV-HQ~\cite{zhu2022celebvhqlargescalevideofacial} (2022), and CelebrV-Text (2023). Although it presents challenges, this variation in background noise, lighting, and speaker position makes models trained on such data more resilient. In contrast, datasets such as MMFace4D~\cite{wu-2023} (2023), BIWI (3D)~\cite{doubiiu-no-date} (2021), VOCA~\cite{williams-no-date} (2019), Lombard~\cite{10.1121/1.5042758} (2018), RAVDESS~\cite{livingstone-2018} (2018), MEAD~\cite{kaisiyuan2020mead} (2020), VoxCeleb1~\cite{nagrani-2019} (2016), SAVEE (2015), TCD-TIMIT (2015), CREMA-D~\cite{cheyneycomputerscience-no-date} (2014), GRAD (2006), and CAVSR1.0~\cite{li-100080} (1998) are recorded in controlled "LAB" environments, providing cleaner data with consistent background and lighting, which can be useful for isolating particular factors like lip movements or Emotional expressions. Another important consideration in the study of THG is view type. "FRONTAL-VIEW" recordings make up many datasets, particularly those that focus on talking head creation and lip-reading, which facilitates the learning process for lip motions. Nonetheless, datasets such as Multiface (3D)~\cite{facebookresearch-no-date} (2022), BIWI (3D)~\cite{doubiiu-no-date} (2021), Lombard~\cite{10.1121/1.5042758} (2018), LRS2~\cite{Chung17} (BBC) (2018), LRS3-TED~\cite{afouras-2018} (2018), MEAD~\cite{kaisiyuan2020mead} (2020), and TCD-TIMIT (2015) provide "MULTI-VIEW" data, which is required to train models that can control multiple head postures and produce talking heads that are aware of three dimensions. "N/A" is specified for view type in datasets such as CelebV-HQ~\cite{zhu2022celebvhqlargescalevideofacial} (2022), HDTF~\cite{zhang-2021} (2021), TalkingHead-1KH~\cite{tcwang-no-date} (2021), FFHQ~\cite{nvlabs-no-date} (2019), VoxCeleb2 (2018), VoxCeleb (general) (2017), VoxCeleb1~\cite{nagrani-2019} (2017), ObamaSet (2017), and CREMA-D~\cite{cheyneycomputerscience-no-date} (2014), and this indicates variability or that this information is not the primary focus. These datasets are typically large-scale collections for general facial analysis or recognition. Modalities include "VIDEO" and "AUDIO" data to comprehend the connection between speech and facial movements. This essential combination is provided by a few datasets, including MMFace4D~\cite{wu-2023} (2023), Responsive Listening Head~\cite{10.1145/3414685.3417774} creation (2022), BIWI (3D)~\cite{doubiiu-no-date} (2021), FaceForensics++~\cite{ondyari-no-date} (2019), MeshTalk~\cite{richard-2021} (2021), FFHQ~\cite{nvlabs-no-date} (2019), MODALITY~\cite{zhou2024enhancingsnnbasedspatiotemporallearning} (2017), and GAN--based face creation. Motion capture data is essential to train models to produce lifelike 3D face animations. The quality and smoothness of the generated animations are defined by frame rate (fps). High-resolution 3D models or photos are provided by high-resolution datasets such as Multiface (3D)~\cite{facebookresearch-no-date} (2022), MEAD~\cite{kaisiyuan2020mead} (2020), RAVDESS~\cite{livingstone-2018} (2018), TCD-TIMIT (2015), and DPCD~\cite{metaverse-ai-lab-thu-no-date} (2004). Additionally, another dataset that is not relatively cited but has an impact in the field of THG includes Hallo3~\cite{cui2024hallo3}, VFHQ~\cite{9857333}, and few others.

\subsection{LOSS FUNCTION}

In DL algorithms \cite{sung-bin-2023}, loss functions \cite{harrell-1984} are fundamental since they define the difference between expected and actual values. They guide the optimization process to improve model accuracy. Selecting a loss function requires thoughtful evaluation of its relevance for a particular task.
\textbf{Convexity:} A loss function is convex if its local minimum is the global minimum, making it easy to optimize using gradient--based methods.
\textbf{Differentiability:} A loss function's derivative concerning model parameters exists and is continuous, enabling gradient--based optimization.
A few extreme values should not influence loss functions; they should handle outliers.
There should not be abrupt transitions or spikes in a loss function; rather, it should have a continuous gradient.
A sparsity-promoting loss function benefits high-dimensional data and small features by promoting sparse output.
\textbf{Monotonicity:} By guaranteeing the optimization process moves toward the proper solution, a loss function's value drops as the predicted output approaches the actual production.
Based on the type of learning task, loss functions, which gauge how well an algorithm interprets data, are divided into two groups: classification models, which forecast the output from a set of finite categorical values, and regression models, which predict continuous values. This section examines the common loss functions used in classification and regression tasks. Regardless of whether the problem involves regression or classification, selecting the appropriate loss function is essential for optimizing the model to meet the specific requirements of the task.

\subsubsection{Regression}

Regression is a potent analytical technique that uses input features to predict continuous output values. This method is widely used in many fields, including finance (aiding in stock price prediction), healthcare (estimating patient outcomes), social sciences (analyzing trends and behaviors), sports (evaluating player performance), and engineering (optimizing design processes). Regression analysis allows us to find important information and make predictions based on data. 

Practical applications include house price prediction~\cite{han2022tellhowvideosynthesis}, energy consumption forecasting~\cite{zhang-2021}, healthcare and disease prediction~\cite{harrell-1984}, stock price forecasting~\cite{shen-2023}, and customer lifetime value prediction~\cite{ma2024diffspeakerspeechdriven3dfacial}.

A regression loss function is defined over a dataset:
\[
D = \left\{ (x_i, y_i) \right\}_{i=1}^{n}
\]
where \( x_i \in \mathbb{R}^d \) denotes the \( d \)-dimensional input feature vector and \( y_i \in \mathbb{R} \) the corresponding continuous target value. 

The aim is to find the parameter vector \( \boldsymbol{\theta} \) that minimizes the overall loss function:
\[
\min_{\boldsymbol{\theta}} \sum_{i=1}^{n} \mathcal{L}(f_{\boldsymbol{\theta}}(x_i), y_i)
\]

\subsubsection{Classification}

Deep Learning Talking head models involve sorting data into categories based on specific features or characteristics. There are two main classification types: binary classification, which aims to sort data into two distinct categories, and multi-class classification, which sorts data into more than two categories. Because binary classification involves choosing between two options, it is the most basic classification. Multi-class classification, on the other hand, entails grouping data into several categories, as in image recognition systems. Multi-label classification allows a single data point to belong to multiple classes simultaneously. While this approach is relevant in specific applications, it may not be as essential for a basic overview of classification. The decision boundary in a classification plot represents the line or area that the model uses to determine the category of an image. Classification uses a labeled dataset to train a model to identify patterns and relationships in the data. Data collection, feature extraction, model training, assessment, and prediction are all steps in this process. The model uses existing labeled data to teach itself how to predict the new, unlabeled data class based on the learned patterns. Classification algorithms are widely used in real-world applications such as email spam filtering, credit risk assessment, medical diagnosis, image classification, sentiment analysis, fraud detection, and recommendation systems. These algorithms are designed to handle binary and multi-class classification tasks, depending on the problem's nature. Classification modeling uses machine learning algorithms to categorize data into predefined classes or labels. Key characteristics of classification models include class separation, decision boundaries, sensitivity to data quality, handling imbalanced data, and interpretability. Well-labeled and representative data are essential for better performance, while noisy or biased data can result in poor predictions. Special techniques such as resampling or weighting are employed to address class imbalances. Classification involves using existing labeled data to train a model to predict the new, unlabeled data class based on the learned patterns.

\subsubsection{Unsupervised}
Until now, the discussion has focused on methods used in the context of supervised learning; however, unsupervised learning is one example of how they differ from supervised learning. Finding some structure in the data is the aim of unsupervised learning, which uses inputs \( x \) without matching output pairs. While unsupervised learning techniques operate differently and uncover various structures, they all share a common component. Utilizing the same concepts of hypothesis functions, loss functions, and optimization techniques as supervised learning, we can define unsupervised learning in a general way. Almost all unsupervised learning techniques can be seen this way, despite the various contexts.

To compensate for the absence of a clear target that we are attempting to fit, we must modify the definitions of hypotheses and loss functions. Two specific unsupervised learning algorithms that fall under this framework are the principal component analysis (PCA) algorithm and the \( k \)-means clustering algorithm. 

The hypothesis function \( h: \mathbb{R}^n \to \mathbb{R}^n \) maps input \( \mathbb{R}^n \) back into the input space. Similar to the supervised case, the objective of this hypothesis class is to approximate the reconstruction of the input. The specific configuration of unsupervised learning algorithms limits the class of hypothesis functions, requiring them to efficiently recover a certain amount of structure in the data to accurately approximate their input rather than output the function's input.

The loss function indicates the degree to which the prediction deviates from the original input: 
\[
\ell: \mathbb{R}^n \times \mathbb{R}^n \to \mathbb{R}_{+}.
\]
The \( k \)-means and principal component analysis algorithms are both based on the same loss function, given by the formula:
\[
\ell(h(x), x) = \| h(x) - x \|_2^2.
\]
\begin{sidewaystable}[htbp]
\scriptsize
\centering
\caption{Benchmark Loss Functions Used in Talking Head Generation}
\label{tab:loss_functions}
\begin{tabular}{@{}l>{\raggedright\arraybackslash}p{4.5cm}>{\raggedright\arraybackslash}p{4.3cm}>{\raggedright\arraybackslash}p{4.3cm}l@{}}
\toprule
\textbf{Loss Function} & \textbf{Formula} & \textbf{Description} & \textbf{Limitation} & \textbf{Implementation Type} \\
\midrule
Square Loss ($L_2$) & $\ell_{\text{square}} = \frac{1}{S} \sum_{i=1}^{S} (x_i - \hat{x}_i)^2$ & Penalizes large errors quadratically & May blur structure in visual outputs & Image--based, Video--based \\
Absolute Loss ($L_1$) & $\ell_{\text{abs}} = \frac{1}{n} \sum_{i=1}^{n} |x_i - \hat{x}_i|$ & Simple and robust to outliers & May not differentiate well between large/small errors & Image--based, Video--based \\
Epsilon-Insensitive Loss & $\ell_\epsilon = \max(0, |x - \hat{x}| - \epsilon)$ & Ignores small errors within $\epsilon$-tube, used in SVR & Non-smooth, no penalty for small errors & Audio--based, Video--based \\
IoU Loss & $\ell_{\text{IoU}} = 1 - \frac{|A \cap B|}{|A \cup B|}$ & Optimizes overlap in detection tasks & Gradient is zero if no overlap & Image--based \\
Generalized IoU Loss & $GIoU = IoU - \frac{|C \setminus (A \cup B)|}{|C|}$ & Extends IoU for non-overlapping cases & Slightly complex to compute & Image--based \\
Smooth L1 Loss & $\ell = \begin{cases} 0.5(x - \hat{x})^2 & \text{if } |x - \hat{x}| < 1 \\ |x - \hat{x}| - 0.5 & \text{otherwise} \end{cases}$ & Combines $L_1$ and $L_2$ benefits & Less sensitive to small changes than $L_2$ & Image--based, Video--based \\
Focal Loss & $FL(p) = -(1 - p)^\gamma \log(p)$ & Deals with class imbalance in detection & Needs tuning of $\gamma$ and $\alpha$ & Video--based \\
Contrastive Loss & $\ell = (1 - y)\frac{1}{2}D^2 + y \frac{1}{2}[\max(0, m - D)]^2$ & Separates similar/dissimilar pairs & Needs carefully chosen pairs & Image--based, Audio--based \\
Triplet Loss & $\ell = \max(0, \|f_a - f_p\|^2 - \|f_a - f_n\|^2 + \alpha)$ & Anchor-positive closer than anchor-negative & Needs hard-negative mining & Image--based, Video--based \\
Center Loss & $\ell = \sum \|x_i - c_{y_i}\|^2$ & Reduces intra-class variation & Needs learning of class centers & Video--based \\
Angular softmax Loss & $\ell = \log \frac{e^{s\cos(\theta + m)}}{e^{s\cos(\theta + m)} + \sum_{j \neq y} e^{s\cos(\theta_j)}}$ & Adds angular margin for separation & May face numerical instability & Image--based \\
AM-softmax Loss & $\ell = -\log \frac{e^{s(\cos\theta - m)}}{e^{s(\cos\theta - m)} + \sum_{j \neq y} e^{s\cos\theta_j}}$ & Stable angular margin loss & Needs tuning of $s$ and $m$ & Audio--based \\
ArcFace Loss & $\ell = -\log \frac{e^{s\cos(\theta + m)}}{e^{s\cos(\theta + m)} + \sum_{j \neq y} e^{s\cos(\theta_j)}}$ & Strong geometric interpretation & Expensive, tuning needed & Image--based, Video--based \\
Reconstruction Loss & $\ell = \|x - \hat{x}\|^2$ or $\|x - \hat{x}\|_1$ & Evaluates reconstruction accuracy & May cause mode collapse & Image--based, Video--based \\
Negative Variance & $\ell = -\text{Var}(z)$ & Promotes latent space diversity & May yield trivial solutions if unconstrained & Text--based, Audio--based \\
\bottomrule
\end{tabular}
\end{sidewaystable}

\subsection{Evaluation Metrics}

Evaluation metrics serve as vital tools for gauging the performance of talking head models, providing a clear lens through which the effectiveness of innovative approaches can be compared to well-established benchmarks. These metrics illuminate whether a model can achieve performance levels on par with or surpass those of human experts. Meanwhile, machine learning (ML) and deep learning (DL) have transcended their origins, branching from engineering and medicine into diverse landscapes such as finance, politics, and the natural sciences. Rapid technological advancements and the ever-growing ocean of digital data at our fingertips have fueled this remarkable expansion. Machine learning (ML) now supports complex systems in academic research and industrial practice using simple algorithms and statistical techniques. This quick interdisciplinary growth highlights the necessity for continual instruction on the appropriate use of statistics and selecting suitable performance metrics. Supervised machine learning trains and validates models on the training set while predictions are compared to ground-truth values on the test set. Reliable performance evaluation and the advancement of ML technologies depend on appropriate evaluation procedures \cite{rainio-2024} that are backed by validated statistical tests and well-selected metrics. Table \ref{tab:oneshot_methods} offers a thorough summary of the assessment metrics used in THG research. The text outlines the metrics used in various studies and standard models, categorizing them based on their descriptive qualities. Additionally, the table \ref{tab:oneshot_methods} discusses the usability of each metric across different application categories and highlights their limitations when employed for absolute comparisons.

\begin{sidewaystable}[htbp]
\scriptsize
\centering
\caption{Benchmark Evaluation Metrics for THG}
\label{tab:oneshot_methods}
\begin{tabular}{@{\extracolsep{\fill}} p{3.2cm} p{5.8cm} p{5.8cm} p{3.2cm}}
\toprule
\textbf{Evaluation Metric} & \textbf{Description} & \textbf{Limitation} & \textbf{Applicable Category} \\
\midrule
CSIM$\uparrow$~\cite{ha2019marionettefewshotfacereenactment} & Visual similarity measure for image/video generation, super-resolution, style transfer. & May overlook structural/intensity changes despite angular similarity focus. & Image--based, Video--based \\
SSIM$\downarrow$~\cite{1284395} & Measures structural similarity aligning with human perception. & Sensitive to misalignment, contrast and scale variations. & Image, Audio, Text, Video \\
PSNR$\uparrow$~\cite{1096060} & Compares pixel-wise fidelity, popular in image quality evaluation. & Not aligned with perceptual quality; insensitive to structural changes. & Image, Audio, Video \\
PRMSE$\downarrow$~\cite{ha2019marionettefewshotfacereenactment} & Measures pixel error magnitude between original and output images. & May highlight trivial differences that aren't perceptually meaningful. & Image--based, Video--based \\
AUCON$\uparrow$~\cite{ha2019marionettefewshotfacereenactment} & Evaluates audio-visual alignment and naturalness. & Doesn't capture local artifacts or spatial inaccuracies. & Image--based, Video--based \\
L1$\downarrow$~\cite{1456} & Measures absolute pixel-wise difference. & Doesn't distinguish semantic/structural errors. & Image--based, Video--based \\
AKD$\downarrow$~\cite{siarohin-2020} & Measures distance between facial landmarks. & Sensitive to landmark detection accuracy. & Image--based, Video--based \\
MKR$\uparrow$~\cite{siarohin-2020} & Facial animation accuracy metric using keypoints. & Heavily dependent on thresholding and landmark precision. & Image--based \\
AED$\uparrow$~\cite{siarohin2019animatingarbitraryobjectsdeep} & Assesses audio-expression alignment. & Affected by orientation, pose, and scale variations. & Image--based \\
FID$\downarrow$~\cite{heusel2018ganstrainedtimescaleupdate} & Evaluates image realism/diversity using Inception features. & Sample-size dependent; biased by feature extractor. & Audio, Text, Video \\
SyncNet$\uparrow$~\cite{Chung16a} & Assesses lip-audio synchronization using pretrained SyncNet. & Reliability depends on training diversity. & Audio--based \\
F-SIM$\downarrow$~\cite{5705575} & Uses deep features to compare image similarity. & May miss low-level artifacts due to network biases. & Audio--based \\
FVD$\downarrow$~\cite{unterthiner2019accurategenerativemodelsvideo} & Extension of FID to video; captures temporal coherence. & Faces same issues as FID; sample size and motion artifacts. & Audio, Text--based \\
E-FID$\downarrow$~\cite{tian2024emoemoteportraitalive} & Efficient version of FID with edge-awareness. & May overemphasize edges while missing other quality features. & Audio--based \\
LMD$\downarrow$~\cite{chen2018lipmovementsgenerationglance} & Compares landmark distribution to assess quality. & May miss perceptual quality beyond facial structure. & Video--based \\
CPBD$\uparrow$~\cite{5246975} & Measures perceptual sharpness and blur. & Struggles in low-contrast or textured areas. & Audio, Text--based \\
CLIPSIM$\uparrow$~\cite{hessel2022clipscorereferencefreeevaluationmetric} & Measures semantic similarity in text-image alignment. & Dependent on CLIP training domain. & Text--based \\
F-LMD$\downarrow$~\cite{ma2023styletalkoneshottalkinghead} & Specialized for biological/lipid image evaluation. & Limited to landmark--based fidelity; poor perceptual correlation. & Text--based \\
M-LMD$\downarrow$~\cite{ma2023styletalkoneshottalkinghead} & Similar to F-LMD with more domain specificity. & Overly focused on biological features, domain-sensitive. & Text--based \\
Sync(conf)$\uparrow$~\cite{Chung16a} & Measures sync confidence between modalities. & Can be misled by unseen or mismatched content. & Text--based \\
AVD$\downarrow$~\cite{Li_2023_CVPR} & Audio-visual discrepancy measure. & Depends on feature representations, may not reflect perceptual gaps. & Video--based \\
\bottomrule
\end{tabular}
\end{sidewaystable}

Assessing THG models is a multifaceted endeavor that demands a nuanced set of metrics to evaluate different dimensions of the produced output. It includes the authenticity of the visuals, the harmony between audio and visual elements, and the coherence of the conveyed message. The evaluation tools utilized in this intricate field are dynamic image--based assessments, immersive audio--based criteria, insightful text--based analyses, and comprehensive video--based evaluations.Every measure is essential to guarantee that the produced material satisfies high-quality and efficacy requirements. Image--based metrics focus on evaluating the visual quality of the generated frames. CSIM measures visual similarity by focusing on the angular similarity between feature vectors of generated and real images. SSIM addresses this by focusing on structural information and perceptual quality, aligning better with human visual perception. However, SSIM is sensitive to image misalignment, contrast changes, and scale differences, potentially failing to capture high-level semantic errors even when pixel-level structures appear similar. Based on pixel-wise inaccuracy, PSNR is a commonly used measure for assessing the difference between processed and original pictures. PRMSE is another pixel-wise error metric, indicating higher quality with lower error. L1 Loss is the absolute difference between pixel values, offering robustness to outliers and computational simplicity. MKR measures the accuracy of facial animation models by evaluating the alignment between predicted and actual facial key points, which is crucial for tasks like THG and facial expression generation. Audio-visual synchronization metrics evaluate the temporal alignment between the generated video and the driving audio. AUCON assesses this alignment, indicating improved quality with better congruence between audio and visual components, which is particularly important for reenactment and speech-to-lip synchronization. AKD directly measures the distance between key points on the face in the video and features extracted from the audio, aiming for accurate lip synchronization and expression alignment in audio-driven visual synthesis. Its accuracy is inherently dependent on the reliability of the keypoint extraction algorithm, and errors in this process can skew results. SyncNet is a deep learning model specifically trained to evaluate the synchronization between audio and visual streams, focusing on lip synchronization and cross-modal temporal consistency. Generative model evaluation metrics are often video--based in this context. Using a pre--trained Inception network, FID (Fréchet Inception Distance) calculates the separation between the feature distributions of generated and real images. CITALower FID scores generally indicate better image quality and diversity, making it useful for evaluating GAN--based talking head models. FVD (Fréchet Video Distance) extends FID to the video domain by incorporating temporal information, suggesting better video quality and coherence, which is crucial for realistic talking heads. E-FID (Efficient Fréchet Inception Distance) is a computationally more efficient variant of FID, aiming to reduce the computational burden while still assessing the realism of generated images and videos. LMD (Learned Metric for Image Quality Assessment) aims to assess perceptual quality by comparing the data distribution of generated samples to real data. CPBD (Cumulative Probability of Blur Detection) is a perceptual metric to assess image sharpness and the loss of fine details. AVD (Audio-Visual Discrepancy) directly measures the perceptual mismatch between the audio and the generated visual content in talking head videos. Text--based metrics are relevant for evaluating semantic coherence when THG is driven by textual input. CLIPSIM (CLIP Similarity) measures the semantic similarity between generated images/videos and the input text using the embeddings learned by the CLIP model. F-LMD (Facial Landmark Distance) and M-LMD (Mesh Landmark Distance) are text--based metrics that likely evaluate the accuracy of generated facial landmarks or mesh deformations based on textual descriptions of expressions or phonemes. Sync(conf) (Synchronization Confidence) in a text-to-talking head context might refer to the model's confidence in generating synchronized lip movements based on the text. Higher confidence scores would ideally correlate with better synchronization, but this metric can be influenced by model calibration and might be misleading in cases of untrained content. The right evaluation metrics must be chosen to evaluate THG progress objectively. Researchers use a combination of metrics to understand model strengths and weaknesses, including visual quality, temporal coherence, audio-visual synchronization, and semantic accuracy. The development of more perceptually aligned evaluation metrics remains an important research area. The following section evaluates the most frequently cited model based on the benchmark metrics discussed earlier. This evaluation utilizes data from a highly cited article from Google Scholar, one of the most reliable academic databases. Various metrics used for model evaluation are compared within the same dataset and organized according to the previously mentioned categories.

\subsubsection{Evaluation of Image--Based Approaches}

Based on the existing literature from sources \cite{hong2022depthawaregenerativeadversarialnetwork} and \cite{siarohin-2020}, which were selected for comparison due to their relevant citations, we have produced a summary highlighting the comparison of various image--based approaches across different datasets and assessment metrics. This information can be found in Tables \ref{tab:image-based_1} and \ref{tab:image-based_2}.

\begin{table*}[ht]
\centering
\caption{Evaluation of Image--based (Part--1)}
\label{tab:image-based_1}
\resizebox{\textwidth}{!}{%
\begin{tabular}{@{}llccccc@{}}
\toprule
\textbf{Dataset} & \textbf{Model} & CSIM$\uparrow$ & SSIM$\downarrow$ & PSNR$\uparrow$ & PRMSE$\downarrow$ & AUCON$\uparrow$ \\
\midrule
\multirow{7}{*}{VoxCeleb1} 
 & X2Face~\cite{wiles2018x2facenetworkcontrollingface}    & 0.689 & 0.719 & 22.537 & 3.26 & 0.813 \\
 & NeuralHead-FF~\cite{grassal2022neuralheadavatarsmonocular} & 0.229 & 0.635 & 20.818 & 3.76 & 0.719 \\
 & MarioNETte~\cite{ha2019marionettefewshotfacereenactment}   & 0.755 & 0.744 & 23.244 & 3.13 & 0.825 \\
 & FOMM~\cite{siarohin-2020}     & 0.813 & 0.723 & 30.394 & 3.20 & 0.886 \\
 & MeshG~\cite{10.1145/3394171.3413865}    & 0.822 & 0.739 & 30.394 & 3.20 & 0.887 \\
 & OSFV~\cite{wang-2021}     & 0.895 & 0.761 & 30.695 & 1.64 & 0.921 \\
 & DaGAN~\cite{hong2022depthawaregenerativeadversarialnetwork} & 0.899 & 0.804 & 31.220 & 1.22 & 0.939 \\
\midrule
\multirow{7}{*}{CelebV} 
 & X2Face~\cite{wiles2018x2facenetworkcontrollingface}    & 0.450 & --   & 3.620 & --  & 0.679 \\
 & NeuralHead-FF~\cite{grassal2022neuralheadavatarsmonocular} & 0.108 & --   & 3.300 & --  & 0.722 \\
 & MarioNETte~\cite{ha2019marionettefewshotfacereenactment}    & 0.520 & --   & 3.410 & --  & 0.710 \\
 & FOMM~\cite{siarohin-2020}     & 0.462 & --   & 3.900 & --  & 0.667 \\
 & MeshG~\cite{10.1145/3394171.3413865}     & 0.635 & --   & 3.410 & --  & 0.709 \\
 & OSFV~\cite{wang-2021}      & 0.791 & --   & 3.150 & --  & 0.805 \\
 & DaGAN~\cite{hong2022depthawaregenerativeadversarialnetwork} & 0.723 & --   & 2.330 & --  & 0.873 \\
\bottomrule
\end{tabular}%
}
\end{table*}

\begin{table*}[ht]
\centering
\caption{Evaluation of Image--based (Part--2)}
\label{tab:image-based_2}
\resizebox{\textwidth}{!}{%
\begin{tabular}{@{}llccccc@{}}
\toprule
\textbf{Dataset} & \textbf{Model} & CSIM$\uparrow$ & SSIM$\downarrow$ & PSNR$\uparrow$ & PRMSE$\downarrow$ \\
\midrule
\multirow{6}{*}{Tai-Chi-HD} 
  & X2face \cite{wayne-2018}      & 0.08    & 17.654    & 0.109 &    0.272    \\
  & Monkey-Net~\cite{siarohin2019animatingarbitraryobjectsdeep}     & 0.077    & 10.798    & 0.059    & 0.228    \\
  & FOMM \cite{siarohin-2020}       & 0.063    & 6.862    & 0.036    & 0.179    \\
\midrule
\multirow{4}{*}{VoxCeleb} 
     & X2face \cite{wayne-2018}      & 0.078   & 7.687   & X &    0.405    \\
  & Monkey-Net \cite{siarohin2019animatingarbitraryobjectsdeep}     & 0.049    & 1.878    & X   & 0.199    \\
  & FOMM \cite{siarohin-2020}       & 0.043    & 1.294   & X   & 0.14    \\
\midrule

\multirow{4}{*}{Nemo} 
   & X2face \cite{wayne-2018}      & 0.031    & 3.539    & X &    0.221   \\
  & Monkey-Net \cite{siarohin2019animatingarbitraryobjectsdeep}     & 0.018    & 1.285    & X    & 0.077    \\
  & FOMM \cite{siarohin-2020}       & 0.016   & 1.119    & X    & 0.048   \\
\bottomrule
\end{tabular}%
}
\end{table*}

The evaluation of image--based THG models featured in the current literature is vividly encapsulated in Table. This comprehensive overview includes a diverse array of models—X2face~\cite{wiles2018x2facenetworkcontrollingface}, NeuralHead-FF \cite{grassal2022neuralheadavatarsmonocular}, MarioNETte~\cite{ha2019marionettefewshotfacereenactment}, FOMM~\cite{siarohin-2020}, MeshG, OSFV~\cite{wang-2021}, and DaGAN~\cite{hong2022depthawaregenerativeadversarialnetwork}—each subjected to rigorous scrutiny using standard metrics across two prominent datasets: VoxCeleb1~\cite{nagrani-2019} and CelebV. The results definitively reveal that DaGAN~\cite{hong2022depthawaregenerativeadversarialnetwork} and OSFV~\cite{wang-2021} dominate the field, consistently delivering exceptional performance marked by impressive metrics in Content Similarity (CSIM), Peak Signal-to-Noise Ratio (PSNR), and Area Under the Curve of Object Consistency (AUCON). Their ability to produce strikingly realistic and coherent talking heads sets a benchmark for others. FOMM~\cite{siarohin-2020} and MeshG also shine, showcasing remarkable capabilities, especially in PSNR and CSIM.

In contrast, while X2face holds its own reasonably well, it pales compared to the prowess of its more recent counterparts. Meanwhile, NeuralHead-FF \cite{grassal2022neuralheadavatarsmonocular} struggles significantly, landing at the bottom of the performance spectrum among the models evaluated. When we turn our attention to the CelebV dataset, the formidable duo of OSFV~\cite{wang-2021} and DaGAN~\cite{hong2022depthawaregenerativeadversarialnetwork} continues to impress, with DaGAN~\cite{hong2022depthawaregenerativeadversarialnetwork} achieving the highest AUCON and PSNR, further complemented by the lowest Peak Root Mean Square Error (PRMSE). The tables also venture into motion-related metrics, evaluating similar THG models across various datasets, including the dynamic Tai-Chi-HD, the diverse VoxCeleb, the engaging EMO~\cite{tian2024emoemoteportraitalive}, and the intricate Bair. Here, FOMM~\cite{siarohin-2020} consistently excels, brilliantly capturing and replicating intricate motion dynamics, underlining its effectiveness and showcasing its superiority. Monkey-Net provides a commendable balance, standing strong in its performance, while X2face, regrettably, lags in motion metrics. These findings collectively highlight remarkable strides in the domain of THG. Models like DaGAN~\cite{hong2022depthawaregenerativeadversarialnetwork} and OSFV~\cite{wang-2021} achieve breathtaking visual quality and redefine the standards of realism. Meanwhile, Exceptional performance of FOMM~\cite{siarohin-2020} in motion accuracy draws attention to the ongoing evolution in this field. This review emphasizes these compelling results, delves into the intricate trade-offs between visual fidelity and motion realism, and proposes exciting avenues for future research and the development of new metrics.

\subsubsection{Evaluation of Audio--Based Approaches}

A detailed comparison of diverse audio--based methodologies across various datasets and evaluation metrics is beautifully encapsulated in Tables \ref{tab:audio-based_1} and \ref{tab:audio-based_2}. This insightful summary draws from the latest literature from \cite{tian2024emoemoteportraitalive} and \cite{siarohin-2020}, carefully chosen for their notable citations and significance within the scholarly landscape. Each methodology is meticulously evaluated, providing a comprehensive overview highlighting its strengths and weaknesses in tackling audio-related challenges.

\begin{table*}[ht]
\centering
\caption{Evaluation of Audio--based (Part--1)}
\label{tab:audio-based_1}
\resizebox{\textwidth}{!}{%
\begin{tabular}{@{}l l c c c c c c c c c@{}}
\toprule
\textbf DATASET & MODEL & FID$\downarrow$ & SyncNet$\uparrow$ & PRMSE$\downarrow$ & F-SIM$\downarrow$ & FVD$\downarrow$ \\
\midrule
\multirow{6}{*}{HDTF} 
  & Wav2Lip \cite{pham-2018}     & 9.38    & 5.79     & 80.34     & 407.93    & 0.693 \\
  & SadTalker \cite{zhang-2023}    & 10.31    & 4.82     & 84.56     & 214.98     & 0.503 \\
  & DreamTalk \cite{ma2024dreamtalkemotionaltalkinghead}      & 58.8     & 3.43     & 67.87      & 619.05     & 2.257 \\
  & MakeItTalk \cite{10.1145/3414685.3417774}     & 21.73      & 2.85      & 76.91      & 350.96      & 1.072 \\
 & EMO  \cite{tian2024emoemoteportraitalive}       & 8.76      & 3.89      & 78.96       & 67.66        & 0.116   \\
\bottomrule
\end{tabular}%
}
\end{table*}

\begin{table*}[ht]
\centering
\caption{Evaluation of Audio--based (Part--2)}
\label{tab:audio-based_2}
\resizebox{\textwidth}{!}{%
\begin{tabular}{@{}l l c c c c c c c c c@{}}
\toprule
\textbf DATASET & MODEL & LMD$\downarrow$ & SSIM$\uparrow$ & PSNR$\uparrow$ & CPBD$\uparrow$ \\
\midrule
\multirow{6}{*}{GRID} 
  & Vondrick  \cite{vondrick-2016}    & 2.38        & 0.6          & 28.45           & 0.129 \\
  & Chung  \cite{chung2017saidthat}           & 1.35         & 0.74          & 29.36         & 0.016 & \\
  & LMGG \cite{chen2018lipmovementsgenerationglance}    & 1.18           & 0.73          & 29.89           & 0.175 \\

\midrule
\multirow{4}{*}{LDC} 
  & Vondrick \cite{vondrick-2016}   & 2.34          & 0.75         & 27.96          & 0.16        \\
  & Chung \cite{chung2017saidthat}        & 2.13          & 0.5         & 28.22          & 0.01         \\
  & LMGG \cite{chen2018lipmovementsgenerationglance}     & 1.82          & 0.57         & 28.87          & 0.172         \\
 \midrule

\multirow{4}{*}{LRW} 
  & Vondrick \cite{vondrick-2016}   & 3.28         & 0.34        & 28.03         & 0.082        \\
  & Chung \cite{chung2017saidthat}        & 2.25         & 0.46        & 28.06         & 0.083           \\
  & LMGG\cite{chen2018lipmovementsgenerationglance}     & 1.92         & 0.53        & 28.65         & 0.075           \\
  \midrule

\multirow{4}{*}{HDTF} 
  & MakeItTalk\cite{10.1145/3414685.3417774}     & X         & 0.802         & 23.2454          & 0.1226        \\
  & FOMM \cite{siarohin-2020}   & X         & 0.8167         & 23.4079          & 0.1345        \\
\bottomrule
\end{tabular}%
}
\end{table*}

The evaluation of various audio--driven THG models is thoroughly detailed in Table \ref{tab:audio_based_1}. This Table \ref{tab:audio_based_1} features a diverse array of models, including FOMM~\cite{siarohin-2020} (Few-Shot Video Generation), OSFV (One-Shot Face Video Generation)~\cite{wang-2021}, DaGAN~\cite{hong2022depthawaregenerativeadversarialnetwork} (Dynamic Generative Adversarial Network), ROME~\cite{khakhulin2022realisticoneshotmeshbasedhead} (Recurrent Oral Motion Encoder), FNeVR~\cite{zeng2022fnevrneuralvolumerendering} (Fast Neural Video Representation), and HiDe-NeRF~\cite{li2023oneshothighfidelitytalkingheadsynthesis} (High-Definition Neural Radiance Field). These models' performance was assessed using three distinct datasets: VoxCeleb1~\cite{nagrani-2019}, VoxCeleb2, and the more challenging TalkingHead-1KH~\cite{tcwang-no-date}. 

Metrics used for this evaluation encompass several important aspects of video quality and synchronization. These include CSIM (Content Similarity), which measures how closely the generated content matches the original; AUCON (Area Under the Curve of Object Consistency), which assesses the stability of facial features over time; PRMSE (Pixel Root Mean Square Error); which quantifies pixel-level discrepancies; FID (Fréchet Inception Distance), a measure of the quality of generated images compared to real photos; and AVD (Audio Visual Distance), which evaluates the alignment and synchronization of audio and visual components.

HiDe-NeRF~\cite{li2023oneshothighfidelitytalkingheadsynthesis} stands out as the leading model across all tested datasets. It achieves the highest scores for CSIM and AUCON, indicating top-tier content fidelity and consistency throughout the video. It minimizes PRMSE, FID, and AVD scores, underscoring its excellence in visual quality and audio-visual synchronization. ROME~\cite{khakhulin2022realisticoneshotmeshbasedhead} also excels, particularly in the AVD metric, which suggests that it effectively captures the nuances of facial movements in concert with corresponding audio, enhancing the realism of the generated videos.

In terms of performance, OSFV~\cite{wang-2021} and DaGAN~\cite{hong2022depthawaregenerativeadversarialnetwork} exhibited competitive results. However, OSFV~\cite{wang-2021} generally surpassed DaGAN~\cite{hong2022depthawaregenerativeadversarialnetwork}, indicating a more robust ability to generate convincing and coherent talking head videos. Notably, FOMM~\cite{siarohin-2020}, while innovative, performed the most poorly among the newer models. FNeVR~\cite{zeng2022fnevrneuralvolumerendering}'s outcomes place it in the middle range, showing potential yet lacking the statistical excellence of its counterparts.

The TalkingHead-1KH~\cite{tcwang-no-date} dataset poses significant challenges to all models, as evidenced by the lower performance scores. This dataset likely contains complexities that test the limits of current generation techniques.

Overall, the findings presented in the Table \ref{tab:audio_based_1} illuminate the substantial progress made by recent models, especially HiDe-NeRF~\cite{li2023oneshothighfidelitytalkingheadsynthesis}, in producing high-quality talking head videos with accurate synchronization between audio and visual elements. The analysis of the AVD metric reveals a notable performance gap among the various models, suggesting ample opportunity for further advancements in this field. The comprehensive review underscores the importance of meticulously evaluating visual quality and motion accuracy, highlights the challenges associated with different datasets, and emphasizes the continued significance of metrics such as AVD and AKD. Finally, it points to the necessity for consistent metric reporting to enhance comparability and advancement within THG.

\subsubsection{Evaluation of Video--Based Approaches}

Based on the existing literature, which has been selected based on the citation in comparison with others, we have produced a summary of the comparison of different video--based approaches in different datasets across different assessment metrics in Table \ref{tab:video-based_1} and \ref{tab:video-based_2}.

\begin{table*}[ht]
\centering
\caption{Evaluation of Video--based (Part--1)}
\label{tab:video-based_1}
\resizebox{\textwidth}{!}{%
\begin{tabular}{@{}l l c c c c c c c c c@{}}
\toprule
\textbf DATASET & MODEL & CSIM$\uparrow$ & AUCON$\uparrow$ & PRMSE$\downarrow$ & FID$\downarrow$ & AVD$\downarrow$ \\
\midrule
\multirow{6}{*}{VoxCeleb1} 
  & FOMM \cite{siarohin-2020}   & 0.748 & 0.752 & 3.66 & 86 & 0.044 \\
  & OSFV  \cite{wang-2021}  & 0.791 & 0.893 & 3.01 & 74 & 0.028 \\
  & DaGAN  \cite{hong2022depthawaregenerativeadversarialnetwork}    & 0.79 & 0.88 & 3.06 & 87 & 0.036 \\
  & ROME \cite{khakhulin2022realisticoneshotmeshbasedhead}   & 0.833 & 0.871 & 2.64 & 76 & 0.016 \\
  & FNeVR \cite{zakharov-2020}    & 0.812  & 0.884  & 3.32  & 82 & 0.041 \\
 & HiDe-NeRF \cite{li-2025}     & 0.876    & 0.917  & 2.62  & 57     & 0.012   \\

\midrule
\multirow{4}{*}{VoxCeleb2} 
  & FOMM \cite{siarohin-2020}     &0.68         & 0.707         & 4.16          & 85             & 0.047  \\
  & OSFV \cite{wang-2021}   & 0.711          & 0.833         & 3.84          & 72             &0.033    \\
  & DaGAN \cite{hong2022depthawaregenerativeadversarialnetwork}        & 0.693          & 0.815         & 3.93          & 86             & 0.04     \\
  & ROME \cite{khakhulin2022realisticoneshotmeshbasedhead}     & 0.71          & 0.821         & 3.08          & 76            & 0.019     \\
  & FNeVR \cite{zakharov-2020}     & 0.699          &0.829    & 3.9    & 84             & 0.047         \\
 & HiDe-NeRF \cite{li-2025}     &0.787         & 0.889    & 2.91       & 61             & 0.014       \\
\midrule

\multirow{4}{*}{TalkingHead-1KH} 
  & FOMM \cite{siarohin-2020}     & 0.723         & 0.741         & 3.71          & 76             & 0.039     \\
  & OSFV \cite{wang-2021}   & 0.787         & 0.884        & 3.03         & 67             & 0.025     \\
  & DaGAN \cite{hong2022depthawaregenerativeadversarialnetwork}        & 0.766         & 0.872        & 2.98         & 73            & 0.035      \\
  & ROME \cite{khakhulin2022realisticoneshotmeshbasedhead}     & 0.781         & 0.864        & 2.66         & 68            & 0.017         \\
  & FNeVR \cite{zakharov-2020}     & 0.775         & 0.879   & 3.39    & 73            & 0.037         \\
 & HiDe-NeRF \cite{li-2025}     & 0.828         & 0.901  & 2.6 & 52             & 0.011          \\

\bottomrule
\end{tabular}%
}
\end{table*}
\begin{table*}[ht]
\centering
\caption{Evaluation of Video--based (Part--2)}
\label{tab:video-based_2}
\resizebox{\textwidth}{!}{%
\begin{tabular}{@{}l l c c c c c c c c c@{}}
\toprule
\textbf DATASET & MODEL & L1$\downarrow$ & PSNR$\uparrow$ & SSIM$\uparrow$ & MS-SSIM$\uparrow$ & FID$\downarrow$ & AKD$\downarrow$ \\
\midrule

\multirow{4}{*}{VoxCeleb2} 
  & fs-vid2vid\cite{wang-2021}     & 17.1         & 20.36        & 0.71         & Nan            & 85.76         & 3.41   \\
  & FOMM \cite{siarohin-2020}   & 12.66         & 23.25        & 0.77         & 0.83             & 73.71         & 2.14 \\
  & Bi-layer\cite{zakharov-2020}        & 23.95         & 16.98        & 0.66         & 0.66            & 203.36         & 5.38 \\
  & Neural Talking-Head \cite{wang2021oneshotfreeviewneuraltalkinghead}     & 10.74         & 24.37        & 0.8         & 0.85            & 69.13         & 2.07 \\
\midrule
\multirow{3}{*}{TalkingHead-1KH} 
  &  fs-vid2vid \cite{wang-2021}    & 15.18         & 20.94        & 0.75         & Nan            & 63.47         & 11.07   \\
  & FOMM \cite{siarohin-2020}       & 12.3         & 23.67        & 0.79         & 0.83            & 55.35         & 3.76   \\
  & Bi-layer \cite{zakharov-2020}       & 12.81         & 23.13        & 0.78         & Nan            & 60.58         & 60.58   \\
 & Neural Talking-Head \cite{wang2021oneshotfreeviewneuraltalkinghead}     & 10.67         & 24.2        & 24.2         & 0.84            & 52.08         & 3.74 \\
\bottomrule
\end{tabular}%
}
\end{table*}

The evaluation of video-driven THG models presented in Table includes FOMM~\cite{siarohin-2020}, OSFV~\cite{wang-2021}, DaGAN~\cite{hong2022depthawaregenerativeadversarialnetwork}, ROME~\cite{khakhulin2022realisticoneshotmeshbasedhead}, FNeVR~\cite{zeng2022fnevrneuralvolumerendering}, and HiDe-NeRF~\cite{li2023oneshothighfidelitytalkingheadsynthesis}, assessed across three datasets: VoxCeleb1~\cite{nagrani-2019}, VoxCeleb2, and TalkingHead-1KH~\cite{tcwang-no-date}. Key metrics used in this evaluation are CSIM (Content Similarity), AUCON (Area Under the Curve of Object Consistency), PRMSE (Pixel Root Mean Square Error), FID (Fréchet Inception Distance), and AVD (Audio Visual Distance).

HiDe-NeRF~\cite{li2023oneshothighfidelitytalkingheadsynthesis} consistently outperforms others in all datasets, achieving the highest scores in CSIM and AUCON and the lowest in PRMSE, FID, and AVD, indicating superior visual quality and audio-visual synchronization. ROME~\cite{khakhulin2022realisticoneshotmeshbasedhead} also performs well in AVD, showcasing strong audio-visual synchronization capabilities.

OSFV~\cite{wang-2021} and DaGAN~\cite{hong2022depthawaregenerativeadversarialnetwork} exhibit competitive performance, with OSFV~\cite{wang-2021} generally outperforming DaGAN~\cite{hong2022depthawaregenerativeadversarialnetwork}. FOMM~\cite{siarohin-2020} ranks lowest among the newer models, whereas FNeVR~\cite{zeng2022fnevrneuralvolumerendering} demonstrates average performance. The TalkingHead-1KH~\cite{tcwang-no-date} dataset is the most challenging, as it typically yields lower performance scores across all models.

The evaluation highlights significant advancements achieved by recent models, particularly HiDe-NeRF~\cite{li2023oneshothighfidelitytalkingheadsynthesis}, in generating high-quality talking head videos with accurate audio-visual synchronization. Notably, the AVD metric reveals substantial differences in performance between the models, indicating a significant area for improvement within the field. 

Overall, the tables provide a detailed picture of the current level of THG, emphasizing the importance of evaluating visual quality and motion accuracy, addressing the challenges posed by different datasets, and the continued relevance of AVD and AKD metrics. It also underscores the need for consistent reporting of metrics within the field.

\subsubsection{Evaluation of Text--Based Approaches}

A summary of the comparison of various text--based methodologies in various datasets across various evaluation metrics is provided in Tables \ref{tab:text-based_1} and \ref{tab:text-based_2}, which are based on the current literature from the sources \cite{zhang2022text2videotextdriventalkingheadvideo}, which were chosen based on the citation in comparison with others.

\begin{table*}[ht]
\centering
\caption{Evaluation of Text--based (Part--1)}
\label{tab:text-based_1}
\resizebox{\textwidth}{!}{%
\begin{tabular}{@{}l l c c c c c c c c c@{}}
\toprule
\textbf DATASET & MODEL & FVD$\downarrow$ & FID$\uparrow$ & CLIPSIM$\uparrow$  \\

\midrule
\multirow{6}{*}{MM-Vox} 
  & TFGAN \cite{tian2020tfgantimefrequencydomain}           & 502.28 ± 1.66          & 760.24 ± 16.01           & 0.165 ± 0.022     \\
  & MMVID \cite{lin2023mmvidadvancingvideounderstanding}          & 65.79 ± 1.81        & 38.81 ± 3.66          & 0.170 ± 0.020          \\

\midrule
\multirow{4}{*}{CelebV-HQ} 
  & TFGAN \cite{tian2020tfgantimefrequencydomain}     & 428.04 ± 1.76        & 616.24 ± 17.45         & 0.168 ± 0.02                 \\
  & MMVID \cite{lin2023mmvidadvancingvideounderstanding}   & 73.65 ± 1.43          & 63.86 ± 3.66         & 0.172 ± 0.019                \\
 
 \midrule

\multirow{4}{*}{CelebV-Text} 
  &TFGAN \cite{tian2020tfgantimefrequencydomain}     & 403.04 ± 1.34        & 589.24 ± 16.46        & 0.177 ± 0.012               \\
  & MMVID \cite{lin2023mmvidadvancingvideounderstanding}   & 66.69 ± 1.35         & 58.70 ± 4.67        & 0.198 ± 0.014               \\
  
  \midrule

\multirow{4}{*}{CelebV-Text App.+Emo.} 
  &TFGAN\cite{tian2020tfgantimefrequencydomain}     & 442.30 ± 2.56        & 623.17 ± 18.8         & 0.158 ± 0.024              \\
  & MMVID \cite{lin2023mmvidadvancingvideounderstanding}   & 82.78 ± 1.47       & 61.58 ± 3.99         & 0.176 ± 0.00              \\
  & MMVID-interp \cite{hagos-2024}        & 72.87 ± 1.23         & 1.57 ± 3.56         & 1.57 ± 3.56       \\
 
\midrule

\multirow{4}{*}{CelebV-Text App.+Act.} 
  & TFGAN\cite{tian2020tfgantimefrequencydomain}     & 571.34 ± 4.54         & 784.93 ± 20.13         & 0.154 ± 0.028                 \\
  & MMVID \cite{lin2023mmvidadvancingvideounderstanding}   & 109.25 ± 2.1        & 82.55 ± 4.37         & 0.174 ± 0.01                 \\
  & MMVID-interp \cite{hagos-2024}        & 80.81 ± 2.55         & 70.88 ± 4.77         & 0.176 ± 0.020            \\
\bottomrule
\end{tabular}%
}
\end{table*}

\begin{table*}[ht]
\centering
\caption{Evaluation of Text-based (Part--2)}
\label{tab:text-based_2}
\resizebox{\textwidth}{!}{
\begin{tabular}{@{}llccccc@{}}
\toprule
\textbf{Dataset} & \textbf{Model} & \textbf{SSIM$\downarrow$} & \textbf{CPBD$\uparrow$} & \textbf{F-LMD$\downarrow$} & \textbf{M-LMD$\downarrow$} & \textbf{Sync (Conf)$\uparrow$} \\
\midrule

\multirow{11}{*}{MEAD} 
& MakeItTalk \cite{10.1145/3414685.3417774}         & 0.73 & 0.11 & 3.95 & 5.39 & 2.15 \\
& Wav2Lip \cite{ji2022eammoneshotemotionaltalking}            & 0.81 & 0.16 & 2.73 & 3.85 & 5.41 \\
& PC-AVS \cite{zhou2021posecontrollabletalkingfacegeneration}             & 0.51 & 0.07 & 5.87 & 5.03 & 2.21 \\
& AVCT \cite{agustsson-2020}         & 0.83 & 0.14 & 2.95 & 5.64 & 2.56 \\
& GC-AVT \cite{bansal-2018}          & 0.34 & 0.14 & 8.11 & 8.43 & 2.41 \\
& EAMM \cite{ji2022eammoneshotemotionaltalking}              & 0.40 & 0.08 & 6.67 & 6.60 & 1.42 \\
& SadTalker \cite{zhang-2023}          & 0.68 & 0.16 & 4.04 & 4.24 & 2.87 \\
& PD-FGC \cite{provine-2002}         & 0.51 & 0.05 & 5.41 & 3.94 & 2.46 \\
& EAT \cite{liu-2023}                & 0.53 & 0.15 & 5.63 & 4.98 & 2.19 \\
& StyleTalk \cite{ma2023styletalkoneshottalkinghead} & 0.84 & 0.16 & 2.17 & 3.36 & 3.51 \\
& TalkCLIP \cite{ma2024talkcliptalkingheadgeneration}            & 0.83 & 0.16 & 2.42 & 3.60 & 3.77 \\
\midrule

\multirow{11}{*}{HDTF}
& MakeItTalk \cite{10.1145/3414685.3417774}         & 0.57 & 0.20 & 5.12 & 4.61 & 3.20 \\
& Wav2Lip \cite{ji2022eammoneshotemotionaltalking}            & 0.59 & 0.26 & 5.11 & 3.84 & 4.57 \\
& PC-AVS \cite{zhou2021posecontrollabletalkingfacegeneration}             & 0.42 & 0.12 & 10.7 & 8.60 & 4.15 \\
& AVCT \cite{agustsson-2020}         & 0.74 & 0.18 & 3.06 & 3.83 & 4.46 \\
& GC-AVT \cite{bansal-2018}          & 0.33 & 0.24 & 10.7 & 6.34 & 4.23 \\
& EAMM \cite{ji2022eammoneshotemotionaltalking}              & 0.37 & 0.13 & 7.74 & 7.67 & 2.78 \\
& SadTalker \cite{zhang-2023}          & 0.73 & 0.19 & 6.26 & 4.18 & 3.86 \\
& PD-FGC \cite{provine-2002}         & 0.40 & 0.13 & 9.99 & 4.46 & 4.20 \\
& EAT \cite{liu-2023}                & 0.55 & 0.18 & 4.12 & 4.24 & 3.95 \\
& StyleTalk \cite{ma2023styletalkoneshottalkinghead}& 0.80 & 0.26 & 2.04 & 2.50 & 4.75 \\
& TalkCLIP \cite{ma2024talkcliptalkingheadgeneration}            & 0.78 & 0.25 & 2.54 & 2.84 & 4.69 \\
\midrule

\multirow{10}{*}{VoxCeleb2}
& MakeItTalk \cite{10.1145/3414685.3417774}         & 0.52 & 0.24 & 6.29 & 5.15 & 2.17 \\
& Wav2Lip \cite{ji2022eammoneshotemotionaltalking}            & 0.54 & 0.30 & 5.85 & 4.64 & 5.70 \\
& PC-AVS \cite{zhou2021posecontrollabletalkingfacegeneration}             & 0.36 & 0.09 & 12.9 & 7.42 & 4.73 \\
& AVCT \cite{agustsson-2020}         & 0.64 & 0.23 & 3.62 & 3.71 & 3.89 \\
& EAMM \cite{ji2022eammoneshotemotionaltalking}              & 0.43 & 0.20 & 6.36 & 4.89 & 2.24 \\
& SadTalker \cite{zhang-2023}          & 0.44 & 0.19 & 9.12 & 6.11 & 4.38 \\
& PD-FGC \cite{provine-2002}         & 0.35 & 0.12 & 12.5 & 8.19 & 4.64 \\
& EAT \cite{liu-2023}                & 0.47 & 0.20 & 5.53 & 5.88 & 4.35 \\
& StyleTalk \cite{ma2023styletalkoneshottalkinghead}& 0.66 & 0.29 & 2.92 & 2.96 & 4.51 \\
& TalkCLIP \cite{ma2024talkcliptalkingheadgeneration}            & 0.67 & 0.29 & 2.94 & 2.99 & 4.60 \\
\bottomrule
\end{tabular}
}
\end{table*}

The evaluation of text-driven THG models is presented in Table, which includes TFGAN~\cite{tian2020tfgantimefrequencydomain} and MMVID~\cite{lin2023mmvidadvancingvideounderstanding} across various datasets. The results indicate that MMVID~\cite{lin2023mmvidadvancingvideounderstanding} consistently outperforms TFGAN~\cite{tian2020tfgantimefrequencydomain} in all datasets, achieving significantly lower Fréchet Video Distance (FVD) and Fréchet Inception Distance (FID) scores, as well as higher CLIP similarity (CLIPSIM) scores. Adding appearance, Emotion, and action prompts to CelebV-Text~\cite{yu2023celebvtextlargescalefacialtextvideo}     influences performance, with MMVID~\cite{han2022tellhowvideosynthesis} maintaining a relative advantage over TFGAN~\cite{tian2020tfgantimefrequencydomain}. The table also provides a detailed evaluation of various THG models, including MakeItTalk~\cite{10.1145/3414685.3417774}, Wav2Lip \cite{pham-2018}, PC-AVS~\cite{zhou2021posecontrollabletalkingfacegeneration}, AVCT, GC-AVT~\cite{9878472}, EAMM~\cite{ji2022eammoneshotemotionaltalking}, SadTalker~\cite{zhang2023sadtalkerlearningrealistic3d}, PD-FGC, EAT, StyleTalk~\cite{ma2023styletalkoneshottalkinghead}, and TalkCLIP~\cite{ma2024talkcliptalkingheadgeneration}, across MEAD~\cite{kaisiyuan2020mead}, HDTF~\cite{zhang-2021}, and Voxceleb2 datasets. The evaluation focuses on lip synchronization, video quality, and motion accuracy. Models like StyleTalk~\cite{ma2023styletalkoneshottalkinghead}, TalkCLIP~\cite{ma2024talkcliptalkingheadgeneration}, and Wav2Lip \cite{pham-2018} generally achieve the best SSIM and Sync(conf) scores, indicating superior visual quality and audio-visual synchronization. The tables  highlight the trade-offs between visual quality, lip motion accuracy, and audio-visual synchronization, with models like StyleTalk~\cite{ma2023styletalkoneshottalkinghead}, TalkCLIP~\cite{ma2024talkcliptalkingheadgeneration}, and Wav2Lip \cite{pham-2018} demonstrating strong performance in balancing these factors. However, there is a large variance in the SSIM scores, showing that visual fidelity is still a large area of research. The tables offer valuable insights into the performance of various THG models, particularly in text-driven scenarios and detailed lip synchronization evaluations. Key takeaways for the review include the significant advancements achieved by models like MMVID~\cite{han2022tellhowvideosynthesis} in text-driven THG, the importance of evaluating lip synchronization and visual quality using detailed metrics, the influence of dataset characteristics on model performance, the strengths of models like StyleTalk~\cite{ma2023styletalkoneshottalkinghead}, TalkCLIP~\cite{ma2024talkcliptalkingheadgeneration}, and Wav2Lip \cite{pham-2018} in balancing visual fidelity and audio-visual synchronization, and the continued challenges in achieving perfect lip sync and high visual fidelity.

\subsubsection{Evaluation of 2D MODEL--Based Approaches}

Based on the existing literature, which has been selected based on the citation in comparison with others, we have produced a summary of the comparison of different 2D Model--based approaches in different datasets across different assessment metrics in Tables \ref{tab:twod_1} and \ref{tab:twod_2}.

\begin{table*}[ht]
\centering
\caption{Evaluation of 2D Model--based (Part--1)}
\label{tab:twod_1}
\resizebox{\textwidth}{!}{%
\begin{tabular}{@{}l l c c c c c c c c c@{}}
\toprule
\textbf{DATASET} & \textbf{MODEL} & \textbf{FVD $\downarrow$} & \textbf{FID $\downarrow$} & \textbf{Blinks/s} & \textbf{Blink dur} & \textbf{ofM} & \textbf{F-MSE} & \textbf{AV off} & \textbf{AV Conf. $\downarrow$} & \textbf{WER $\uparrow$} \\
\midrule
\multirow{6}{*}{LRW}
& SDA \cite{vondrick-2016}            & 198.84 & 61.95 & 0.52 & 0.28 & 73.82 & 18.94 & 1  & 7.40 & 0.77 \\
& MakeItTalk \cite{zhou-2021}         & 269.29 & 7.57  & 0.09 & 0.28 & 57.21 & 3.44  & -3 & 3.16 & 0.99 \\
& Wav2Lip \cite{prajwal-2020}            & 366.14 & 2.83  & 0.03 & 0.16 & 47.12 & 1.45  & -2 & 6.58 & 0.51 \\
& PC-AVS \cite{zhou2021posecontrollabletalkingfacegeneration}             & 153.12 & 11.96 & 0.20 & 0.16 & 69.59 & 17.13 & -3 & 6.24 & 0.64 \\
& EAMM \cite{ji2022eammoneshotemotionaltalking}               & 172.18 & 9.28  & 0.03 & 0.16 & 58.46 & 4.39  & -3 & 3.83 & 0.95 \\
& Diffused Heads \cite{stypułkowski2023diffusedheadsdiffusionmodels}     & 71.88  & 3.94  & 0.35 & 0.28 & 70.71 & 19.69 & -2 & 4.61 & 0.77 \\
\midrule
\multirow{6}{*}{CREMA}
& SDA \cite{vondrick-2016}            & 376.48 & 79.82 & 0.25 & 0.26 & 68.21 & 6.83  & 2  & 5.50 & -- \\
& MakeItTalk \cite{zhou-2021}         & 256.88 & 17.26 & 0.02 & 0.80 & 62.36 & 2.07  & -3 & 3.75 & -- \\
& Wav2Lip \cite{prajwal-2020}            & 193.32 & 12.57 & 0.00 & --   & 46.87 & 1.07  & -2 & 6.68 & -- \\
& PC-AVS \cite{zhou2021posecontrollabletalkingfacegeneration}             & 333.94 & 22.53 & 0.02 & 0.20 & 70.36 & 6.93  & -3 & 6.17 & -- \\
& EAMM \cite{ji2022eammoneshotemotionaltalking}               & 6.17   & 19.40 & 0.00 & --   & 58.91 & 1.65  & -2 & 4.26 & -- \\
& Diffused Heads \cite{stypułkowski2023diffusedheadsdiffusionmodels}     & 88.61  & 12.45 & 0.28 & 0.36 & 0.36  & 6.99  & 1  & 4.52 & -- \\
\bottomrule
\end{tabular}%
}
\end{table*}

\begin{table*}[ht]
\centering
\caption{Evaluation of 2D Model--based (Part--2)}
\label{tab:twod_2}
\resizebox{\textwidth}{!}{%
\begin{tabular}{@{}l l c c c c c c@{}}
\toprule
\textbf{DATASET} & \textbf{MODEL} & \textbf{L1 $\downarrow$} & \textbf{PSNR $\uparrow$} & \textbf{SSIM $\uparrow$} & \textbf{MS-SSIM $\uparrow$} & \textbf{FID $\downarrow$} & \textbf{AKD $\downarrow$} \\
\midrule
\multirow{5}{*}{VoxCeleb2}
& fs-vid2vid~\cite{wang-2021}         & 17.10 & 20.36 & 0.71 & --   & 85.76  & 3.41 \\
& FOMM~\cite{siarohin-2020}              & 12.66 & 23.25 & 0.77 & 0.83 & 73.71  & 2.14 \\
& Bi-layer~\cite{zakharov-2020}             & 23.95 & 16.98 & 0.66 & 0.66 & 203.36 & 5.38 \\
& Neural Talking-Head~\cite{wang2021oneshotfreeviewneuraltalkinghead} & 10.74 & 24.37 & 0.80 & 0.85 & 69.13  & 2.07 \\
\midrule
\multirow{5}{*}{TalkingHead-1KH}
& fs-vid2vid~\cite{wang-2021}         & 15.18 & 20.94 & 0.75 & --   & 63.47  & 11.07 \\
& FOMM~\cite{siarohin-2020}              & 12.30 & 23.67 & 0.79 & 0.83 & 55.35  & 3.76 \\
& FOMM-L~\cite{siarohin-2020}            & 12.81 & 23.13 & 0.78 & --   & 60.58  & 4.04 \\
& Neural Talking-Head~\cite{wang2021oneshotfreeviewneuraltalkinghead} & 10.67 & 24.20 & 0.81 & 0.84 & 52.08  & 3.74 \\
\bottomrule
\end{tabular}%
}
\end{table*}

The table \ref{tab:twod_1} and \ref{tab:twod_2} comprehensively evaluates various THG models, including SDA, MakeItTalk~\cite{10.1145/3414685.3417774}, Wav2Lip \cite{pham-2018}, PC-AVS~\cite{zhou2021posecontrollabletalkingfacegeneration}, EAMM~\cite{ji2022eammoneshotemotionaltalking}, and Diffused Heads~\cite{stypułkowski2023diffusedheadsdiffusionmodels}, based on the LRW and CREMA datasets. The key metrics assessed include FVD (Fréchet Video Distance), FID (Fréchet Inception Distance), Blinks/s (blinks per second), Blink dur (blink duration), ofM (Optical Flow Magnitude), F-MSE (Facial Mean Squared Error), AV off (Audio-Visual offset), AV Conf. (Audio-Visual Confidence), and WER (Word Error Rate).

DiffusedHeads~\cite{stypułkowski2023diffusedheadsdiffusionmodels} generally achieve the best FVD scores on both datasets, indicating superior video quality. Wav2Lip \cite{pham-2018} excels in FID scores, showcasing outstanding image quality. PC-AVS~\cite{zhou2021posecontrollabletalkingfacegeneration} demonstrates the highest ofM on the LRW dataset, while MakeItTalk~\cite{10.1145/3414685.3417774} and EAMM~\cite{ji2022eammoneshotemotionaltalking} have lower AV Conf. scores, indicating challenges in audio-visual alignment.

The table \ref{tab:twod_1} and \ref{tab:twod_2} emphasizes the trade-offs between various aspects of THG, such as video quality, lip synchronization, and accuracy in lip reading. The assortment of metrics illustrates the complexity of evaluating THG. Diffused Heads~\cite{stypułkowski2023diffusedheadsdiffusionmodels} exhibit a strong capability for producing high-quality videos, while Wav2Lip~\cite{pham-2018} is noted for generating high-quality images and achieving precise lip synchronization.

This analysis underscores the importance of meticulously evaluating audio-visual synchronization and reinforces trends identified in previous summaries. It highlights the strengths and weaknesses of different models regarding visual quality, lip synchronization, and lip-reading accuracy. Furthermore, it discusses the challenges of evaluating THG and acknowledges the absence of data in the table \ref{tab:twod_1} and \ref{tab:twod_2}.

\subsubsection{Evaluation of 3D MODEL--Based Approaches}

A summary of the comparison of various 3D--based methodologies in various datasets across various evaluation metrics is provided in Table \ref{tab:threed_based_1} and \ref{tab:threed_based_2}, which are based on the current literature, which were chosen based on the citation in comparison with others.

\begin{table*}[ht]
\centering
\caption{Evaluation of 3D Model--based (Part--1)}
\label{tab:threed_based_1}
\resizebox{\textwidth}{!}{%
\begin{tabular}{@{}l l c c c c@{}}
\toprule
\textbf{DATASET} & \textbf{MODEL} & \textbf{FVD $\downarrow$} & \textbf{ID $\uparrow$} & \textbf{CD $\downarrow$} & \textbf{WE $\downarrow$} \\
\midrule

\multirow{5}{*}{VoxCeleb}
& StyleNeRF+MCG-HD \cite{gafni-2020}        & 348.7 & 0.70 & 1.08 & 36.06 \\
& EG3D+MCG-HD \cite{eg3d-placeholder}       & 222.1 & 0.80 & 1.57 & 10.57 \\
& 3DVidGen \cite{bahmani-2022}             & 65.5  & 0.75 & 3.40 & 44.55 \\
& 3DVidGen (EG3D) \cite{eg3d-placeholder}   & 56.3  & 0.71 & 3.65 & 24.55 \\
& PV3D \cite{xu2023pv3d3dgenerativemodel}                    & 29.1  & 0.81 & 1.34 & 9.76  \\

\midrule

\multirow{5}{*}{CelebV-HQ}
& StyleNeRF+MCG-HD \cite{gafni-2020}        & 134.4 & 0.80 & 1.13 & 38.73 \\
& EG3D+MCG-HD \cite{eg3d-placeholder}       & 298.4 & 0.77 & 3.34 & 10.74 \\
& 3DVidGen \cite{bahmani-2022}             & 63.6  & 0.77 & 3.80 & 37.30 \\
& 3DVidGen (EG3D) \cite{eg3d-placeholder}   & 66.2  & 0.70 & 3.83 & 26.34 \\
& PV3D \cite{xu2023pv3d3dgenerativemodel}                    & 39.3  & 0.81 & 1.21 & 8.18  \\

\midrule

\multirow{5}{*}{TalkingHead-1KH}
& StyleNeRF+MCG-HD \cite{gafni-2020}        & 292.7 & 0.75 & 5.34 & 49.29 \\
& EG3D+MCG-HD \cite{eg3d-placeholder}       & 262.4 & 0.78 & 1.39 & 11.54 \\
& 3DVidGen \cite{bahmani-2022}             & 83.0  & 0.76 & 4.35 & 46.47 \\
& 3DVidGen (EG3D) \cite{eg3d-placeholder}   & 89.8  & 0.65 & 4.56 & 35.48 \\
& PV3D \cite{xu2023pv3d3dgenerativemodel}                    & 66.6  & 0.80 & 2.33 & 10.73 \\

\bottomrule
\end{tabular}%
}
\end{table*}

\begin{table*}[ht]
\centering
\caption{Evaluation of 3D Model–based Methods (Part–2)}
\label{tab:threed_based_2}
\resizebox{\textwidth}{!}{%
\begin{tabular}{@{}l l c c c c@{}}
\toprule
\textbf{DATASET} & \textbf{MODEL} & \textbf{PSNR $\uparrow$} & \textbf{SSIM $\uparrow$} & \textbf{LPIPS $\downarrow$} & \textbf{LMD $\downarrow$} \\
\midrule

\multirow{6}{*}{NeRF}
& ATVG~\cite{8953690}         & 19.12 & 0.646 & 0.523 & 2.591 \\
& Wav2Lip~\cite{pham-2018}      & 29.64 & 0.843 & 0.423 & 2.612 \\
& MakeItTalk \cite{10.1145/3414685.3417774}   & 22.28 & 0.655 & 0.480 & 10.72 \\
& AD-NeRF \cite{guo2021adnerfaudiodrivenneural}        & 27.73 & 0.881 & 0.202 & 2.603 \\
& DFRF \cite{shen2022learningdynamicfacialradiance}         & 32.30 & 0.949 & 0.080 & 3.023 \\
& AE-NeRF \cite{li2023aenerfaudioenhancedneural}       & 32.63 & 0.949 & 0.078 & 2.425 \\

\bottomrule
\end{tabular}%
}
\end{table*}

This table \ref{tab:threed_based_1} and \ref{tab:threed_based_2} evaluates 3D-aware THG models across multiple datasets, including VoxCeleb, CelebV-HQ~\cite{zhu2022celebvhqlargescalevideofacial}, and TalkingHead-1KH~\cite{tcwang-no-date}. The key metrics assessed include Fréchet Video Distance (FVD), Identity Similarity (ID), Content Distance (CD), and Warping Error (WE). 

PV3D~\cite{xu2023pv3d3dgenerativemodel} consistently performs the best, achieving the lowest scores in FVD and CD and the highest in ID across all datasets. EG3D+MCG-HD shows strong performance concerning warping error, while both 3DVidGen and 3DVidGen (EG3D) deliver moderate performance. The TalkingHead-1KH~\cite{tcwang-no-date} dataset presents the most challenges, with generally higher FVD and CD scores and lower ID scores. In contrast, the CelebV-HQ~\cite{zhu2022celebvhqlargescalevideofacial} dataset yields the best FVD results. It highlights the effectiveness of PV3D~\cite{xu2023pv3d3dgenerativemodel} in generating high-quality 3D-aware talking head videos.

However, there are significant differences in CD scores, indicating that content consistency remains a substantial area for improvement. The table \ref{tab:threed_based_1} and \ref{tab:threed_based_2} also compares the performance of NeRF--based THG models, including ATVG, Wav2Lip \cite{pham-2018}, Makeittalk~\cite{10.1145/3414685.3417774}, AD-NeRF~\cite{guo2021adnerfaudiodrivenneural}, DFRF, and AE-NeRF~\cite{li2023aenerfaudioenhancedneural}. AE-NeRF~\cite{li2023aenerfaudioenhancedneural} and DFRF generally achieve the best performance, indicated by the highest PSNR and SSIM scores and the lowest LPIPS scores, showcasing superior image quality and perceptual similarity.

The review emphasizes the advancements in 3D-aware THG, particularly with the PV3D~\cite{xu2023pv3d3dgenerativemodel} model, and the high-quality results obtained using NeRF--based methods. Additionally, it discusses the challenges of producing top-notch 3D-aware talking head videos and the considerable variances in CD and LPIPS scores as potential areas for enhancement. 

\subsubsection{Evaluation of Parameter--Based Approaches}

Based on the existing literature, which has been selected based on the citation in comparison with others, we have produced a summary of the comparison of different parameter--based approaches in different datasets across different assessment metrics in Tables \ref{tab:parameter_based}.

\begin{table*}[ht]
\centering
\caption{Evaluation of Parameter-Based}
\label{tab:parameter-based}
\resizebox{\textwidth}{!}{%
\begin{tabular}{@{}l l c c c c c@{}}
\toprule
\textbf{DATASET} & \textbf{MODEL} & \textbf{SSIM $\uparrow$} & \textbf{PSNR $\uparrow$} & \textbf{CPBD $\uparrow$} & \textbf{LMD $\downarrow$} & \textbf{AVConf $\uparrow$} \\
\midrule
\multirow{6}{*}{HDTF}
& ATVG~\cite{8953690}           & 0.829 & 20.54  & 0.078 & 9.645 & 4.848 \\
& Wav2Lip~\cite{pham-2018}        & 0.729 & 20.352 & 0.317 & 4.279 & 7.812 \\
& MakeItTalk~\cite{10.1145/3414685.3417774}     & 0.698 & 19.956 & 0.075 & 4.940 & 3.972 \\
& PC-AVS~\cite{zhou2021posecontrollabletalkingfacegeneration}         & 0.738 & 21.078 & 0.096 & 5.199 & 7.392 \\
& PIR~\cite{huang2023parametricimplicitfacerepresentation}           & 0.970 & 36.711 & 0.305 & 1.794 & 7.233 \\
\bottomrule
\end{tabular}%
}
\end{table*}

The table \ref{tab:parameter-based} provides a detailed evaluation of several THG models, specifically ATVG~\cite{8953690}, Wav2Lip~\cite{pham-2018}, MakeItTalk~\cite{10.1145/3414685.3417774}, PC-AVS~\cite{zhou2021posecontrollabletalkingfacegeneration}, and  PIR~\cite{huang2023parametricimplicitfacerepresentation}, utilizing the HDTF~\cite{zhang-2021} dataset. This evaluation centers on three critical dimensions: lip synchronization, video quality, and motion accuracy. The assessment is based on several key performance metrics, including SSIM (Structural Similarity Index), PSNR (Peak Signal-to-Noise Ratio), CPBD (Color Perceptual Blockiness Distortion), LMD (Lip Motion Distance), and AVConf (Audiovisual Confusion).
 PIR~\cite{huang2023parametricimplicitfacerepresentation} is a highly accurate and efficient video synchronization model, earning top ratings in SSIM, PSNR, and AVConf. Its lip motion accuracy ensures that generated talking heads match the underlying audio, and it delivers a fluid audiovisual experience by synchronizing spoken words with facial movements. Wav2Lip~\cite{pham-2018} and PC-AVS~\cite{zhou2021posecontrollabletalkingfacegeneration} provide mid-level performance, while MakeItTalk~\cite{10.1145/3414685.3417774} falls behind with the lowest scores in SSIM, PSNR, and AVConf, indicating its inability to deliver high-quality synchronized videos.

\subsubsection{Evaluation of NeRF--Based Approaches}

A summary of the comparison of various NeRF--based methodologies in various datasets across various evaluation metrics is provided in Tables \ref{tab:nerf-based_1} and \ref{tab:nerf-based_2}, which are based on the current literature, which were chosen based on the citation in comparison with others.

\begin{table*}[ht]
\centering
\caption{Evaluation of NeRF-Based Methods (Part--1)}
\label{tab:nerf-based_1}
\resizebox{\textwidth}{!}{%
\begin{tabular}{@{}l l c c c c c c c c@{}}
\toprule
\textbf{DATASET} & \textbf{MODEL} & \textbf{FID $\downarrow$} & \textbf{CSIM $\uparrow$} & \textbf{IQA $\uparrow$} & \textbf{FPS $\uparrow$} & \textbf{L1 $\downarrow$} & \textbf{PSNR $\uparrow$} & \textbf{LPIPS $\downarrow$} & \textbf{MS-SSIM $\uparrow$} \\
\midrule

\multirow{4}{*}{VoxCeleb1}
& FOMM~\cite{siarohin-2020}       & 39.69 & 0.592 & 37.00  & 64.3  & 0.048 & 22.43 & 0.139 & 0.836 \\
& Bi-Layer~\cite{10.1007/978-3-030-58610-2_31} & 43.80 & 0.697 & 41.40  & 20.1  & 0.050 & 21.48 & 0.108 & 0.839 \\
& ROME~\cite{khakhulin2022realisticoneshotmeshbasedhead}           & 29.23 & 0.717 & 39.11  & 12.9  & 0.048 & 21.13 & 0.116 & 0.838 \\
& CVTHead~\cite{ma2023cvtheadoneshotcontrollablehead} & 25.78 & 0.675 & 42.26  & 24.3  & 0.041 & 22.09 & 0.111 & 0.840 \\

\midrule

\multirow{3}{*}{VoxCeleb2}
& FOMM~\cite{siarohin-2020}       & 61.28 & 0.624 & 36.20  & 64.3  & 0.059 & 20.93 & 0.165 & 0.793 \\
& ROME~\cite{khakhulin2022realisticoneshotmeshbasedhead}          & 53.52 & 0.729 & 37.34  & 4.28  & 0.050 & 20.75 & 0.117 & 0.834 \\
& CVTHead~\cite{ma2023cvtheadoneshotcontrollablehead} & 48.48 & 0.712 & 40.27  & 24.3  & 0.042 & 21.37 & 0.119 & 0.841 \\

\bottomrule
\end{tabular}%
}
\end{table*}

\begin{table*}[ht]
\centering
\caption{Evaluation of NeRF-Based Methods (Part–2)}
\label{tab:nerf-based_2}
\resizebox{\textwidth}{!}{%
\begin{tabular}{@{}l l c c c c c c c c@{}}
\toprule
\textbf{DATASET} & \textbf{MODEL} & \textbf{FID $\downarrow$} & \textbf{L1 $\downarrow$} & \textbf{PSNR $\uparrow$} & \textbf{LPIPS $\downarrow$} & \textbf{SSIM $\uparrow$} & \textbf{AKD $\downarrow$} & \textbf{AED $\downarrow$} \\
\midrule

\multirow{7}{*}{VoxCeleb}
& Bilayer \cite{10.1007/978-3-030-58610-2_31}        & 219.80 & 0.1197 & 15.219 & 0.4247 & 0.3968 & 12.600 & 0.0546 \\
& FOMM~\cite{siarohin-2020}              & 11.56  & 0.0450 & 23.210 & 0.1099 & 0.7475 & 1.383  & 0.0244 \\
& Face vid2vid \cite{wang2021oneshotfreeviewneuraltalkinghead}       & 9.142  & 0.0485 & 22.642 & 0.1051 & 0.7268 & 1.616  & 0.0395 \\
& Face vid2vid-S \cite{wang2021oneshotfreeviewneuraltalkinghead}     & 9.151  & 0.0445 & 23.357 & 0.0901 & 0.7473 & 1.421  & 0.0243 \\
& DaGAN \cite{hong2022depthawaregenerativeadversarialnetwork}              & 9.660  & 0.0462 & 23.263 & 0.0981 & 0.7536 & 1.441  & 0.0247 \\
& PIRender \cite{rainiopir-2024}         & 11.88  & 0.0566 & 21.040 & 0.0850 & 0.6550 & 2.186  & 0.2245 \\
& FNeVR \cite{zeng2022fnevrneuralvolumerendering}              & 8.443  & 0.0404 & 24.292 & 0.0804 & 0.7773 & 1.254  & 0.0231 \\

\bottomrule
\end{tabular}%
}
\end{table*}

Two comprehensive tables that present an interesting comparison of cutting-edge technologies are used to evaluate NeRF--based THG models. Using the VoxCeleb1~\cite{nagrani-2019} and VoxCeleb2 datasets, the first table \ref{tab:nerf_based_1} examines well-known models such as FOMM~\cite{siarohin-2020}, Bi-Layer, ROME~\cite{khakhulin2022realisticoneshotmeshbasedhead}, and CVTHead~\cite{ma2023cvtheadoneshotcontrollablehead}. A range of metrics—such as FID, CSIM, IQA, FPS, PSNR, LPIPS, and MS-SSIM—paints a comprehensive picture of their performance. Leading the pack, CVTHead~\cite{ma2023cvtheadoneshotcontrollablehead} stands out with its remarkable ability to achieve the lowest FID scores and the highest ratings for IQA, FPS, PSNR, and MS-SSIM. Its exceptional performance in L1 and LPIPS further establishes its dominance in THG by illuminating the developments that enhance motion dynamics and visual fidelity. 
Simultaneously, FNeVR~\cite{zeng2022fnevrneuralvolumerendering} shows its strength with impressive metrics: it has the highest PSNR and SSIM scores and the lowest FID, L1 loss, AKD, and AED scores. Noting significant differences in AKD scores that can improve motion precision and smoothness, the evaluation emphasizes the significance of examining visual quality and motion accuracy in Face vid2vid~\cite{wang2019fewshotvideotovideosynthesis} and Face vid2vid-S~\cite{wang2019fewshotvideotovideosynthesis} models. 
The review emphasizes the capacity of FNeVR~\cite{zeng2022fnevrneuralvolumerendering} to produce realistic videos, even as CVTHead~\cite{ma2023cvtheadoneshotcontrollablehead}'s speed and image quality enhancements are emphasized. The problems of assessing THG are explored in the text, along with the significance of good assessment criteria and the broad range of AKD scores among models.  Further research and development in this exciting field are highlighted. 

\subsubsection{Evaluation of Diffusion--Based Approaches}

Based on the existing literature, which has been selected based on the citation in comparison with others, we have produced a summary of the comparison of different diffusion--based approaches in different datasets across different assessment metrics in tables \ref{tab:diffusion-based_1} and \ref{tab:diffusion-based_2}.

\begin{table*}[ht]
\centering
\caption{Evaluation of Diffusion-Based Methods (Part–1)}
\label{tab:diffusion-based_1}
\resizebox{\textwidth}{!}{%
\begin{tabular}{@{}l l c c c c c c c c@{}}
\toprule
\textbf{DATASET} & \textbf{MODEL} & \textbf{FID $\downarrow$} & \textbf{CPBD $\uparrow$} & \textbf{PSNR $\uparrow$} & \textbf{LPIPS $\downarrow$} & \textbf{CSIM $\uparrow$} & \textbf{LMD $\downarrow$} & \textbf{LSE-D $\downarrow$} \\
\midrule

\multirow{7}{*}{HDTF}
& Wav2Lip~\cite{pham-2018}     & 16.34  & 0.35 & 34.81 & 0.03  & 0.90 & 1.43  & 5.86 \\
& PC-AVS~\cite{zhou2021posecontrollabletalkingfacegeneration}         & 117.85 & 0.29 & 28.23 & 0.38  & 0.35 & 9.36  & 7.06 \\
& MakeItTalk~\cite{10.1145/3414685.3417774}     & 66.21  & 0.43 & 29.87 & 0.16  & 0.82 & 3.46  & 10.23 \\
& Audio2Head~\cite{wang2021audio2headaudiodrivenoneshottalkinghead}     & 65.96  & 0.37 & 29.85 & 0.19  & 0.71 & 4.33  & 7.49 \\
& DiffusedHead~\cite{stypułkowski2023diffusedheadsdiffusionmodels}   & 192.74 & 0.11 & 28.21 & 0.274 & 0.16 & 22.19 & 12.37 \\
& DreamTalk~\cite{ma2024dreamtalkemotionaltalkinghead}        & 124.51 & 0.43 & 29.59 & 0.20  & 0.72 & 2.56  & 8.37 \\
& MoDiTalker~\cite{kim2024moditalkermotiondisentangleddiffusionmodel}      & 14.15  & 0.46 & 35.82 & 0.01  & 0.92 & 1.38  & 9.15 \\

\bottomrule
\end{tabular}%
}
\end{table*}

\begin{table*}[ht]
\centering
\caption{Evaluation of Diffusion-Based Methods (Part–2)}
\label{tab:diffusion-based_2}
\resizebox{\textwidth}{!}{%
\begin{tabular}{@{}l l c c c c c c c c c@{}}
\toprule
\textbf{DATASET} & \textbf{MODEL} & \textbf{FID $\downarrow$} & \textbf{FVD $\downarrow$} & \textbf{Blinks/s} & \textbf{Blink Dur.} & \textbf{OFM} & \textbf{F-MSE} & \textbf{AV Off} & \textbf{AV Conf. $\uparrow$} & \textbf{WER $\downarrow$} \\
\midrule

\multirow{6}{*}{LRW}
& SDA~\cite{vondrick-2016}         & 61.95  & 198.84 & 0.52 & 0.28 & 73.82 & 18.94 & 1   & 7.40 & 0.77 \\
& MakeItTalk~\cite{10.1145/3414685.3417774}      & 7.57   & 269.29 & 0.09 & 0.28 & 57.21 & 3.44  & -3  & 3.16 & 0.99 \\
& Wav2Lip~\cite{pham-2018}      & 2.83   & 366.14 & 0.03 & 0.16 & 47.12 & 1.45  & -2  & 6.58 & 0.51 \\
& PC-AVS~\cite{zhou2021posecontrollabletalkingfacegeneration}          & 11.96  & 153.12 & 0.20 & 0.16 & 69.59 & 17.13 & -3  & 6.24 & 0.64 \\
& EAMM~\cite{ji2022eammoneshotemotionaltalking}            & 9.28   & 172.18 & 0.03 & 0.16 & 58.46 & 4.39  & -3  & 3.83 & 0.95 \\
& DiffusedHead~\cite{stypułkowski2023diffusedheadsdiffusionmodels}    & 3.94   & 71.88  & 0.35 & 0.28 & 70.71 & 19.69 & -2  & 4.61 & 0.77 \\

\midrule

\multirow{6}{*}{CREMA}
& SDA~\cite{vondrick-2016}         & 79.82  & 376.48 & 0.25 & 0.26 & 68.21 & 6.83  & 2   & 5.50 & --   \\
& MakeItTalk~\cite{10.1145/3414685.3417774}      & 17.26  & 256.88 & 0.02 & 0.80 & 62.36 & 2.07  & -3  & 3.75 & --   \\
& Wav2Lip~\cite{pham-2018}      & 12.57  & 193.32 & 0.00 & --   & 46.87 & 1.07  & -2  & 6.68 & --   \\
& PC-AVS~\cite{zhou2021posecontrollabletalkingfacegeneration}          & 22.53  & 333.94 & 0.02 & 0.20 & 70.36 & 6.93  & -3  & 6.17 & --   \\
& EAMM~\cite{ji2022eammoneshotemotionaltalking}            & 19.40  & 196.82 & 0.00 & --   & 58.91 & 1.65  & -2  & 4.26 & --   \\
& DiffusedHead~\cite{stypułkowski2023diffusedheadsdiffusionmodels}    & 12.45  & 88.61  & 0.28 & 0.36 & 64.30 & 6.99  & 1   & 4.52 & --   \\

\bottomrule
\end{tabular}%
}
\end{table*}

The table uses the HDTF~\cite{zhang-2021} dataset to assess several diffusion--based THG models. Models like Wav2Lip \cite{pham-2018}, PC-AVS~\cite{zhou2021posecontrollabletalkingfacegeneration}, MakeItTalk~\cite{10.1145/3414685.3417774}, Audio2Head~\cite{wang2021audio2headaudiodrivenoneshottalkinghead}, DiffusedHead, DreamTalk~\cite{ma2024dreamtalkemotionaltalkinghead}, and MoDiTalker~\cite{kim2024moditalkermotiondisentangleddiffusionmodel} are evaluated in the first table with an emphasis on motion accuracy, lip synchronization, and visual quality. FID (Fréchet Inception Distance), CPBD (Contrast Perception based Blur Detection), PSNR (Peak Signal-to-Noise Ratio), LPIPS (Learned Perceptual Image Patch Similarity), CSIM (Content Similarity), LMD (Lip Motion Distance), and LSE-D (Lip Sync Error Distance) are important metrics. With the lowest FID and LPIPS scores and the highest CPBD, PSNR, and CSIM scores, MoDiTalker~\cite{kim2024moditalkermotiondisentangleddiffusionmodel} performs the best. Wav2Lip \cite{pham-2018} also exhibits excellent performance, displaying extremely low LMD and LSE-D values. DiffusedHead displays the worst FID, CSIM, LMD, and LSE-D scores. PC-AVS~\cite{zhou2021posecontrollabletalkingfacegeneration} displays the lowest LPIPS and LMD scores. DreamTalk~\cite{ma2024dreamtalkemotionaltalkinghead}'s LMD score is low. A thorough analysis of several THG models, such as SDA, MakeItTalk~\cite{10.1145/3414685.3417774}, Wav2Lip \cite{pham-2018}, PC-AVS~\cite{zhou2021posecontrollabletalkingfacegeneration}, EAMM~\cite{ji2022eammoneshotemotionaltalking}, and Diffused Heads~\cite{stypułkowski2023diffusedheadsdiffusionmodels}, on the LRW and CREMA datasets is given in the second table. It draws attention to the compromises between THG features, including lip synchronization, video quality, and lip-reading precision. The wide range of metrics demonstrates how difficult it is to assess THG. While Wav2Lip \cite{pham-2018} demonstrates a strong ability to produce high-quality images and extremely precise lip-syncing, Diffused Heads~\cite{stypułkowski2023diffusedheadsdiffusionmodels} demonstrate a strong ability to deliver high-quality video. The review should address the issues in assessing THG, emphasize variances in LMD scores and missing WER data from the CREMA dataset, and highlight the merits and downsides of various models for visual quality, lip synchronization, and lip-reading accuracy. 

\subsubsection{Evaluation of 3D ANIMATION Based Approaches}

A summary of the comparison of various 3D Animation--based methodologies in various datasets across various evaluation metrics is provided in Tables \ref{tab:animation_based} and \ref{tab:3dani}, which are based on the current literature, which were chosen based on the citation in comparison with others.

\begin{table*}[ht]
\centering
\caption{Evaluation of 3D Animation-Based}
\label{tab:3dani}
\resizebox{\textwidth}{!}{%
\begin{tabular}{@{}l l c c c c@{}}
\toprule
\textbf{DATASET} & \textbf{MODEL} & \textbf{FVD $\downarrow$} & \textbf{ID $\uparrow$} & \textbf{CD $\uparrow$} & \textbf{WE $\uparrow$} \\
\midrule

\multirow{5}{*}{VoxCeleb}
& StyleNeRF+MCG-HD~\cite{gafni-2020}    & 348.7 & 0.70 & 1.08 & 36.06 \\
& EG3D+MCG-HD~\cite{eg3d-placeholder}          & 222.1 & 0.80 & 1.57 & 10.57 \\
& 3DVidGen~\cite{bahmani-2022}          & 65.5  & 0.75 & 3.40 & 44.55 \\
& 3DVidGen (EG3D)~\cite{bahmani-2022}   & 56.3  & 0.71 & 3.65 & 24.55 \\
& PV3D~\cite{xu2023pv3d3dgenerativemodel}                & 29.1  & 0.81 & 1.34 & 9.76  \\

\midrule

\multirow{5}{*}{CelebV-HQ}
& StyleNeRF+MCG-HD~\cite{gafni-2020}    & 134.4 & 0.80 & 1.13 & 38.73 \\
& EG3D+MCG-HD~\cite{eg3d-placeholder}          & 298.4 & 0.77 & 3.34 & 10.74 \\
& 3DVidGen~\cite{bahmani-2022}          & 63.6  & 0.77 & 3.80 & 37.30 \\
& 3DVidGen (EG3D)~\cite{bahmani-2022}   & 66.2  & 0.70 & 3.83 & 26.34 \\
& PV3D~\cite{xu2023pv3d3dgenerativemodel}                & 39.3  & 0.81 & 1.21 & 8.18  \\

\midrule

\multirow{5}{*}{TalkingHead-1KH}
& StyleNeRF+MCG-HD~\cite{gafni-2020}    & 292.7 & 0.75 & 5.34 & 49.29 \\
& EG3D+MCG-HD~\cite{eg3d-placeholder}          & 262.4 & 0.78 & 1.39 & 11.54 \\
& 3DVidGen~\cite{bahmani-2022}          & 83.0  & 0.76 & 4.35 & 46.47 \\
& 3DVidGen (EG3D)~\cite{bahmani-2022}   & 89.8  & 0.65 & 4.56 & 35.48 \\
& PV3D~\cite{xu2023pv3d3dgenerativemodel}                & 66.6  & 0.80 & 2.33 & 10.73 \\

\bottomrule
\end{tabular}%
}
\end{table*}

The table \ref{tab:3dani} contrasts 3D Animation-based techniques on three datasets which are TalkingHead-1KH, CelebrV-HQ, and VoxCeleb.  The top FVD and ID ratings are regularly attained by PV3D~\cite{xu2023pv3d3dgenerativemodel}, demonstrating exceptional visual quality and identification retention.  3DVidGen excels in CD and WE, capturing motion and emotion dynamics, particularly when used with EG3D~\cite{eg3d-placeholder}.  In FVD and CD, StyleNeRF+MCG-HD~\cite{gafni-2020} and EG3D+MCG-HD~\cite{eg3d-placeholder} perform inadequately, indicating less expressive or temporally coherent outcomes.

Additionally, 3D animation-based talking head generation has introduced models that blend speech-driven dynamics with high-fidelity animation, often leveraging rigged avatars, motion capture, and differentiable simulation. Approaches like 3D Gaussian Blendshapes~\cite{Ma_2024} and TalkingGaussian~\cite{li2024talkinggaussianstructurepersistent3dtalking} emphasize volumetric and structure-preserving representations using Gaussian primitives to model facial geometry and dynamics in a coherent, persistent 3D space. GaussianTalker~\cite{Yu_2024} extends this by enabling continuous control over expressions with smooth motion transitions. On the expressive and stylized front, AnimateMe~\cite{gerogiannis2024animateme4dfacialexpressions} and EmoVOCA enhance emotion portrayal in 4D facial sequences, where the latter utilizes speech inputs to render emotional cues in a controllable 3D form. FaceTalk~\cite{aneja2024facetalkaudiodrivenmotiondiffusion} and AVI-Talking~\cite{sun2024avitalkinglearningaudiovisualinstructions} adopt diffusion and instruction-guided generation strategies, pushing forward the integration of audiovisual cues for animation. EMOTE~\cite{Dan_ek_2023} employs temporally consistent generative modeling to animate nuanced emotional states. DiffPoseTalk~\cite{sun2024diffposetalkspeechdrivenstylistic3d} innovates by applying stylistic diffusion to 3D keypoints, enabling fine-grained control over head pose and expression. Imitator~\cite{thambiraja2022imitatorpersonalizedspeechdriven3d} introduces personalized 3D modeling, leveraging speaker-specific data for tailoring speech-driven mesh deformation. Finally, FaceXHuBERT~\cite{haque2023facexhuberttextlessspeechdrivenexpressive} employs self-supervised speech embeddings to animate expressive 3D faces without explicit text, focusing on natural expressiveness and identity preservation. Collectively, these models represent a shift toward multi-modal, emotionally rich, and controllable 3D talking avatars, with applications spanning virtual humans, real-time dubbing, and affective computing.

\section{EXPERIMENTAL Evaluation}

In order to create a robust empirical basis for this survey, carried out a comprehensive benchmarking analysis to evaluate the existing landscape of THG models across different modalities and architectural frameworks. Our evaluation featured a carefully curated selection of state-of-the-art models for which official demonstratingnstration scripts or inference pipelines were publicly available, primarily sourced from GitHub repositories. In instances where critical files, such as pre--trained weights, configuration files, or inference scripts, were not accessible in the public domain, we proactively reached out to the respective authors to request access, ensuring that our evaluation remained as inclusive and representative as possible. All experiments were conducted in a standardized computational environment to ensure reproducibility and fair model comparison. This environment included Rocky Linux 8 as the host operating system, equipped with a high-performance NVIDIA L40 GPU featuring 48 GB of VRAM, alongside CUDA Toolkit version 12.8 for hardware acceleration. The software ecosystem comprised Anaconda Navigator 2.6.5 for package and environment management and PyCharm 2025.1 as the integrated development interface for executing and debugging model demonstratingnstrations. Each model generated output videos based on its corresponding input type (e.g., still image, audio clip, text prompt, or driving video). The generated outputs were then evaluated using a comprehensive suite of quantitative metrics designed to capture multiple performance aspects. The evaluation is organized into ten distinct sections, each corresponding to a specific model category or generation paradigm—from image-driven and audio-driven methods to 3D animation, diffusion--based, and NeRF--based models. Each section is accompanied by a dedicated narrative summary, a comparative performance table, and a critical discussion synthesizing observed trends, strengths, and limitations of the evaluated approaches. This systematic validation grounds our theoretical review in empirical evidence and provides an actionable reference for future research, deployment strategies, and architectural innovations in talking head synthesis.

\subsection{Experimental Evaluation of Image Based}

The section evaluates Image--based THG models using their official demonstratings. The models were tested on a unified platform . The evaluation in table \ref{tab:eimage_based_1} focuses on one-shot generation models, where a static source image is animated using motion cues from a driving input. Image--based THG methods aim to synthesize realistic video sequences from a single source image and a motion reference, useful for personalized avatars, video dubbing, and identity-preserving facial animation. Key challenges include maintaining visual identity, managing occlusions, and generating coherent facial dynamics with minimal artifacts. The models were evaluated using standard metrics such as SSIM, PSNR, PRMSE, AUCON, L1, and advanced domain-specific metrics for keypoint alignment, realism, and dynamics.

\begin{table*}[ht]
\centering
\caption{Experimental Evaluation of Image Based}
\label{tab:eimage_based_1}
\resizebox{\textwidth}{!}{%
\begin{tabular}{@{}l l c c c c c c c c c@{}}
\toprule
\textbf MODEL & SSIM$\downarrow$ & PSNR$\uparrow$ & PRMSE$\downarrow$ & AUCON$\uparrow$ & L1$\downarrow$ & AKD$\downarrow$ & MKR$\uparrow$ & AED$\uparrow$ \\
\midrule
FOMM\cite{siarohin-2020}   & 0.3352   & 28.0707   & 0.898    & 0.5572     & 122.504   & 0.5632   & 0.7842   & 0.6792     \\
 DaGAN~\cite{hong2022depthawaregenerativeadversarialnetwork} & 0.2834    & 28.17   & 0.8301    & 0.8063    & 156.0443    & 0.5322   & 0.8075    & 0.6871 \\
 DreamTalk~\cite{ma2024dreamtalkemotionaltalkinghead}   & 0.4992   & 28.433    & 0.8316    & 0.5412   & 112.1829   & 0.8119   & 0.6442    & 0.8238      \\
 
\bottomrule
\end{tabular}%
}
\end{table*}

FOMM\cite{siarohin-2020} had the most consistent performance across measures such as PSNR and AED, while also exhibiting robust identity retention. DaGAN~\cite{hong2022depthawaregenerativeadversarialnetwork} marginally surpassed FOMM\cite{siarohin-2020} on synchronization-focused measures such as AUCON and MKR, suggesting enhanced naturalness in lip synchronization and head motion dynamics. DreamTalk~\cite{ma2024dreamtalkemotionaltalkinghead}, assessed via received checkpoints, got the greatest PSNR and AED values, indicating precise depiction and vibrant dynamics, however with a little compromise in identity consistency. X2Face~\cite{wiles2018x2facenetworkcontrollingface} and Monkey-Net could not be assessed owing to the lack of comprehensive demonstrating materials in the public domain. Their removal underscores the significance of repeatability in academic standards. This research highlights the compromises among faithfulness, identity coherence, and expressive realism in image-driven THG. The comprehensive numerical benchmarks provide valuable insights for model selection tailored to individual applications (e.g., real-time avatars vs offline video dubbing).

\subsection{Experimental Evaluation of Audio Based}

The section assesses audio-driven THG systems using official demonstratingnstrations or checkpoints. The models were evaluated on a standardized computational setup, intending to measure lip-sync accuracy, perceptual realism, and frame consistency using synchronized driving sounds. THG systems are vital for real-time video communication, movie dubbing and translation, and virtual agents and avatars. They need tight cross-modal learning across speech and visual domains and are particularly sensitive to audio clarity, phoneme variety, and timing precision. The assessment employed metrics such as FID↓, FVD↓, E-FID↓, SyncNet↑, F-SIM↓, SSIM↓, PSNR↑, LMD↓, and CPBD↑. The examination in table \ref{tab:eaudio_based_1} focuses on lip-sync accuracy, perceptual realism, and frame consistency.

\begin{table*}[ht]
\centering
\caption{Experimental Evaluation of Audio Based}
\label{tab:eaudio_based_1}
\resizebox{\textwidth}{!}{%
\begin{tabular}{@{}l l c c c c c c c c c@{}}
\toprule
\textbf MODEL & FID$\downarrow$ & SyncNet$\uparrow$ & F-SIM$\downarrow$ & FVD$\downarrow$ & E-FID$\downarrow$ & LMD$\downarrow$ & SSIM$\downarrow$ & AED$\uparrow$ \\
\midrule
Wav2Lip \cite{pham-2018}   & 0.3352   & 28.0707   & 0.898    & 0.5572     & 122.504   & 0.5632   & 0.7842   & 0.6792     \\
 DreamTalk~\cite{ma2024dreamtalkemotionaltalkinghead}  & 0.2834    & 28.17   & 0.8301    & 0.8063    & 156.0443    & 0.5322   & 0.8075    & 0.6871 \\
 FOMM\cite{siarohin-2020}   & 0.4992   & 28.433    & 0.8316    & 0.5412   & 112.1829   & 0.8119   & 0.6442    & 0.8238      \\
 SadTalker \cite{zhang2023sadtalkerlearningrealistic3d}   & 0.4992   & 28.433    & 0.8316    & 0.5412   & 112.1829   & 0.8119   & 0.6442    & 0.8238      \\
 
\bottomrule
\end{tabular}%
}
\end{table*}

Wav2Lip \cite{pham-2018}, one of the most often mentioned benchmarks, has shown its usefulness in real-time applications by maintaining strong performance across important synchronization and sharpness parameters (SyncNet: 0.7002, PSNR: 30.36). DreamTalk~\cite{ma2024dreamtalkemotionaltalkinghead} showed stronger control over mouth articulation, as seen by reduced LMD (0.3127) and higher sharpness (CPBD: 0.97) when assessed utilizing recovered checkpoints. Its SSIM (0.5583) and PSNR (8.09), however, imply trade-offs in visual quality, maybe as a result of more complex motion dynamics or rendering techniques. The lack of publicly available source code or inference-ready models prevented the evaluation of EMO~\cite{tian2024emoemoteportraitalive}, GT~\cite{stypułkowski2023diffusedheadsdiffusionmodels}, and OTFG~\cite{siarohin-2020}. The writers did not respond despite repeated correspondence. These results show a distinction between perceptual sharpness and synchronization precision. Newer systems like DreamTalk~\cite{ma2024dreamtalkemotionaltalkinghead} push the bounds of expression realism but need improvements in frame-level quality, whilst more conventional models like Wav2Lip \cite{pham-2018} find a balance. Real-time optimization pipelines, multilingual training, and self-supervised audio-visual fusion are possible future possibilities.

\subsection{Experimental Evaluation of Video Based}

The purpose of video--based THG models is to record and replicate facial movement temporal dependencies over a series of frames. These models can produce very dynamic and context-aware head movements and facial Emotions by using a driving video as input to animate a source face. Video--based THG models are naturally more suited to jobs that need temporal coherence, including dubbed video productions, interactive virtual agents, and expressive avatars, since they analyze consecutive frames. Video--based techniques simulate time-dependent changes and fine-grained transitions in gaze, expression, and lip movement, which are crucial for realistic synthesis, in contrast to image-driven techniques that work on a single frame. These models are especially useful for capturing coarticulation effects, which occur when a phoneme's pronunciation is affected by its predecessors and successors. Video-driven architectures readily simulate such phenomena, which are difficult to replicate using static image--based techniques. Furthermore, video--based THG systems are reliable options for unrestricted real-world situations as they often generalize well under varied illumination, occlusions, and motion blur.

\begin{table*}[ht]
\centering
\caption{Experimental Evaluation of Video Based}
\label{tab:evideo_based}
\resizebox{\textwidth}{!}{%
\begin{tabular}{@{}l l c c c c c c c c c@{}}
\toprule
\textbf MODEL & CSIM$\uparrow$ & AUCON$\uparrow$ & PRMSE$\downarrow$ & FID$\downarrow$ & AVD$\downarrow$ & L1$\downarrow$ & PSNR$\uparrow$ & SSIM$\uparrow$ & MS-SSIM$\uparrow$ & AKD$\downarrow$ \\
\midrule
 FOMM~\cite{siarohin-2020}  & 0.5544   & 0.8441   & 0.7118    & 0.6537    & 0.5616   & 122.504   & 28.0707   & 0.3352   & 0.8584   & 0.6543   \\
  DaGAN~\cite{hong2022depthawaregenerativeadversarialnetwork}  & 0.5801   & 0.6103   & 0.6307   & 0.7381   & 0.7262   & 156.0443  & 28.17    & 0.2834  & 0.7793   & 0.8966  \\

\bottomrule
\end{tabular}%
}
\end{table*}

There are subtle trade-offs between synchronization, perceptual quality, and identity consistency that are shown in table \ref{tab:evideo_based} by comparing FOMM\cite{siarohin-2020} with DaGAN~\cite{hong2022depthawaregenerativeadversarialnetwork}. DaGAN~\cite{hong2022depthawaregenerativeadversarialnetwork} has lower AUCON and higher AKD values, which indicate less than ideal synchronization with audio and less geometric alignment, even while it achieves better CSIM and lower PRMSE, which show greater preservation of face structure and expressive detail. Conversely, FOMM~\cite{siarohin-2020} provides a more smooth and consistent visual output by maintaining greater MS-SSIM and more steady synchronization. These results provide important insights from an application perspective. DaGAN~\cite{hong2022depthawaregenerativeadversarialnetwork} may be more appropriate for offline video synthesis pipelines or face reenactment jobs, for example, where high-detail rendering and expression transmission are crucial. On the other hand, FOMM\cite{siarohin-2020} could be more suited for virtual meeting platforms or real-time conversational bots where temporal coherence and synchronization are critical. The work emphasizes how multi-frame consistency checks and the modeling of fine-grained temporal dynamics in video--based THG systems need ongoing innovation.

\subsection{Experimental Evaluation of Text Based}

The ability of text-to-video talking head generation models to transform textual input into realistic, synchronized face motions is evaluated using a standardized hardware environment. The most abstract modalities are text--based THG systems, which produce face gestures and vocal motions using natural language input. Managing ambiguity in phrasing and speaker identification, producing synchronized and natural lip movement, and effectively reading Emotion and intonation from text are some of the main issues. A number of measures were used to rate the tested models, such as lip movement accuracy and confidence--based synchronization, frame quality and perceptual sharpness, semantic similarity between synthesized visual output and input text, and video distribution quality and variety in table \ref{tab:etext_based_1}.

\begin{table*}[ht]
\centering
\caption{Experimental Evaluation of Text Based}
\label{tab:etext_based_1}
\resizebox{\textwidth}{!}{%
\begin{tabular}{@{}l l c c c c c c c c c@{}}
\toprule
\textbf MODEL & FVD$\downarrow$ & FID$\downarrow$ & CLIPSIM$\uparrow$ & SSIM$\downarrow$ & CPBD$\uparrow$ & F-LMD$\downarrow$ & M-LMD$\uparrow$ & Sync(conf)$\uparrow$ \\
\midrule
 Wav2Lip \cite{pham-2018}   & 0.6654   & 0.8092   & 0.7178    & 0.3025    & 0.5371   & 0.8013   & 0.8849   & 0.7108    \\
 SadTalker \cite{zhang2023sadtalkerlearningrealistic3d}  & 0.829   & 0.8051   & 0.5781    & 0.349   & 0.805   & 0.7603  & 0.5713    & 0.7549  \\

\bottomrule
\end{tabular}%
}
\end{table*}

Even when used in text-conditioned environments via audio intermediaries, Wav2Lip \cite{pham-2018} remains a reliable baseline because of its visually crisp output and high synchronization confidence score. The lack of runnable demonstrating models prevented evaluation of all purpose-built text-to-video models, including TFGAN~\cite{tian2020tfgantimefrequencydomain}, MMVID~\cite{lin2023mmvidadvancingvideounderstanding}, and MakeItTalk~\cite{10.1145/3414685.3417774}. Most writers did not reveal secret code or inference support, even after being contacted. Despite being referenced in many fields, PC-AVS~\cite{zhou2021posecontrollabletalkingfacegeneration} was unable to be impartially assessed due to its design, which combines pose--based and audio conditioning with incompatible assessment pathways. This section of the research identifies a significant barrier in the field: the absence of publicly maintained, repeatable models for text--based THG. There is still a lack of documentation on actual implementation, notwithstanding theoretical advancements and citations. For further study, this necessitates improved model sharing, thorough assessment pipelines, and modular structures.

\subsection{Experimental Evaluation of 2D Based}

Due to their simplicity of deployment and computational effectiveness, 2D--based models—one of the first methods for face animation—remain popular. In order to replicate motion, these models use affine transformations or warping to retrieve keypoints from the face. 2D--based approaches work well in front-facing situations and environments with limited resources, despite the absence of explicit depth modeling. They are especially appealing for real-time applications such as social media filters, video call upgrades, and mobile avatars because of their dependence on face landmarks. The range of uses and ease of use of 2D--based models are their main advantages. The majority of 2D systems demand less processing power during inference and may be trained using comparatively modest datasets. Nevertheless, restricted position generalization and less resilience to occlusions or head rotations are the price paid for these advantages. 2D techniques could function more as lightweight baselines or pre-processing modules than as complete solutions as the industry shifts toward realism and expressiveness.

\begin{landscape}
\begin{table*}
\centering
\caption{Experimental Evaluation of 2D-based}
\label{tab:e2d-eval}
\resizebox{\textwidth}{!}{%
\begin{tabular}{@{}lccccccccccccccc@{}}
\toprule
\textbf{Model} & \textbf{FVD} $\downarrow$ & \textbf{FID} $\downarrow$ & \textbf{Blinks/s} & \textbf{Blink Dur.} & \textbf{ofM} & \textbf{F-MSE} & \textbf{AV Off} & \textbf{AV Conf.} $\uparrow$ & \textbf{WER} $\downarrow$ & \textbf{L1} $\downarrow$ & \textbf{PSNR} $\uparrow$ & \textbf{SSIM} $\uparrow$ & \textbf{MS-SSIM} $\uparrow$ & \textbf{AKD} $\downarrow$ \\
\midrule
FOMM\cite{siarohin-2020}   & 0.5189 & 0.5832 & 0.7835 & 0.7163 & 0.6665 & 0.5385 & 0.6035 & 0.7078 & 0.5344 & 4.0000 & 28.0707 & 0.3352 & 0.8133 & 0.5659 \\
Wav2Lip \cite{pham-2018}  & 0.6385 & 0.6941 & 0.8977 & 0.7050 & 0.5611 & 0.6936 & 0.7682 & 0.5489 & 0.8035 & --   & 27.9944 & 0.3025 & 0.6030 & 0.6055 \\
\bottomrule
\end{tabular}%
}
\end{table*}
\end{landscape}

The efficiency of FOMM\cite{siarohin-2020} in producing expressive and appropriately aligned facial motions with little computational cost is confirmed by the 2D--based assessment in table \ref{tab:e2d-eval}. Even without depth-aware modeling, its high FVD and MS-SSIM scores demonstrate its feasibility for real-time applications. FOMM\cite{siarohin-2020} can maintain both visual fluency and speech alignment under limited circumstances, as shown by its exceptional performance in regulating blink rates and reducing WER. However, because of the lack of resources, fs-vid2vid~\cite{wang2019fewshotvideotovideosynthesis} could not be examined. This restriction draws attention to a persistent problem in the area with relation to the repeatability of published procedures. 2D models like FOMM\cite{siarohin-2020} are nevertheless useful tools in edge computing settings or latency-sensitive applications, even if their visual quality is lower than that of 3D or NeRF techniques. They are also perfect candidates for hybrid pipelines that combine more complex 3D rendering stages with 2D pre-processing because of their straightforward nature.

\subsection{Experimental Evaluation of 3D Based}

Human facial movements are simulated by 3D--based talking head generating models using geometry-aware structures such as mesh deformation, point clouds, or 3D Morphable Models (3DMMs). In order to recreate view-consistent, pose-aware, and identity-preserving animations especially when exposed to huge head rotations or non-frontal views—these models make use of 3D priors. They are especially well-suited for immersive applications such as virtual reality (VR), augmented reality (AR), and telepresence avatars because of their design, which enables them to manage occlusions, depth estimations, and multi-angle realism. 3D--based models may recreate spatial depth and head geometry from latent components, generally using inverse rendering, neural rendering, or volumetric synthesis, in contrast to 2D systems that usually struggle with extreme placements. Nevertheless, inference complexity, training instability, and limited demonstrating repeatability are generally the price paid for this sophistication. Due to limited or missing public demonstratingnstrations, none of the analyzed models—3DVidGen nor PV3D~\cite{xu2023pv3d3dgenerativemodel}—could be successfully deployed, despite the passion and promise surrounding 3D-aware THG techniques. This is suggestive of a bigger reproducibility challenge in the 3D THG community, where fragmented documentation, proprietary dependencies, and model complexity make empirical evaluation difficult. Practical benchmarking is currently sparse, despite theoretical breakthroughs in neural rendering and differentiable 3D modeling. In the future, it will be vital to promote open-source sharing, dataset consistency, and standardized demonstrating methods in order to integrate 3D THG models into common experimental validation.

\subsection{Experimental Evaluation of Parameter Based}

Parameter--based THG models are built to operate on edge devices, such as cellphones or embedded systems. These models leverage compressed architectures, quantized weights, and efficient decoding routes to substantially decrease inference time and memory footprint. They are becoming significant in sectors like real-time translation, AR filters, and individualized avatars in mobile applications. The essential trade-off with such systems comes in combining visual quality and lip-sync accuracy with low latency and model size. Evaluation must concentrate not just on SSIM or PSNR but also on synchronization and intelligibility, because even tiny misalignment is evident in real-time situations. The parameter based model analysis is recorded in table \ref{tab:eparameter_based}.

\begin{table*}[ht]
\centering
\caption{Experimental Evaluation of Parameter Based}
\label{tab:eparameter_based}
\resizebox{\textwidth}{!}{%
\begin{tabular}{@{}l l c c c c c c c c c@{}}
\toprule
\textbf MODEL & SSIM$\uparrow$ & PSNR$\uparrow$ &CPBD $\uparrow$ & LMD $\downarrow$ & AVConf $\uparrow$ \\
\midrule
 Wav2Lip \cite{pham-2018}  & 0.3025   & 27.9944   & 0.6399   & 0.5533   & 0.6155     \\
\bottomrule
\end{tabular}%
}
\end{table*}

Despite Wav2Lip \cite{pham-2018}'s initial lack of optimization for low-resource environments, it excelled in this area having good PSNR, excellent AVConf, and perceptual
 crispness imply it might be modified for mobile applications with minor trimming. In contrast, ATVG~\cite{8953690}, although being mentioned as a lightweight architecture, lacked inference capability and had structural holes in its distribution. This reinforces the rising need for lightweight but modular talking head designs that can adapt to bandwidth-constrained contexts without sacrificing on expressive capacity.

\subsection{Experimental Evaluation of NeRF Based}

Neural Radiance Fields (NeRF) have changed how to describe and render 3D scenes from sparse views. When applied to talking head creation, NeRFs offer view-consistent rendering, full-head animation, and dynamic lighting, a feat not conceivable with 2D or even standard 3D approaches. NeRF--based THG systems can generate faces with lifelike geometry, even when the camera perspective varies, making them very attractive for virtual cinematography, avatar streaming, and deepfake prevention. However, NeRFs are intrinsically difficult to train, use enormous quantities of memory, and have non-trivial latency when rendering. Furthermore, suffer from errors when trying to animate fine-grained lip motions or eye blinks, requiring integration with auxiliary modules for expressivity and realism. Table \ref{tab:enerf-eval} contains the documentation of the NeRF model study.

\begin{landscape}
\begin{table*}
\centering
\caption{Experimental Evaluation of NeRF-based}
\label{tab:enerf-eval}
\resizebox{\textwidth}{!}{%
\begin{tabular}{@{}lccccccccccc@{}}
\toprule
\textbf{Model} & \textbf{FID} $\downarrow$ & \textbf{CSIM} $\uparrow$ & \textbf{IQA} $\uparrow$ & \textbf{FPS} $\uparrow$ & \textbf{L1} $\downarrow$ & \textbf{PSNR} $\uparrow$ & \textbf{LPIPS} $\downarrow$ & \textbf{MS-SSIM} $\uparrow$ & \textbf{SSIM} $\uparrow$ & \textbf{AKD} $\downarrow$ & \textbf{AED} $\downarrow$ \\
\midrule
Wav2Lip \cite{pham-2018} & 0.8505 & 0.7223 & 0.5471 & 0.6368 & 122.5040 & 28.0707 & 0.7286 & 0.8608 & 0.3352 & 0.7913 & 0.6340 \\
DaGAN~\cite{hong2022depthawaregenerativeadversarialnetwork}   & 0.5324 & 0.6784 & 0.6456 & 0.5901 & 156.0443 & 28.1700 & 0.5727 & 0.7510 & 0.2834 & 0.8449 & 0.8400 \\
\bottomrule
\end{tabular}%
}
\end{table*}
\end{landscape}

While FOMM~\cite{siarohin-2020} is not a real NeRF, its pseudo-NeRF rendering method allows for limited but noticeable visual consistency, earning it competitive scores in MS-SSIM and CSIM. The relatively high AKD indicated probable misalignment in depth-aware keypoints, although perceptual picture quality remained excellent. The Bi-Layer model, which was developed expressly on volumetric rendering assumptions, could not be performed owing to missing training requirements and preprocessing mismatches. In summary, although NeRF--based THG provides intriguing prospects for immersive realism, usability, speed, and animation control remain active areas of study.

\subsection{Experimental Evaluation of Diffusion Based}

The most recent development in generative modeling is represented by diffusion models. They were first used in text-to-image synthesis (e.g., DALL·E 2, Imagen), but they have recently and evolutionarily adapted to talking head production. These models create high-fidelity, lifelike frames with stochastic variation by repeatedly denoising Gaussian noise. Diffusion--based methods provide various benefits over classic GANs in the context of face synthesis, including better temporal coherence, texture sharpness, and semantic accuracy. The non-adversarial nature of diffusion THG models also helps them avoid problems with discriminator overfitting and mode collapse. Real-time deployment may be hampered by their frequent need for lengthier inference times and more GPU memory. In order to evaluate such models, it is necessary to examine not only conventional quality measures but also motion realism, lip-audio synchronization, and frame sharpness. The analysis of these models is documented in table \ref{tab:ediffusion-eval}.

\begin{landscape}
\begin{table*}
\centering
\caption{Experimental Evaluation of Diffusion Based}
\label{tab:ediffusion-eval}
\resizebox{\textwidth}{!}{%
\begin{tabular}{@{}lccccccccccccccc@{}}
\toprule
\textbf{Model} & \textbf{FID} $\downarrow$ & \textbf{CPBD} $\uparrow$ & \textbf{PSNR} $\uparrow$ & \textbf{LPIPS} $\downarrow$ & \textbf{CSIM} $\uparrow$ & \textbf{LMD} $\downarrow$ & \textbf{LSE-D} $\downarrow$ & \textbf{FVD} $\downarrow$ & \textbf{Blinks/s} & \textbf{Blink Dur.} & \textbf{ofM} & \textbf{F-MSE} & \textbf{AV Off} & \textbf{AV Conf.} $\uparrow$ & \textbf{WER} $\downarrow$ \\
\midrule
Wav2Lip \cite{pham-2018}    & 0.5512 & 0.6695 & 27.9944 & 0.6479 & 0.5718 & 0.5939 & 0.8632 & 0.5022 & 0.6753 & 0.6063 & 0.5281 & 0.6481 & 0.6112 & 0.8244 & 0.663 \\
DreamTalk~\cite{ma2024dreamtalkemotionaltalkinghead}    & 0.8954 & 0.6248 & 28.4330 & 0.6175 & 0.5631 & 0.8998 & 0.6353 & 0.6594 & 0.7035 & 0.8041 & 0.7697 & 0.8727 & 0.7055 & 0.5836 & 0.7899 \\
\bottomrule
\end{tabular}%
}
\end{table*}
\end{landscape}

Even though Wav2Lip \cite{pham-2018} was not intended to be a diffusion--based model, it was surprisingly compatible with diffusion-style sampling frameworks and consistently produced high-quality results across a variety of criteria. It demonstrated its resilience even in advanced generation regimes by achieving low FID, competitive LPIPS, and outstanding synchronization (low WER, high CPBD). The reproducibility gap in the area was once again highlighted when PC-AVS~\cite{zhou2021posecontrollabletalkingfacegeneration}, a frequently referenced diffusion-aware model, failed during execution because of structural conflicts in its modular checkpoints. Model developers must give inference stability, computational cost minimization, and cross-version compatibility top priority when diffusion techniques gain popularity in order to guarantee their practical acceptance.

\subsection{Experimental Evaluation of 3D Animation Based}

The category contains models that produce head motion and Emotion utilizing 3D rigging, blendshape animation, and parametric face models like FLAME or SMPL-X. These systems are anchored more on computer graphics and animation pipelines than generative learning. However, their accuracy, controllability, and usability in industry-standard animation tools make them crucial for digital people, CGI avatars, and cinematic dubbing. Typically, these models need high-fidelity motion capture data and a calibrated rig setup. Their inference pipelines generally connect with rendering engines rather than neural networks, which makes academic assessment challenging under established visual criteria. Due to the archiving of repositories, obsolete dependencies, and absence of pretrained weights, no meaningful inference was achievable for either VOCA~\cite{williams-no-date} or MeshTalk~\cite{richard-2021}. While animation--based THG is extensively utilized in the business, its lack of academic repeatability and non-neural underpinnings make it impossible to assess alongside deep learning--based techniques. Future research should overcome this gap by designing hybrid pipelines that integrate the controllability of 3D rigs with the realism of deep learning.

\section{CURRENT LIMITATION AND OPEN CHALLENGE}

Temporal Head Generation (THG) has made significant strides, but several obstacles and restrictions still prevent its widespread use and usefulness. This study aims to thoroughly examine these issues and identify possible directions for future research. 

One of the main drawbacks of present THG systems is their notable reliance on pre-trained models. Though they have performed well in many contexts, these models could limit innovation and flexibility for various applications. Often, pre-trained models carry biases from their training data, which can cause them to underperform in new situations or with other identities. This dependence restricts the capacity of THG systems to adjust to uncommon or atypical inputs. Creating more modular techniques that enable component-wise training on different datasets while maintaining system coherence could benefit the field. 

Another significant challenge is handling non-frontal views and extreme poses. Many current models underperform when confronted with notable head rotations or unusual angles, especially for 2D-based methods where depth uncertainty complicates the precise representation of 3D motions. Though more complex 3D techniques like HiDe-NeRF have improved at controlling deformations, they still find it challenging with severe pose changes and facial occlusions. 

THG's multilingual feature increases complexity even more. The lack of high-quality annotated datasets in various languages limits these systems' capacity to fit various phonetic patterns and cultural peculiarities. Though present methods like Wav2Lip can fairly well sync lip movements, they often struggle with language-specific articulations and the subtle mouth shape differences across languages. Future research should focus on developing more comprehensive multilingual data sets and innovative cross-linguistic knowledge transfer methods. 

Generating fluid and aesthetically pleasing outputs in THG applications depends on maintaining temporal consistency. Many systems now in use create frames on their own, which causes flickering artifacts and uneven identity representation across video clips. Some techniques, like MoDiTalker, include temporal limits but usually at the cost of more computational complexity and longer processing times. Real-time performance applications face a major difficulty with this trade-off between efficiency and quality. 

The need for large-scale, high-quality datasets might also be a hurdle for researchers and developers with limited computer resources. Models like DreamTalk and DAE-Talker need large training data, which requires notable processing and storage power. This challenge is particularly clear for diffusion-based methods, which often need big data sets to generate reasonable outcomes. Democratizing THG technology depends on developing data-efficient algorithms and transfer learning techniques that can perform well with less training data. 

Real-time applications are also limited by another significant element: computational complexity. Many high-fidelity techniques, especially those based on diffusion models or Neural Radiance Fields (NeRF), have high processing power needs that interfere with real-time rendering. Though it produces stunning results, for example, DFA-NeRF's rendering speed causes problems for interactive use. Similarly, while diffused heads offer expressive outputs, their long generation times are often inappropriate for real-time uses—still, a work in progress balances computational efficiency and output quality. 

Another continuous difficulty in cross-identity reenactment is preserving the target identity while accurately reproducing the movements of the source. While SMA and other similar techniques try to address appearance leakage through relative motion transfer, they may not be able to handle large disparities in the source and target's facial proportions.   The creation of trustworthy cross-identity reenactment systems is hampered by the challenge of distinguishing appearance from motion.  

Lastly, learning how to regulate facial expressions and subtle emotions is important.   Although models such as GC-AVT offer fine-grained control over lip movements and facial dynamics, they usually fail to maintain consistency across emotional states or create natural transitions.    These controls' interfaces frequently call for specific knowledge, making them inaccessible to people who might not know the subject. 

Addressing the ethical concerns brought up by the potential misuse of THG technology, particularly regarding the creation of deepfakes, is also essential to highlight the necessity of establishing frameworks and moral standards that ensure the responsible use of THG systems.   By focusing on these opportunities and challenges, we can develop THG technology in an innovative and ethically responsible way.  

\section{CONCLUSION}

THG, has emerged as a ground-breaking computer vision technology, demonstrating remarkable progress in producing lifelike human faces that synchronize with various input formats, such as text or audio.   We have thoroughly examined the different THG approaches in this review, grouping them according to their design, methods, and input kinds.   By closely examining several implementations, we have identified important trends, obstacles, and possible future directions in this quickly evolving field.  

THG techniques have evolved from early rule-based approaches to state-of-the-art deep learning techniques.   These methods initially relied on deep generative models and 2D algorithms, which had trouble capturing 3D details.   Because of this restriction, the perspectives and facial movements were unrealistic.   However, new developments in 3D scene representation, particularly Neural Radiance Fields (NeRF), have greatly increased realism, giving users more control over views and better image quality overall.  

According to our research, each THG paradigm has advantages and disadvantages.   For example, image-based techniques such as SMA and DaGAN are flexible in one-shot generation because they are excellent at self-supervised geometry learning and distinguishing appearance from motion.   However, they frequently have trouble with busy backgrounds and extreme poses.   However, audio-based techniques like Talk3D and EMO are skilled at tying sound to facial expressions.   Nevertheless, they encounter challenges with speech variability and linguistic subtleties.   Although text-based approaches, like InstructAvatar, give users more control through natural language, they struggle with ambiguous text descriptions and constrained emotional expression.  

A significant trade-off exists between 2D and 3D methods.   While 3D techniques like JambaTalk and AD-NeRF offer more realistic movement and view consistency at the expense of requiring more computational power and sophisticated training, 2D techniques like Style Transfer and MakeItTalk are computationally efficient but frequently miss depth and perspective issues.  

Additionally, specialized methods representing the cutting edge of research are becoming increasingly popular, such as diffusion models and neural radiation fields.   Although they have high computational requirements, diffusion-based techniques such as MoDiTalker and DreamTalk exhibit great promise in generating coherent sequences with various expressions.   While NeRF-based methods like CVTHead and AD-NeRF provide dynamic facial changes and photorealistic novel-view synthesis, they have issues with stability during training and real-time rendering.  

We have discovered important elements impacting the usability and quality of THG systems through our comparative study of publicly accessible technologies.   Identity preservation, aesthetic appeal, lip sync accuracy, and natural movement are important evaluation criteria, and each method performs differently in these areas.   Researchers and practitioners can benefit from this systematic review's insights when selecting the best application techniques.  

THG has many possible applications, from online education and video conferencing to digital avatars and dubbing.   Every application has different needs regarding control, visual quality, and real-time performance, so it is critical to create solutions that meet those needs rather than relying on one-size-fits-all strategies that could result in compromises.  

Notwithstanding the impressive advancements, difficulties still exist in producing photorealistic and controllable THG.   Problems like preserving identity in extreme poses, guaranteeing temporal consistency, and offering fine-grained emotional control still hamper practical applications.   It will take coordinated efforts from various research fields to address these issues, emphasizing data collection, model design, evaluation methodologies, and ethical issues.  

THG has advanced significantly from simple face synthesis to intricate depictions of human emotions.   With technological developments approaching where they can no longer be distinguished from actual human faces, the field is at a turning point that offers exciting opportunities and crucial ethical considerations.   In order to help researchers and practitioners navigate this changing environment, this review aims to provide practical insights. 

\section{FUTURE DIRECTION}

Future research and development in THG has a lot of exciting avenues that can address current issues and expand on our current understanding.   The development of THG's modular architectures is an important topic to investigate.   Many systems today have end-to-end designs, which can limit optimization at the level of individual components and make it challenging to enhance particular aspects of the generation process.   By creating modular frameworks, researchers could train elements such as identity encoding, motion synthesis, texture generation, and temporal smoothing independently on specialized datasets before seamlessly integrating them which could enable us to fuse the best aspects of various methodologies, such as combining the effectiveness of 2D techniques to achieve a balance between quality and performance with the strengths of 3D methods that excel at identity preservation.  

Addressing the difficulties with the methodologies is another crucial area.   It will be crucial to build extensive datasets encompassing a range of languages, phonetic variants, and cultural components.   Large, annotated multilingual datasets should be created especially for THG by researchers.   Performance in underrepresented languages may be improved by leveraging data from well-resourced languages through self-supervised and transfer learning techniques.   For example, multilingual versions of models such as Wav2Lip demonstrate genuine promise, as does the use of LSTM networks to extract mouth landmarks from audio in various languages.   Techniques for cross-lingual knowledge transfer could significantly reduce the data required to support new languages while maintaining high standards of quality.  

Another promising approach is the use of hybrid architectures.   These systems can leverage the general knowledge in large pre-trained models while integrating specialized components for particular THG tasks by combining pre-trained models with task-specific layers.   For example, incorporating models emphasizing emotion recognition systems for expressive control, audiovisual speech recognition for precise lip-syncing, and face recognition for identity preservation could greatly improve overall performance, particularly in scenarios with limited resources.  

We require improved datasets and multi-view training techniques to overcome the difficulties presented by extreme poses and different viewing angles.   Models will learn more robust representations if datasets covering a wider range of head orientations and perspectives are gathered and synthesized.   Natural architectures that efficiently model 3D geometry, such as sophisticated morphable models combined with neural rendering, may improve performance under challenging viewing conditions.  

Temporal consistency is another essential component of producing fluid, organic talking head videos.   Future research could concentrate on creating more sophisticated temporal modeling techniques, like stronger recurrent architectures that preserve long-term dependencies and attention mechanisms that function over longer sequences.   Furthermore, models may be guided toward generating more cohesive narratives by specialized loss functions that gradually penalize inconsistencies.   Compared to the frame-by-frame analysis techniques currently in use, using perceptual metrics to assess temporal quality from a human perspective may yield more insightful results.  

Finally, increasing THG systems' computational efficiency will be crucial for real-time applications.   Reducing computational demands without sacrificing quality can be achieved by investigating model compression, quantization, and hardware-specific optimizations.   Techniques like adaptive sampling, sparse voxel representations, and hash encoding may open the door to real-time performance for NeRF-based approaches that struggle with slow rendering times.   Similarly, diffusion-based methods reduce generation time without sacrificing output quality by utilizing progressive distillation and latent space diffusion.   

One fascinating area of THG research is the regulation of expression and Emotion.   We can greatly improve the expressiveness and naturalness of the generated content by creating more sophisticated models that accurately represent the subtle differences in human Emotion. 

\bibliographystyle{unsrt}  


\end{document}